\documentclass[smallcondensed]{svjour3}     
\usepackage{graphicx}
\usepackage{amssymb}
\usepackage{amsmath}
\usepackage{subfigure}
\usepackage{multirow,tabularx}
\usepackage{booktabs}
\usepackage{algorithm}
\usepackage{algorithmic}
\usepackage{threeparttable}
\newcommand*{\affaddr}[1]{#1} 
\newcommand*{\affmark}[1][*]{\textsuperscript{#1}}

\begin{document}

\title{A Block-based Generative Model for Attributed Network Embedding}

\author{Xueyan~Liu\protect\affmark[1, 2]  \and
        Bo~Yang\affmark[1, 2] \footnote{Corresponding author}\and 
        Wenzhuo~Song\affmark[1, 2] \and
        Katarzyna~Musial\affmark[3] \and
        Wanli~Zuo\affmark[1, 2] \and
        Hongxu~Chen\affmark[3]   \and
        Hongzhi~Yin\affmark[4]
}

\authorrunning{X. Liu, B. Yang, W. Song et al} 

\institute{Xueyan~Liu \at
              \email{xueyanl17@mails.jlu.edu.cn}
           \and
           Bo~Yang \at
              \email{ybo@jlu.edu.cn}
           \and
           Wenzhuo~Song \at
              \email{songwz17@mails.jlu.edu.cn}
           \and
           Katarzyna~Musial\at
           \email{Katarzyna.Musial-Gabrys@uts.edu.au}
           \and
           Wanli~Zuo\at
            \email{zuowl@jlu.edu.cn}
            \and
            Hongxu~Chen \at
           \email{Hongxu.Chen@uts.edu.au}
           \and
            Hongzhi~Yin \at
            \email{h.yin1@uq.edu.au}
            \\
           \affaddr{\affmark[1] School of Computer Science and Technology, Jilin University, Changchun 130012, China}\\
          \affaddr{\affmark[2] Key Laboratory of Symbolic Computation and Knowledge Engineer (Jilin University), Ministry of Education, Changchun 130012, China}\\
          \affaddr{\affmark[3] Advanced Analytics Institute, School of Computer Science, Faculty of Engineering and IT, University of Technology Sydney, NSW, Australia, 2007}\\
          \affaddr{\affmark[4] School of Information Technology and Electrical Engineering, The University of Queensland, Brisbane, QLD, Australia, 4072}\\
}

\date{Received: date / Accepted: date}

\maketitle

\begin{abstract}
Attributed network embedding has attracted plenty of interest in recent years. It aims to learn task-independent, low-dimensional, and continuous vectors for nodes preserving both topology and attribute information. Most of the existing methods, such as random-walk based methods and GCNs, mainly focus on the local information, i.e., the attributes of the neighbours. Thus, they have been well studied for assortative networks (i.e., networks with communities) but ignored disassortative networks (i.e., networks with multipartite, hubs, and hybrid structures), which are common in the real world. To enable model both assortative and disassortative networks, we propose a block-based generative model for attributed network embedding from a probability perspective. Specifically, the nodes are assigned to several blocks wherein the nodes in the same block share the similar linkage patterns. These patterns can define assortative networks containing communities or disassortative networks with the multipartite, hub, or any hybrid structures. To preserve the attribute information, we assume that each node has a hidden embedding related to its assigned block. We use a neural network to characterize the nonlinearity between node embeddings and node attributes. We perform extensive experiments on real-world and synthetic attributed networks. The results show that our proposed method consistently outperforms state-of-the-art embedding methods for both clustering and classification tasks, especially on disassortative networks.
\end{abstract}

\section{Introduction}
 Attributed networks model complex systems, with sometimes very complex features, in a systematic and simplified way. As oppose to the pure networks only considering the relationships between the nodes, attributed networks provide much richer and heterogeneous information about the systems due to the fact that, during the modelling process, they include node features. For example, in social networks, the attributes provide individuals' gender, nationality, location, and interests. In the protein-protein interaction networks, a protein is defined by the amino acid types, the protein structures ($\alpha$-helices, $\beta$-sheet or turns), etc. A paper consists of a title, keywords, authors, and venue in the academic citation networks. Thus, studying attributed networks is particularly important for real-world networks and their applications.

Recently, attributed network embedding or representation learning (RL) has become a research hotspot. RL methods aim to map nodes to low-dimension and continuous embeddings, while preserving both the topological properties and the attribute information of the attributed networks. Compared with traditional methods for specific network analysis tasks, including node clustering \cite{gao2010community,ruan2013efficient}, node classification \cite{silva2018social}, link prediction \cite{barbieri2014follow,wahid2019predict}, and outlier and change point detection \cite{perozzi2014focused} \cite{kendrick2018change}, RL methods are task-independent. Therefore, the learned embeddings can be used by off-the-shelf methods to perform the downstream network analysis tasks. 

In general, the representations of nodes in the same cluster (or block) \footnote{In this paper, \emph{cluster}, \emph{group}, and \emph{block} are interchangeable.} are similar so that traditional cluster/classification methods can use them for network analysis. Learning representations for assortative networks and disassortative networks are both critical because these two types of networks are common in the real world. Both assortative and disassortative networks are defined by the kinds of groups they contained. In \cite{newman2016structure}, Newman and Clauset described the assortative structure as a group of nodes, in which the links are denser in the same group than between different groups. In contrast, the links are sparser intra-groups than inter-groups for disassortative structures. Given this, we refer to the networks that contain only assortative structures as assortative networks. In other words, if a network is assortative, it satisfy the following condition: the nodes are connected densely in the same group and sparsely between the different groups. Similarly, we define the disassortative networks as networks containing at least a disassortative structure, i.e., at least a block containing nodes that connect sparsely within the block and densely between the blocks. For example, the academic citation networks are assortative networks since the papers are more likely to cite other papers in the same field. A food web is a disassortative network because a predator links densely to the preys and rarely connects to other predators. According to the above definition, a node's representation is highly correlated to its proximal node for assortative networks. However, for disassortative networks, the representations of two nodes in the same cluster also should be similar even if they are far away from each other in a geodesic sense.

However, most of the existing RL methods, such as random-walk based methods \cite{pan2016tri,huang2019large,gao2018deep,liao2018attributed,ijcai2018-438} and graph neural networks (GNNs) based methods \cite{kipf2016semi,hamilton2017inductive,kipf2016variational,velivckovic2017graph,pan2018adversarially,mehta2019stochastic}, are designed for assortative networks, ignoring the disassortative networks. For example, random-walk based methods \cite{pan2016tri,huang2019large,gao2018deep,liao2018attributed,ijcai2018-438} learn the structural similarity between node pairs using the concept of neighbours defined by $k-$step random walk before learning embeddings. Intuitively, using random walk approaches, ``close'' nodes are more likely to co-occur in node sequences. Thus, the representation of a node is more similar to that of nodes in short distance than long distance \cite{ribeiro2017struc2vec}.
Most of the GNN-based methods learn node embeddings by aggregating only the information from nodes in close proximity \cite{ICLR2020GeomGCN}. In this way, the embeddings of nodes depend greatly on their neighbour's messages. Besides, many matrix factorization-based methods \cite{huang2017accelerated,yang2018enhanced,yang2018binarized}  learned attributed network representation by factoring the matrices that are constructed based on the network topology and node features. However, they cannot depict the relations between the different properties of the attributes in a nonlinear way. More details about these methods are discussed in Related Work (Section \ref{RelatedWork}).

Although many efforts were devoted to addressing this problem, to develop a unified representation learning method for both assortative and disassortative attributed networks is still a challenging problem. For example, Gao \emph{et al.} \cite{gao2018bine} proposed BiNE (Bipartite Network Embedding) to learn node embeddings for bipartite networks, in which the nodes connect sparsely to each other in the same cluster/type but densely between different clusters/types. They first constructed two new networks for each kind of nodes and then performed random walk on the original and newfound networks, respectively. BiNE requires prior knowledge about types of networks and labels of the nodes before designing an appropriate RL model. However, in real-world networks, those are unknown and expensive to be collected. It is essential to design a general attributed network embedding method for both assortative and disassortative networks, especially when their types are unknown.

Another approach that attempts to deal with both assortative and disassortative networks is the stochastic block model (SBM) \cite{holland1983stochastic}. SBM is commonly used to characterize networks with complicated structural patterns, including communities, multipartite, hubs, and hybrid structures. Recently, various extensions of SBM are presented for different tasks, including structural pattern detection \cite{abbe2017community}, link prediction \cite{guimera2009missing}, signed networks analysis \cite{jiang2015stochastic,yang2017stochastic}, and dynamic networks evolution \cite{yang2011detecting}. However, these SBMs only consider the network topology, so they are not suitable for attributed networks. Additionally, the obtained embeddings of nodes by SBM are not general features but defined as assignment relationships between nodes and blocks, which will limit its application.

In light of the above problems, we propose a novel attributed network generative model (ANGM) for both assortative and disassortative networks and its learning algorithm inspired by the stochastic block model and neural networks. Specifically, we use the concept of ``block'' and ``block-block'' link probability matrix as model parameters to describe the generative process of the topology of networks with diverse structural patterns. It is worth noting that the block-block link probability matrix defines the types of networks in terms of the structural patterns. For example, if the diagonal entries of the matrix are higher than the off-diagonal entries, it depicts assortative networks, otherwise, it describes disassortative networks. This matrix will be optimized if we fit the model to real-world networks. Thus, ANGM is able to model both assortative and disassortative networks without prior knowledge about the types of networks. We introduce neural networks to integrate node attributes to our model. We assume the embeddings of nodes in the same block are similar, and then use a neural network to characterize the nonlinearity between the node embeddings and node attributes. Different from SBM, we use two latent variables to model node assignment and node embedding, respectively. Thus, the embeddings in our model can be used by different downstream tasks. Finally, we unify the generative process of nodes' links and attributes to a probabilistic graph model.

The main contributions of this paper are as follows:
\begin{itemize}
  \item[(1)] We propose a block-based attributed generative model (ANGM) for attributed network embedding. Besides assortative networks, ANGM can deal with \emph{disassortative networks}, which are almost ignored by existing methods.
  \item[(2)] We propose a variational learning method for estimating the parameters and the latent variables of ANGM by maximizing the ELBO, which can use a simple distribution to approximate the intractable distribution.
  \item[(3)] We conduct extensive validations and comparisons on different downstream tasks, including node clustering, classification, and visualization, using both assortative and disassortative attributed networks. The results show that ANGM outperforms many state-of-the-art algorithms, especially for dealing with disassortative networks.
\end{itemize}

The rest of the paper is organised as follows. In Section \ref{RelatedWork}, we review and discuss the state-of-the-art methods for attributed network embedding. In Sections \ref{model} and \ref{Method}, we present the attributed network generative model and its learning method. In Section \ref{Experiments}, we test our method on both synthetic and real-world datasets for different network analysis tasks, including node clustering, node classification, and visualization of representation. Finally, we conclude and summarize our proposed model and method in Section \ref{Conclusion}.

\section{Related Work} \label{RelatedWork}
In recent years, attributed network embedding or representation learning (RL) has become a prominent research area that focuses on learning low-dimension, continuous, and task-dependent node embeddings/representations. Unlike the graph embedding methods designed for pure networks \cite{perozzi2014deepwalk,node2vec-kdd2016,jiang2018spectral}, the attributed network embedding methods can preserve both the topology and the attribute information. In this section, we review and discuss the state-of-the-art attributed network embedding methods. According to their principles, we divide them into three folds: matrix factorization based methods, random-walk based methods, and graph neural network based methods.

\subsection{Matrix factorization based methods}
Matrix factorization (MF) based methods construct matrices based on the network properties, such as the network topology and the node attributes, and then factorize them to obtain the node representations. In 2015, Yang \emph{et al.} first proved that the DeepWalk algorithm \cite{perozzi2014deepwalk} is equal to factorizing a matrix built based on the walking probability. Inspired that, they introduced text information into MF to use both structural and text information for network embedding. Accelerated attributed network embedding (AANE) \cite{huang2017accelerated} transformed the node attributes into the similarity matrix and then decomposed the matrix by cooperating the edge-based penalty to learn node representations efficiently. Binarized attributed network embedding (BANE) \cite{yang2018binarized} constructed a Weisfeiler-Lehman proximity matrix to aggregate structural and attributed information and then formulated a factorization learning function for the proximity matrix to learn binary node embeddings faster. To solve the problem that networks are sparse in the real world, Yang \emph{et al.} \cite{yang2018enhanced} considered the node text groups and then used the consistent relationships between the text clustering and node representations to learn the network embeddings under the nonnegative matrix factorization framework. However, the above MF based methods use the liner functions to learn network representations. Thus, they could not characterize the relations between different properties of the networks nonlinearly.

\subsection{Random-walk based methods}
Random-walk based methods for attributed networks are mainly extended from DeepWalk \cite{perozzi2014deepwalk} and Node2Vec \cite{node2vec-kdd2016} to learn the node embeddings. For example, Pan \emph{et al.} \cite{pan2016tri} used the random-walk to model the structural information and then adopted Paragraph2Vec to describe the relations among the nodes, the attributes, and the labels. Feat-Walk \cite{huang2019large} learnt node sequences by performing random-walk on the node-node network and distributed feature-walk on the node-attribute network, and then fed node sequences to the scalable word embedding algorithms to learn node embedding. Deep attributed network embedding (DANE) \cite{gao2018deep}, attributed social network embedding (ASNE) \cite{liao2018attributed}, and attributed network representation learning (ANRL) \cite{ijcai2018-438} first learnt the structural proximity through executing random-walk or calculating the $k-$order neighbours and then combined Word2Vec and deep neural networks together to encode structural and attributed proximity to the embeddings nonlinearly. As we know, if two nodes are closer, they are more likely to co-occur in the node sequences after executing random walk on the network. The co-occurrence of nodes makes their embbeddings similar after we feed node sequences to the word2vec method. Thus, we can deduce that the embeddings of nodes are similar if they are densely connected. Therefore, these methods are only suitable for assortative networks but not applicable to disassortative networks.

\subsection{Graph neural networks}
Unlike random-walk based methods, graph neural network (GNN) based methods for attributed network embedding are inductive.  It means that we can learn the node embeddings for newly coming nodes without retraining the models. Among all GNNs, graph convolutional network (GCN) \cite{kipf2016semi} is the most popular one. Hamilton \emph{et. al.} concluded the GCN and its variations to message passing algorithms. They adopted various aggregators to learn the node embeddings by aggregating the local attribute information \cite{hamilton2017inductive}.
Graph attention network (GAT) \cite{velivckovic2017graph,knyazev2019understanding} introduced attention mechanism to describe the impact of valuable information on node embeddings. Graph wavelet neural network (GWNN) used graph wavelet as a set of bases and regarded wavelet transform as the convolution operator \cite{xu2018graph}. Graph U-Nets \cite{gao2019graph} extended pooling operations to attributed network embedding. Gamma \emph{et al.} \cite{gama2019diffusion} defined graph scattering transformation using diffusion wavelets to obtain stable network representation. These methods are discriminative, which means they usually require prior knowledge about labels of nodes to predefine the objective functions elaborately, which profoundly influence their performance. However, gaining proper prior information is expensive. To solve this problem, many researchers proposed generative GNN methods, which generate new samples according to probability theory, and then they regarded the gap between real and generative samples as the objective function. For example, variational graph auto-encoder (VGAE) \cite{kipf2016variational} considered a two-layer GCN as an encoder to learn the node embeddings, then calculated the link probability between two nodes according to the inner product of their embeddings, finally decoded the network topology according to the link probability. Adversarial regularized variational graph auto-encoder (ARVGA) \cite{pan2018adversarially} incorporated the adversarial model to the VGAE for robust representation learning. Based on the decoding process in terms of inner product, VGAE and ARVGA assume that the more similar the embeddings of two nodes are, the more likely they are connected, which are consistent with assortative networks. From the perspective of the generative model, these two methods can only generate the topological structure of networks but not the node attributes. The graph attention auto-encoder (GATE) \cite{salehi2019graph} used an attention machine to encode the node attributes to node representations and reversed the encoding process to generate the node attributes to solve this problem. Also, it utilized the links to minimize the difference between the two linked nodes' embeddings. Deep generative latent feature relational model (DGLFRM) \cite{mehta2019stochastic}, like VGAE and ARVGA, also used GCN as an encoder to obtain node embeddings, but then it used latent feature relational model and neural networks as decoders to generate links and attributes, respectively. DGLFRM defined node embedding by nodes' membership and the strength of the membership, which limits the application scopes of the embeddings. Besides, Graph2Gauss (G2G) \cite{bojchevski2018deep} first obtained each node embedding's distribution. It derived an unsupervised loss function, which satisfied that the shortest path length between two nodes is smaller, their distributions are more similar.   All the above GNN methods learn embedding of a node by aggregating attribute information of its neighbours, which means that the embeddings of two nodes in the long distance are irrelevant. Therefore, they can only perform well on assortative networks but not on disassortative networks.

\section{Problem Statement}
In this section, we first summarize the main notations used in this paper and formally define the problem of attributed network embedding.

Let $\mathcal{G} = (\mathcal{V}, \mathcal{E}, \pmb{X})$ denote an attributed network with $n$ nodes, and each node has $M$-dimension attributes. $\mathcal{V}$ and $\mathcal{E}$ are the sets of nodes and edges, respectively.  $\pmb{X} \in \{0,1\}^{n\times M}$ or $\pmb{X} \in \mathbb{R}^{n\times M}$ denotes the binary or continuous attribute matrix, and each row $\pmb{x}_i$ refers to the attributes of node $i$. $\pmb{A} \in \{0,1\}^{n\times n}$ is the adjacency matrix of $\mathcal{G}$, where $a_{ij} =1$ denotes node $i$ links to node $j$, otherwise $a_{ij}=0$.Table \ref{tab:nota} shows the main notations for describing the attributed network and the proposed model in this paper.

We define the problem of attributed network embedding as follows: Given an attributed network $\mathcal{G}$, we aim to learn attributed networks' embedding $\pmb{Z} \in \mathbb{R}^{n\times D}$, where $\pmb{z}_i \in \mathbb{R}^{1\times D}$ is a low-dimensional vector representation of node $i$, and $D$ is the dimension of the embedding for each node.

\begin{table}[htpb]
    \centering
    \begin{tabular}{m{0.35\columnwidth}<{\centering}m{0.65\columnwidth}}
    \toprule
      notations   &  Definitions\\
    \midrule
     $\mathcal{G} = (\mathcal{V}, \mathcal{E}, \pmb{X})$  &  Attributed network\\
     $n$ & Number of nodes in $\mathcal{G}$\\
     $M$ & Dimension of attributes in $\mathcal{G}$\\
     $\pmb{A} \in \{0,1\}^{n\times n}$ & Adjacency matrix of  $\mathcal{G}$\\
    $\pmb{X} \in \{0,1\}^{n\times M}$ or $\pmb{X} \in \mathbb{R}^{n\times M}$ & Binary or continuous attribute matrix of $\mathcal{G}$\\
    $K$ & Number of blocks in $\mathcal{G}$\\
    $D$ & Dimension of node embeddings \\
    $\pmb{Z} \in \mathbb{R}^{n\times D}$ & Node embedding matrix\\
    $c_i \in \{1,2,...,K\}$ & Block of node $i$\\
    $\pmb{\omega}  \in [0, 1]^{1\times K}$ & Node assignment probability vector\\
    $\pmb{\Pi} \in [0,1]^{1\times K}$ & Block-block link probability matrix\\
    $\pmb{\mu}_{k} \in \mathbb{R}^{1\times D}$ & Mean of node embeddings in block $k$\\
    $\pmb{\sigma}_{k}\in \mathbb{R}^{1\times D}$ & Standard deviation of node embeddings in block $k$\\
    $\pmb{\upsilon}_i \in [0,1]^{1\times M}$ or $\pmb{\upsilon}_i \in \mathbb{R}^{1\times M}$ & Probability of node $i$ have the attributes if $\pmb{X} \in \{0,1\}^{n\times M}$  or mean of the attributes of node $i$ if $\pmb{X} \in \mathbb{R}^{n\times M}$\\
    $\pmb{\lambda}_i$ & Standard deviation of attributes of node $i$ if $\pmb{X} \in \mathbb{R}^{n\times M}$\\
    $i$, $j$ & index of nodes\\
    $k$, $l$ & index of blocks\\
    $m$ & index of attribute for each node\\
    $d$ & index of embedding for each node\\
       \bottomrule
    \end{tabular}
    \caption{Table of Notations}
    \label{tab:nota}
\end{table}

\section{The Attributed Network Generative Model}\label{model}
In this section, we propose an attributed network generative model for both assortative and disassortative attributed network embedding.
\begin{figure}[t]
\centering
\includegraphics[width=0.5\linewidth]{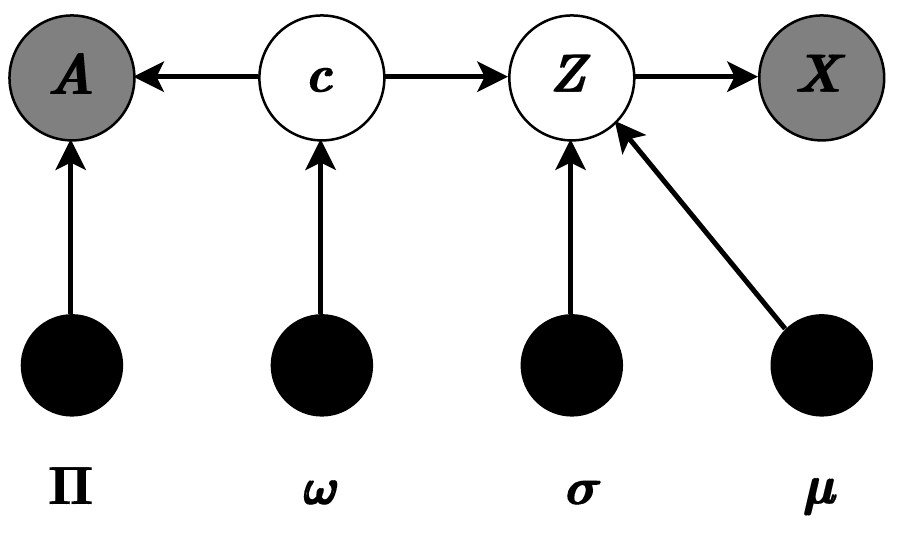}
\caption{The probabilistic graphical model of ANGM.}
\label{graphic_model}
\end{figure}

In this work, we introduce a concept of ``block'' to our embedding method, and we can use ``block" to model the hidden patterns for both attributes and topology of attributed networks. In the standard stochastic blockmodel, a block is a subset of similar nodes in terms of connections in a given network \cite{yang2011characterizing}. In our paper, we add node attributes to the concept of ``block", assuming that the nodes' attributes in the same block are similar. For example, we can group the papers from the same field into a \emph{block} in the citation networks. The papers in the same block are more likely to cite each other and seldom mention the papers from the other fields. Their attributes, such as keywords, venues, and titles, are also more similar in the same areas (blocks) than those from different fields. Specifically, we make four assumptions for our model: (a) a node belongs to one of $K$ blocks; (b) embeddings of nodes in the same block are similar; (c) node embeddings and node attributes are related nonlinearly; (d) nodes in the same blocks share similar linkage patterns. For example, we can describe an assortative networks with communities as follows: the link probability between any two nodes intra-blocks is larger than that inter-blocks. We can also depict a disassortative networks with multipartite structures. The link probability between any two nodes intra-blocks is less than that inter-blocks. Under these assumptions, the node embeddings depends more on the nodes who have the similar structural patterns with them than their neighbors. In this way, we can learn the node embeddings for both the assortative and the disassortative networks.

Mathematically, we define an attributed network generative model (ANGM) as a 4-tuple:
 \begin{equation}\label{ANGM}
  ANGM = (\pmb{\omega}, \pmb{\Pi}, \pmb{\mu}, \pmb{\sigma}).
\end{equation}
The $K$-dimensional vector $\pmb{\omega}$ refers to the node assignment probability, wherein $\omega_k$ denotes how likely it is that a node belongs to block $k$, and it satisfies the criterion $\sum_{k=1}^{K}\omega_k=1$. $\pmb{\Pi}$ is a $K \times K$ matrix, where $\pi_{kl}$ denotes the probability that two nodes in blocks $k$ and $l$ will be connected. $\pmb{\mu}$ and $\pmb{\sigma}$ are two $K\times D$ matrices. $\pmb{\mu}_k$ and $\pmb{\sigma}_k$ denote the mean and the standard deviation of the embeddings of the nodes in block $k$, respectively.

Given an attributed network, we can deduce two latent variables: membership vector $\pmb{c} = <c_1, c_1,..., c_n>$ and embedding matrix $\pmb{Z} \in \mathbb{R}^{n\times D}$, wherein $c_i\in\{1,2,...,K\}$ denotes that node $i$ belongs to block $c_i$, and vector $\pmb{z}_i\in \mathbb{R}^{1\times D}$ denotes the embedding of the node $i$. Figure \ref{graphic_model} shows the probabilistic graphical model of ANGM. In this Figure, $\pmb{A}$ and $\pmb{X}$ are observed data; $\pmb{Z}$ and $\pmb{c}$ are latent variables; $\pmb{\Pi}$, $\pmb{\mu}$, $\pmb{\sigma}$ are model parameters.

Based on ANGM and our assumptions, the generation process of an attributed network is designed as follows:
\begin{itemize}
\item[1.] For each node $i$:
       \begin{itemize}
        \item[(a)] Assign node $i$ to one of $K$ blocks according to a multinomial distribution: $c_i\sim mul(\pmb{\omega})$;
        \item[(b)] Generate the embedding of node $i$ according to its membership and a Gaussian distribution: $\pmb{z}_i\sim \mathcal{N}(\pmb{\mu}_{c_i}, \pmb{\sigma}_{c_i}^2\textbf{I})$;
        \item[(c)] Generate the attributes of node $i$:
                \begin{itemize}
                  \item[$\bullet$] if $\pmb{x}_i$ is binary, i.e., $\pmb{x}_i \in \{0,1\}^{1\times M}$, $\pmb{x}_i$ is generated according to a Bernoulli distribution: $\pmb{x}_i\sim Ber(\pmb{\upsilon}_i)$, where $\pmb{\upsilon}_i= f(\pmb{z}_i;\theta)$, $\pmb{\upsilon}_i \in [0,1]^{1\times M}$.
                  \item[$\bullet$] if $\pmb{x}_i$ is continuous, i.e., $\pmb{x}_i \in \mathbb{R}^{1\times M}$, $\pmb{x}_i$ is generated according to a Gaussian distribution: $\pmb{x}_i\sim \mathcal{N}(\pmb{\upsilon}_i,\pmb{\lambda}_{i}^2\textbf{I})$, where $[\pmb{\upsilon}_i,\log\pmb{\lambda}_{i}^2]= f(\pmb{z}_i;\theta)$, and $\pmb{\upsilon}_i \in  \mathbb{R}^{1\times M}, \pmb{\lambda}_{i} \in  \mathbb{R}^{1\times M}$.
                \end{itemize}
                $f(\pmb{z}_i;\theta)$ denotes the neural networks parameterized by $\theta$, the input of $f$ is $\pmb{z}_i$, and the output is the parameters of the Bernoulli distribution or the Gaussian distribution for the generation of node attributes. $f$ models the nonlinearity between node embeddings and node attributes.
       \end{itemize}
\item[2.] For each node pair ($i,j$):
        \begin{itemize}
        \item[$\bullet$] Generate the link between node $i$ and node $j$ according to their memberships and a Bernoulli distribution: $a_{ij}\sim Ber(\pi_{c_i, c_j})$.
       \end{itemize}
\end{itemize}

According to the probabilistic graphical model shown in Figure~\ref{graphic_model} and the generation process, the likelihood of the complete-data is written as (see Appendix for more details):
\begin{equation}\label{joint_prob}
\begin{split}
  &p(\pmb{X},\pmb{A},\pmb{Z},\pmb{c}|\pmb{\Pi},\pmb{\omega}, \pmb{\sigma},\pmb{\mu})\\
  &=p(\pmb{A}|\pmb{c},\pmb{\Pi})p(\pmb{X}|\pmb{Z})p(\pmb{Z}|\pmb{c},\pmb{\sigma},\pmb{\mu})p(\pmb{c}|\pmb{\omega})\\
  &=\prod_{ij}\pi_{c_ic_j}^{a_{ij}}(1-\pi_{c_ic_j})^{1-a_{ij}}\times\prod_{im}\upsilon_{im}^{x_{im}}(1-\upsilon_{im})^{(1-x_{im})}\\
 &\quad\times\prod_{id}\frac{1}{\sqrt{2\pi}\sigma_{c_id}}e^{-\frac{(z_{id}-\mu_{c_id})^2}{2\sigma_{c_id}^2}}\times\prod_i\omega_{c_i}.
\end{split}
\end{equation}

The proposed generative model for attributed networks has two advantages. (1) It can generate networks with different structural patterns by setting different $\pmb{\Pi}$. For example, we can generate networks with communities by setting $\pi_{kk} > \pi_{kl}$ for $k \neq l$ and  multipartite structures if $\pi_{kk} < \pi_{kl}$. (2) It defines the similarity of the node embeddings from the perspective of ``block'' instead of ``neighbours''. Thus, it considers global information of networks.

\section{The Learning Method}\label{Method}
In this section, we will introduce the learning algorithm for ANGM by fitting the model to a given attributed network and maximizing the likelihood of the data.

Based on Eq.~(\ref{joint_prob}), the log-likelihood of the observed data is 
\begin{equation}\label{log-likh}
  \log p(\pmb{A},\pmb{X}|\pmb{\Pi},\pmb{\omega}, \pmb{\sigma},\pmb{\mu}) = \log\int_{\pmb{Z}} \sum_{\pmb{c}} p(\pmb{X},\pmb{A},\pmb{Z},\pmb{c}|\pmb{\Pi},\pmb{\omega}, \pmb{\sigma},\pmb{\mu})d\pmb{Z}.
\end{equation}

Our goal is to maximize $\log p(\pmb{A},\pmb{X}|\pmb{\Pi},\pmb{\omega}, \pmb{\sigma},\pmb{\mu})$ for finding the optimal model for a given attributed network. However, it is intractable to calculate Eq.~(\ref{log-likh}) directly. Thus, we introduce a decomposable variational distribution $q(\pmb{Z}, \pmb{c}|\pmb{X})$, which is approximated to the intractable posterior distribution $p(\pmb{Z},\pmb{c}|\pmb{A},\pmb{X})$, and then we use Jensen's inequality to gain the lower bound of Eq.~(\ref{log-likh}).  Alternatively, we will maximize the log-likelihood's lower bound as shown in Eq. (\ref{ELBO-orig}).
\begin{equation}
\begin{split}\label{ELBO-orig}
\log p(\pmb{A},\pmb{X}|\pmb{\Pi},\pmb{\omega}, \pmb{\sigma},\pmb{\mu})
 &= \log\int_{\pmb{Z}} \sum_{\pmb{c}} q(\pmb{Z},\pmb{c}|\pmb{X})\frac{p(\pmb{X},\pmb{A},\pmb{Z},\pmb{c}|\pmb{\Pi},\pmb{\omega}, \pmb{\sigma},\pmb{\mu})}{q(\pmb{Z},\pmb{c}|\pmb{X})}d\pmb{Z}\\
  &\geq \int_{\pmb{Z}} \sum_{\pmb{c}} q(\pmb{Z},\pmb{c}|\pmb{X}) \log \frac{p(\pmb{X},\pmb{A},\pmb{Z},\pmb{c}|\pmb{\Pi},\pmb{\omega}, \pmb{\sigma},\pmb{\mu})}{q(\pmb{Z},\pmb{c}|\pmb{X})}d\pmb{Z}\\
  &=E_{q(\pmb{Z}, \pmb{c}|\pmb{X})}\big[\log\frac{p(\pmb{X},\pmb{A},\pmb{Z},\pmb{c}|\pmb{\Pi},\pmb{\omega},\pmb{\sigma},\pmb{\mu})}{q(\pmb{Z},\pmb{c}|\pmb{X})}\big]\\
 &=\mathcal{L}(\pmb{A},\pmb{X})
\end{split}
\end{equation}
According to the mean-field theory, we know  $q(\pmb{Z},\pmb{c}|\pmb{X})$ can be factorized as
\begin{equation}\label{mean-field}\notag
  q(\pmb{Z},\pmb{c}|\pmb{X}) = q(\pmb{Z}|\pmb{X})q(\pmb{c}).
\end{equation}

We use neural networks $g$ parameterized by $\phi$ to calculate $q(\pmb{Z}|\pmb{X})$. The input is the node attribute $\pmb{X}$, and the outputs are the parameters of the Gaussian distribution for the embeddings of nodes. For each node $i$,
\begin{equation}
\begin{split}\label{q-Z}
  [\pmb{\hat{\mu}}_i,\log\pmb{\hat{\sigma}}_i^2]= g(\pmb{x}_i;\phi), \\
  q(\pmb{z}_i|\pmb{x}_i) = \mathcal{N}(\pmb{\hat{\mu}}_i,\pmb{\hat{\sigma}}_i^2\textbf{I}),
\end{split}
\end{equation}
where $\pmb{\hat{\mu}}_i, \pmb{\hat{\sigma}}_i^2 \in \mathbb{R}^{1 \times D}$.

Then we assume that
\begin{equation}\label{q-C}
  q(\pmb{c}_i) = mul(\tau_{i1}, \tau_{i2},...,\tau_{iK})
\end{equation}
where $\tau_{ik}$ denotes the probability of node $i$ belonging to block $k$.

Thus, we can obtain $\mathcal{L}(\pmb{A},\pmb{X})$ according to Eqs.~(\ref{joint_prob}), (\ref{ELBO-orig}), (\ref{q-Z}), and (\ref{q-C}) as follows:
\begin{equation}\label{ELBO}
\begin{split}
&\mathcal{L}(\pmb{A},\pmb{X})\\
& = \sum_{ij}\sum_{kl}\tau_{ik}\tau_{jl}[a_{ij}\log\pi_{kl}+(1-a_{ij})\log(1-\pi_{kl})]\\
&\quad +\frac{1}{L}\sum_{l=1}^{L}\sum_{i=1}^n\sum_{m=1}^M[x_{im}\log \upsilon_{im}^{(l)}+(1-x_{im})\log(1-\upsilon_{im}^{(l)})]\\
&\quad -\frac{1}{2}\sum_{i=1}^n\sum_{k=1}^K\sum_{d=1}^D\tau_{ik}(\log\sigma_{kd}^2+\frac{\hat{\sigma}_{id}^2}{\sigma_{kd}^2} +\frac{(\hat{\mu}_{id}-\mu_{kd})^2}{\sigma_{kd}^2})
\\&\quad  +\sum_{i=1}^n\sum_{k=1}^K\tau_{ik}\log\frac{\omega_k}{\tau_{ik}}+\frac{1}{2}\sum_{i=1}^n\sum_{d=1}^M(1+\log\hat{\sigma}_{id}^2).
\end{split}
\end{equation}
$L$ is sampling frequency for $\pmb{Z}$. Note, we assume $\pmb{X} \in \{0,1\}^{1\times M}$ here. It is easy to extend to $\pmb{X} \in \mathbb{R}^{n\times M}$ by using Gaussian distribution.

To minimize the $-\mathcal{L}(\pmb{A},\pmb{X})$, we will use the coordinate descent to optimize $\pmb{\tau}$, $\pmb{\Pi}$, $\pmb{\omega}$, $\pmb{\mu}$ and $\pmb{\sigma}$, and then use Adam to optimize the parameters of neural networks, i.e., $\theta$ and $\phi$.

In Eq.~(\ref{ELBO}), $\pmb{\hat{\mu}}_i$ and $\pmb{\hat{\sigma}}_i^2$ are computed by Eq.~(\ref{q-Z}). $\pmb{\upsilon}_{i}^{(l)}$ can be calculated by $\pmb{\upsilon}_i^{(l)}= f(\pmb{z}_i^{(l)};\theta)$, and $\pmb{z_i}^{(l)}$ is sampled by Eq.~(\ref{q-Z}). Using reparameterized trick \cite{kingma2013auto},  $\pmb{z}_i^{(l)} = \pmb{\hat{\mu}}_i + \pmb{\hat{\sigma}}_i \circ \pmb{\epsilon}^{(l)}$, where $\pmb{\epsilon}^{(l)}\sim \mathcal{N}(0,\textbf{1})$, and $\circ$ denotes Hadamard product.

We derived the update formulas of $\pmb{\tau}$, $\pmb{\Pi}$, $\pmb{\omega}$, $\pmb{\mu}$ and $\pmb{\sigma}$ as follows by making the derivative of $-\mathcal{L}(\pmb{A},\pmb{X})$ with respect to them equal to zero (see Appendix for more details), respectively.

\begin{equation}
\begin{split}\label{tau}
  \tau_{ik} \propto &\exp (\sum_{j}\sum_{l}\tau_{jl}[a_{ij}\log\pi_{kl}+(1-a_{ij})\log(1-\pi_{kl})]\\
& \quad-\frac{1}{2}\sum_{d}^D(\log\sigma_{kd}^2+\frac{\hat{\sigma}_{id}^2}{\sigma_{kd}^2} +\frac{(\hat{\mu}_{id}-\mu_{kd})^2}{\sigma_{kd}^2}) + \log\omega_k).
\end{split}
\end{equation}

\begin{equation}\label{omega}
  \omega_k = \frac{1}{n}\sum_{i}\tau_{ik},
\end{equation}
\begin{equation}\label{pi}
  \pi_{kl} = \frac{\sum_{ij}\tau_{ik}\tau_{jl}a_{ij}}{\sum_{ij}\tau_{ik}\tau_{jl}},
\end{equation}
\begin{equation}\label{mu}
 \mu_{kd}= \frac{\sum_i^n\tau_{ik}\hat{\mu}_{id}}{\sum_i^n\tau_{ik}},
\end{equation}
and
\begin{equation}\label{sigma}
 \sigma_{kd}= \frac{\sum_i^n\tau_{ik}(\hat{\sigma}_{id}+(\hat{\mu}_{id}-\mu_{kd})^2)}{\sum_i^n\tau_{ik}}.
\end{equation}

Finally, we summarize the learning algorithms in Algorithm \ref{algorithm:1}. According to Algorithm \ref{algorithm:1}, we analyse the time complexity of the process.  In each iteration, it takes $O(K^2n^2+KDn)$, $O(K^2n^2)$, and $O(Kn)$ to update $\pmb{\tau}$, $\pmb{\Pi}$, and $\pmb{\omega}$. Calculating $\pmb{\mu}$ and $\pmb{\sigma}$ takes $O(KDn)$. Thus, the total time complexity is $O(K^2n^2+KDn + W)$ per iteration, where $W$ is the scale of parameters of the neural networks.
\begin{algorithm}
    \caption{Learning algorithm for ANGM}\label{algorithm:1}
        \begin{algorithmic}[1]
            \renewcommand{\algorithmicrequire}{ \textbf{Input:}}
            \renewcommand{\algorithmicensure}{ \textbf{Output:}}
            \REQUIRE Adjacency and attribute matrices of the network: $\pmb{A}$, $\pmb{X}$;
            \ENSURE Node embeddings: $\pmb{Z}$;
            \STATE initialize $\pmb{\tau}, \pmb{\Pi}, \pmb{\omega}, \pmb{\mu}, \pmb{\sigma}$ and $\theta, \phi$;
           \REPEAT
            \FOR{ node $i=1$ to $n$ }
            \FOR{$k=1$ to $K$}
            \STATE update $\tau_{ik}$ according to Equation (\ref{tau});
            \ENDFOR
            \ENDFOR
            \FOR{$k=1$ to $K$}
            \STATE update $\omega_k$ according to Equation (\ref{omega});
            \FOR{$l=1$ to $K$}
            \STATE update $\pi_{kl}$ by using Equation (\ref{pi});
            \ENDFOR
            \FOR{$m=1$ to $D$}
            \STATE update $\mu_{kd}$ according to Equation (\ref{mu});
            \STATE update $\sigma_{kd}$ according to Equation (\ref{sigma});
            \ENDFOR
            \ENDFOR
            \FOR{node $i=1$ to $n$ } \STATE//forward-propagation
            \STATE calculate $\pmb{\hat{\mu}}_i$ and $\log\pmb{\hat{\sigma}}_i^2$ by $[\pmb{\hat{\mu}}_i,\log\pmb{\hat{\sigma}}_i^2]= g(\pmb{x}_i;\phi)$;
            \STATE sample $\pmb{\epsilon}$ according to $\pmb{\epsilon}^{(l)}\sim \mathcal{N}(0,\textbf{1})$;
            \STATE calculate node embedding $\pmb{z}_i$ by $\pmb{z}_i^{(l)} = \pmb{\hat{\mu}}_i + \pmb{\hat{\sigma}}_i \circ \pmb{\epsilon^{(l)}}$;
            \STATE calculate $\pmb{\upsilon_i}$ by $\pmb{\upsilon_i}= f(\pmb{z}_i;\theta)$;
            \ENDFOR
            \STATE calculate loss function by Equation (\ref{ELBO}) and update $\theta$ and $\phi$ by back-propagation;
            \UNTIL{convergence}
        \end{algorithmic}
    \end{algorithm}

\section{Experiments}\label{Experiments}
In this section, we first introduce the state-of-the-art approaches that are compared with ANGM method proposed in this study. Then, we test our method on node clustering and node classification tasks on both assortative and disassortative real-world networks. Finally, we visualize and cluster the learned embeddings on the synthetic networks, which are generated by ANGM, to show the performance of the methods. The code of ANGM is available in https://github.com/liuxyjlu/ANGM.
\subsection{Baselines}
Since our method is unsupervised, we compare our method with several unsupervised network representation learning methods, which fall into four categories: classical pure network embedding (Node2Vec \cite{node2vec-kdd2016}, NOBE \cite{jiang2018spectral}), matrix factorization based method (BANE \cite{yang2018binarized}), random walk based methods (ASNE \cite{liao2018attributed}, and ANRL \cite{ijcai2018-438}) and deep neural networks based methods (VGAE \cite{kipf2016variational}, ARVGE \cite{pan2018adversarially}, G2G \cite{bojchevski2018deep}, and GATE \cite{salehi2019graph}). They are different types of state-of-the-art methods for attributed network embedding.
The details of these methods are as follows.
\begin{itemize}
  \item[$\bullet$]\textbf{NOBE} \cite{jiang2018spectral} is a spectral embedding method based on the non-backtracking strategy to exploits nonlinear structure of graphs.
  \item[$\bullet$]\textbf{Node2Vec} \cite{node2vec-kdd2016} is a random-walk based method for learning node embeddings using only network topology.
  \item[$\bullet$]\textbf{BANE} is a \cite{yang2018binarized} is a matrix factorization model. It constructs Weisfeiler-Lehman proximity matrix that aggregated structural and attributed information and then factorize the matrix to learn the binarized embeddings.
  \item[$\bullet$] \textbf{ASNE} \cite{liao2018attributed} first learns the structure embeddings by performing Node2Vec, then feeds the structure embeddings and attributes to the deep neural networks to determine  the final embeddings.
  \item[$\bullet$] \textbf{ANRL} \cite{ijcai2018-438} is a neighbour-enhancement auto-encoder. It uses the random-walk to learn structural proximity and then adopts the attribute-aware skip-gram model to merge the topology and the attributes information.
  \item[$\bullet$] \textbf{VGAE} \cite{kipf2016variational} is a variational graph auto-encoder method. The network topology and attributes are mapped to vectors by GCN, and then the vectors are decoded into the networks using the inner product of embeddings.
  \item[$\bullet$] \textbf{ARVGE} \cite{pan2018adversarially} adds the adversarial model to VGAE to learn robust embeddings.
  \item[$\bullet$] \textbf{G2G} \cite{bojchevski2018deep} transforms the node attributes to the Gaussian distribution of the node embeddings, and leverages the a personalized ranking to constraint the similarity between two node's embeddings.
  \item[$\bullet$] \textbf{GATE} \cite{salehi2019graph} uses an attention machine to encode the node attributes to node representations and reversed the encoding process to generate the node attributes. Then, it utilizes the links to minimize the difference between the two linked nodes' embeddings.
\end{itemize}

\subsection{Node Clustering and Node Classification on Real-world Networks}
In this section, we test our method for node clustering and node classification tasks on the real-world networks.

\subsubsection{Real-World Networks}

We use eight real-world networks to test our proposed method as shown in Table \ref{tab:realnetwork}, where $n$, $m$, $K$, and $D$ are the numbers of nodes, edges, blocks, and attributes in the networks respectively, and \textit{Type} denotes the types of the networks.

\begin{table}[htpb]
\caption{Statistic features of six real-world networks}
\centering
\begin{tabular}{cccccc}
\toprule
Network    & $n$    & $m $ & $K$ &D & Type\\
\midrule
Corn.        &195     &304  & 5      & 1,703 & disassortative    \\
Texa.     &  187   & 328  &  5    &   1,703 & disassortative   \\
Wash.     & 230    &446 &    5   & 1,703  & disassortative    \\
Wisc.      &  265 & 530 &    5    &  1,703 & disassortative    \\
Cite.  & 3,312   &4,715 &   6    &   3,703 & assortative   \\
Actor & 7,600 & 33,544 & 4 & 931 & disassortative\\
Blog. & 5,196 & 17,143 & 6 & 8,189 & assortative\\
Flickr & 7,575 & 239,738 & 9 &  12,047 & assortative\\
\bottomrule
\end{tabular}\label{tab:realnetwork}
\end{table}

\textit{Cornell}, \textit{Texas}, \textit{Washington}, \textit{Wisconsin} (Corn., Texa., Wash., and Wisc., for short) are hypertext datasets from four universities \cite{craveny1998learning}. The nodes denote web pages. The connections indicate hyperlinks. Node labels are types of web pages, including student, staff, faculty, course, and research project. The attributes refer to the words in web pages. \textit{Citeseer} (Cite. for short) is an academic citation network \cite{namata2012query}. The nodes are academic papers. The edges represent citation relations. The labels denote the research areas of the papers, and attributes are words in the papers.
\textit{Actor} is a cooperation network \cite{ICLR2020GeomGCN}. The nodes refer to actors. An edge between two nodes means that they co-occur in the same Wikipedia web pages. The labels refer to categories of the actors in Wikipedia. The attributes are some words in the actors' Wikipedia web pages. \textit{BlogCatalog} (Blog. for short) and \textit{Flickr} are social networks \cite{li2015unsupervised}. In BlogCatalog, the nodes denote bloggers, and a link between two nodes means that one blogger follows the other one. The attributes represent the keywords in the blogs of the bloggers. Moreover, the labels are the interests of the bloggers. In Flickr, the nodes are the users, and edges are friendships between the users, the attributes denote the users' interests, and the labels refer to the groups that the users joined.

\begin{figure}[htbp]
\centering
\subfigure[Block matrix of Corn.]{
\includegraphics[width=0.4\columnwidth]{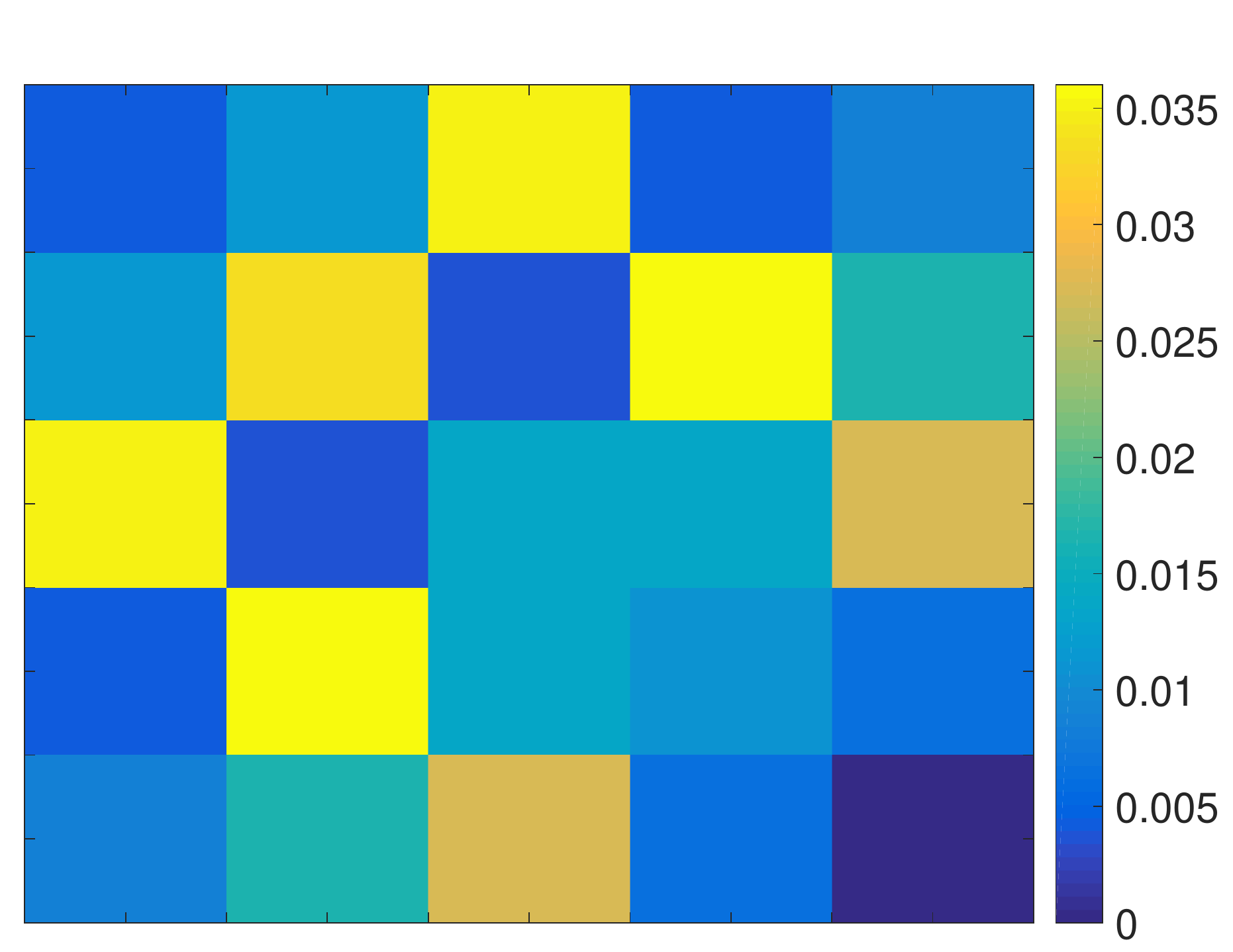}}
\subfigure[Block model of Corn.]{
\includegraphics[width=0.4\columnwidth]{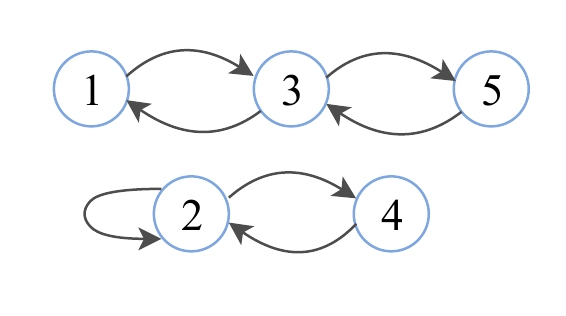}}
\subfigure[Block matrix of Cite.]{
\includegraphics[width=0.4\columnwidth]{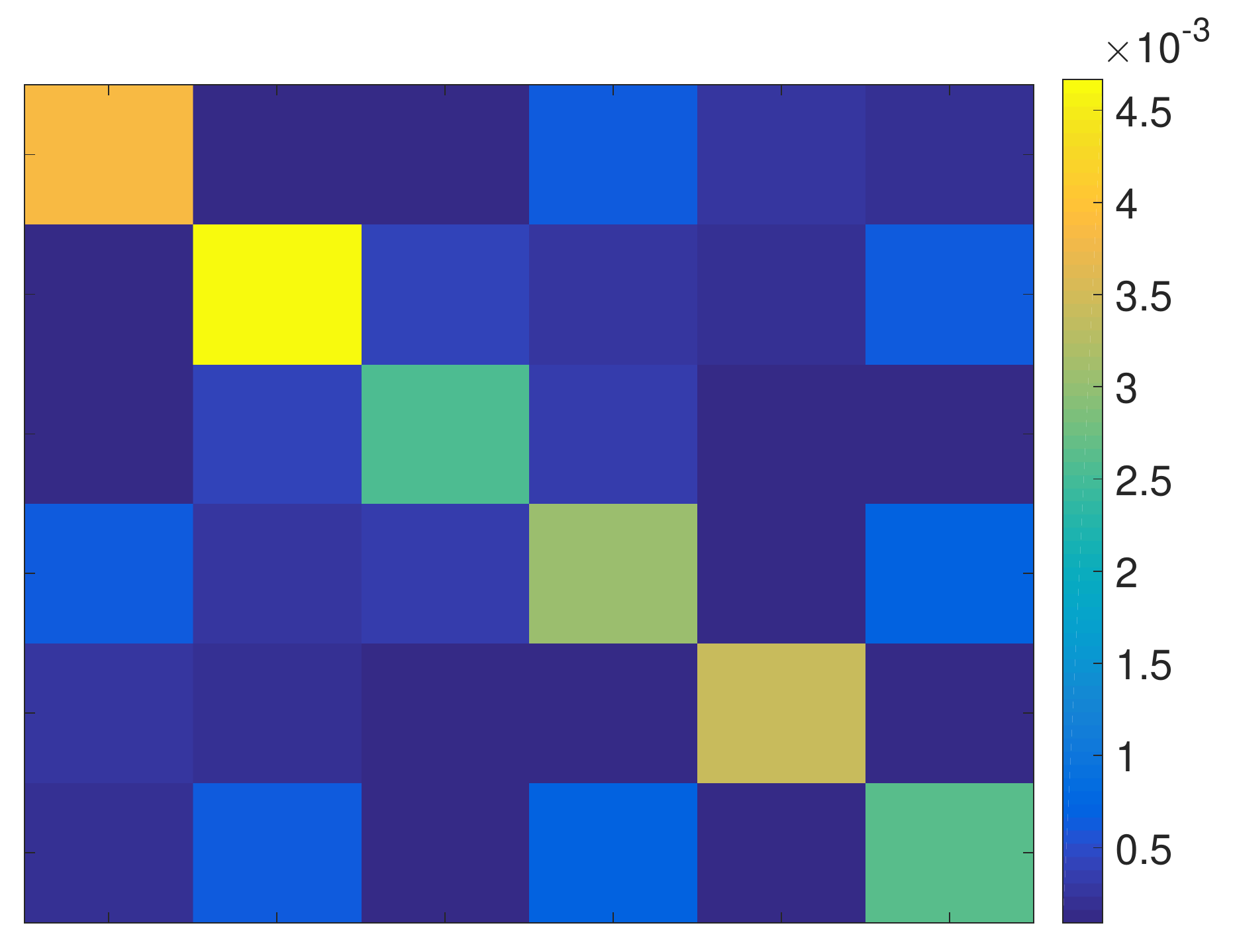}}
\subfigure[Block model of Cite.]{
\includegraphics[width=0.4\columnwidth]{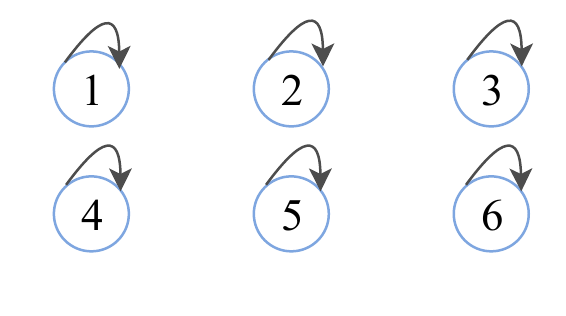}}
\caption{The block matrices and block models of Cornell and Citeseer. The color in (a) and (c) denotes the link density of nodes in each blocks. The circles in (b) and (d) denote the blocks and the arrows denote the higher link probabilities.}
\label{data-anlysis}
\end{figure}

First, we analyse the structural patterns contained in real-world networks and infer the types of the networks. According to the definition of the structural patterns \cite{yang2012characterizing}, we show the block-block link probability matrices and block models of two selected networks (i.e., Cornell and Citeseer) in Figure \ref{data-anlysis}.

Based on the ground truth and edges of the networks, the elements in the block matrices are calculated by $\pi_{kl} =\frac{e_{kl}}{s_{kl}}$, where $e_{kl}$ denotes the number of links between block $k$ and $l$ in the real-world network, and $s_{kl}$ represents the number of links between block $k$ and $l$ in a fully-linked network with the same ground truth and the same nodes as the real-world network. For communities, generally speaking, node $i$ is connected to node $j$ with higher probability if they belong to the same blocks. For multipartite structures, two nodes in different blocks are more likely to connect with each other. From Figure \ref{data-anlysis} (a) and (b), Cornell is a disassortative network containing a community (block 2) and three multipartite structures (blocks 1-3, blocks 3-5, and blocks 2-4). From Figure \ref{data-anlysis} (c) and (d), the structural patterns in Citeseer are all identified as communities, which means Citeseer is an assortative network. Thus, the structural patterns in Cornell are more complicated than those in Citeseer. Using the same approach, we can infer that Texas, Washington, Wisconsin, and Actor are also disassortative networks. BlogCatalog and Flickr are assortative networks. 

\subsubsection{Experiments Settings}
The experiments are run in two steps: 1) \textit{Learning step}: We use each embedding method to learn node representations, i.e., low-dimension and task-independent vectors or embeddings from the topology and attributes information of a network. 2) \textit{Evaluation step}: We evaluate the quality of the learned embeddings in machine learning-based network analysis tasks, including node clustering and node classification.

In the learning step, we use grid search to obtain the hyperparameters of our method and the details are as follows. For all of the real-world networks, each neural network contains two layers, the dimension of the node embddings is 20, and the optimizer is Adam. For Cornell, Texas, Washington, and Wisconsin, each layer consists of 32 hidden unites, the learning rate is 0.001, the number of iteration is 600. For Citeseer, each layer includes 32 hidden unites,  the learning rate is 0.005, the number of iteration is 1000. For Actor, BlogCatalog, and Flickr, each layer consists of 128 hidden unites and the learning rate is 0.01, the number of iteration is 2000.

In the evaluation step, we feed the learned node embeddings to the Gaussian mixture model (GMM) for node clustering. Then, we choose the normalized mutual information (NMI) \cite{kuncheva2004using} and accuracy (AC) \cite{xu2003document} to evaluate the quality of the node embeddings by their results on GMM. For node classification, we first fix the ratio of test set to 20\% and increase the ratio of training set from 10\% to 80\% by a step of 10\%. Then, we train an SVM using the labels and the learned embedding of the nodes in the training set. Next, we use the trained SVM to predict the labels of nodes in the testing set. Finally, we choose Macro-F1 and Micro-F1 \cite{pillai2012f} to evaluate the quality of the learned node embeddings through the performances of the SVM. For each setting, we randomly sample nodes ten times and show the average Macro-F1 and Micro-F1.

For all baselines, we use the implementation released by the original authors and retain to the hyperparameter settings in their implementation except for the dimension of the node embeddings, which is set to 20 to be same as our method for fairness. We use NOBE-GU and ANRL-WAN for NOBE and ANRL, which perform best among their variants.

\begin{table}[t]
\caption{NMI (\%) and AC (\%) of the methods and improvement ratio (\%) on node clustering for eight real-world networks}
\centering
\resizebox{\columnwidth}{!}{
\begin{tabular}{cccccccccc}
\toprule
Metric &Method    &  Corn. & Texa.  & Wash. &Wisc. & Cite.  & Actor & Blog. & Flickr\\
\midrule
\multirow{11}{*}{NMI}
& NOBE & 3.29 & 4.87 & 2.97 & 6.46 & 7.69 & 0.12 & 1.79 & 1.72\\
&Node2Vec  &  7.60  & 5.57  & 3.67  & 2.52  & 17.03 & 0.09& 20.26 & 17.94\\ 
& BANE & 12.57 & 17.67 & 16.82& 20.00 &  9.24 & 0.44 & 3.38 & 1.83\\
&ASNE     & 8.57  & 16.23 & 22.77    & 19.10   & 15.72  & \textbf{4.04} & 5.45 & 7.60\\
&ANRL     & 12.98 & 15.28 & 16.56    & 10.46   & \textbf{35.43}  & 1.00 & 4.31 & 5.70 \\
&VGAE     & 7.31  & 5.53  & 10.33    & 8.23    & 17.80  & 0.62 & 11.57 & 15.40 \\
&ARVGE    & 10.80 & 12.64 & 10.80    & 8.18    & 19.71  & 1.69 & 23.83 & 12.89\\
& G2G & 9.25 & 5.00 & 4.89 & 9.39 & 33.54 & 0.90 & 14.45 & 5.45\\
& GATE& 8.10 & 10.69 & 10.94 & 8.73 & 35.09 & 0.49 & 22.12 & 10.74\\
&ANGM     & \textbf{29.09} & \textbf{28.64} & \textbf{34.73}    & \textbf{40.10}   & 29.04
  & 2.45 & \textbf{23.87} & \textbf{21.49}\\
& Improvement & 124.11 & 62.08 & 52.53 & 100.50 & -18.03 & -39.36& 0.17&19.79\\
\midrule
\multirow{11}{*}{AC}
& NOBE & 33.85 & \textbf{50.81} & 37.33 & 38.93 & 27.14 & 25.82 & 17.86 &  11.80\\
&Node2Vec  & 41.54 & 32.62 & 47.39 & 40.75 & 41.49 & 24.05 & 36.86 & 32.75\\
& BANE& 38.05 & 45.56 & 42.91 & 43.62 & 33.34 & 23.96 &
22.92 & 15.76\\
&ASNE     & 40.51 & 39.04 & 43.91 & 43.77 & 42.91 & \textbf{28.92} & 26.35 & 20.86\\
&ANRL     & 36.92 & 42.62 & 45.43    & 38.57   & 51.76  & 23.33 & 27.12 & 17.54\\
&VGAE     & 34.62 & 34.06 & 39.17    & 30.30   & 37.00  & 22.62 & 31.32 & 30.07\\
&ARVGE    & 31.90 & 42.73 & 37.87    & 35.25   & 42.65  & 22.79 & 40.57 & 24.30\\
& G2G & 35.39 & 35.83 & 32.61 & 32.45 & \textbf{55.34} & 23.97 & 33.74 & 20.07\\
&GATE& 37.95 & 48.13 & 41.30 & 34.72 & 51.54 & 21.67& 38.11 & 22.24\\
&ANGM     & \textbf{44.21} & 50.59 & \textbf{54.35}    & \textbf{56.60}   & 54.98
  & 25.14 & \textbf{43.79} & \textbf{33.74}\\
& Improvement & 6.43 & -0.43 & 14.69 & 29.31 & -0.65 & -13.07& 7.94&3.02\\
\bottomrule
\end{tabular}}\label{cluster:NMI-AC}
\end{table}
\subsubsection{Experimental Results}
\begin{figure}
    \centering
    \includegraphics[width=\textwidth]{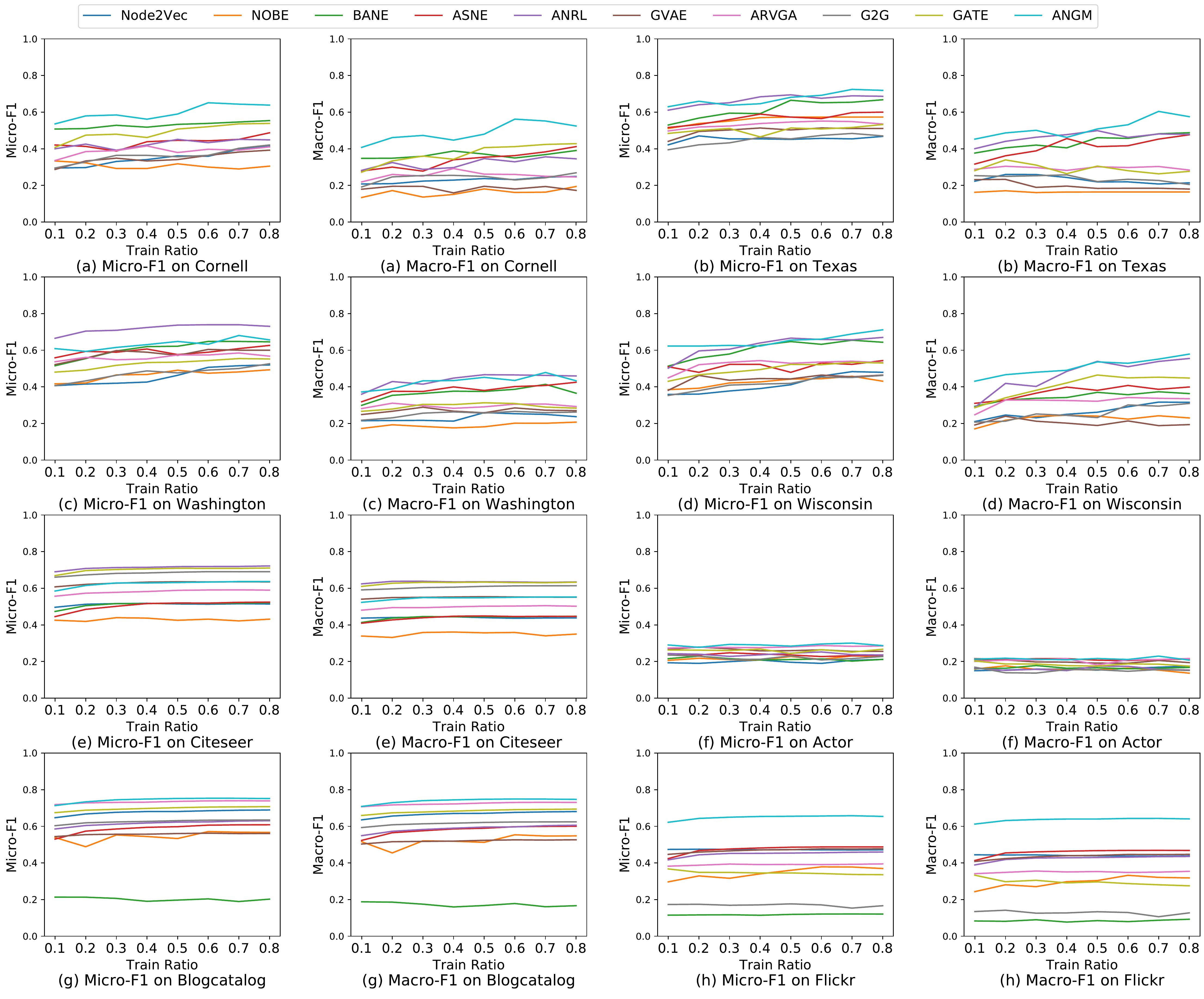}
    \caption{Micro-F1 and Macro-F1 of the methods on the node classification for six real-world networks}
    \label{fig:classification}
\end{figure}
Table \ref{cluster:NMI-AC} and Figure \ref{fig:classification} show the results of our method and the compared algorithms for node clustering and node classification tasks. Specially, ``Improvement'' means the ratio of the improvement of ANGM over the best performed baseline method in Table \ref{cluster:NMI-AC}. For example, the ``Improvement'' of NMI on Cornell dataset can be calculated by $\frac{29.09\%-12.98\%}{12.98\%} = 124.11\%$, where $29.09\%$ and $12.98\%$ are the NMIs of our method (ANGM) and the best performing baseline method (ANRL), respectively. 

Among the eight networks, ANGM outperforms all baselines on six and five networks under the NMI and AC metrics, respectively, as shown in Table \ref{cluster:NMI-AC}.  ANGM improves the NMI score more than $15\%$ on Cornell dataset compared ANRL. In the cases when considering AC, ANGM increases more than $10\%$ compared with BANE on Wisconsin dataset and the ratio of the improvement is $29.31\%$.  On Citeseer with communities, ANGM is in the second place under AC, and its AC score is only 0.4\% less than that of G2G, which is the best performing baseline method. On another two assortative networks, i.e., BlogCatalog and Flickr, ANGM outperforms the baselines. This because the community in Citeseer is more obvious than that in BlogCatalog and Flickr. Therefore, the baselines designed for assortative networks are superior to ANGM on Citeseer but perform worse on the other two. On Actor,
all algorithms perform poorly because there is little difference in the link probability between the different blocks. For the datasets for which  ANGM performs best, especially on disassortative networks, the improvement over the second best performing method is significant as in majority of cases we notice more than $50\%$ and $5\%$ improvement ratio with respect to NMI and AC, respectively.
From Figure \ref{fig:classification} we can see that Macro-F1 and Micro-F1 maintain steady growth for all algorithms on small-scale networks (Cornell, Texas, Washington and Wisconsin). On Citeseer, Actor, BlogCatalog, and Flickr, the Macro-F1 and Micro-F1 are stable with the increasing of training ratio, because the small ratios of nodes are enough to train a SVM on these four networks. Our method (ANGM) outperforms the baselines on most of the networks, especially on Cornell and Flickr dataset, with more than $40\%$ (Macro-F1) and more than $50\%$ (Micro-F1) in each training ratio on Cornell, which are at least $10\%$ more than those of the best performing baseline method (ASNE), respectively.  
From Table \ref{cluster:NMI-AC} and Figure \ref{fig:classification}, we can conclude that ANGM performs better on disassortative networks with complicated structural patterns (Cornell, Taxes, Washington, Wisconsin, and Actor datasets), which is the main goal of our research. ANGM results are also very good and although, not the best, are comparable with other state-of-the-art algorithms on Citesser (an assortative network). Lower ANGM's performance on assortative networks than disassortative ones, when compared with other methods, is down to the fact that our proposed method uses $\pmb{\Pi}$ to fit networks with different structures and other baseline embedding methods are designed for assortative networks.

\subsection{Visualization of Representations on Synthetic Networks}
In this section, we will test and visualize the performance of ANGM and the baselines on different kinds of synthetic networks, including assortative networks with communities  and disassortative networks with multipartite structures, hubs, and hybrid structures. We first show how to generate synthetic networks step by step, and then we run the algorithms on these networks and show their results.
\subsubsection{Generation Model for Synthetic Networks}
To test and visualize the performance of our method on networks with different structures, we generate four types of attributed networks, which are networks with communities, multipartite structures, hubs, and hybrid structures, respectively.

We denote the model for generating networks as $(n, K, \pmb{\omega}, \pmb{\Pi}, \pmb{\upsilon})$, which can be regarded as a simplified version of our ANGM omitting the neural networks or an extension of standard SBM adding node attributes.  $n$ and $K$ are the numbers of nodes and blocks, respectively; $\pmb{\omega}, \pmb{\Pi}, \pmb{\upsilon}$ have the same meaning as they are in Section \ref{model}. The generation process are as follows: (a) Assign each  node $i$ to one of $K$ blocks according to a multinomial distribution: $c_i\sim mul(\pmb{\omega})$; (b) Generate node attributes $\pmb{x}_i$ according to a Bernoulli distribution: $\pmb{x}_i\sim Ber(\pmb{\upsilon}_i)$; (c) For each pair of nodes ($i,j$), determine if there is a link between them according to a Bernoulli distribution: $a_{ij}\sim Ber(\pi_{c_ic_j})$.

\begin{figure}[htbp]
\centering
\subfigure[]{
\includegraphics[width=0.171\columnwidth]{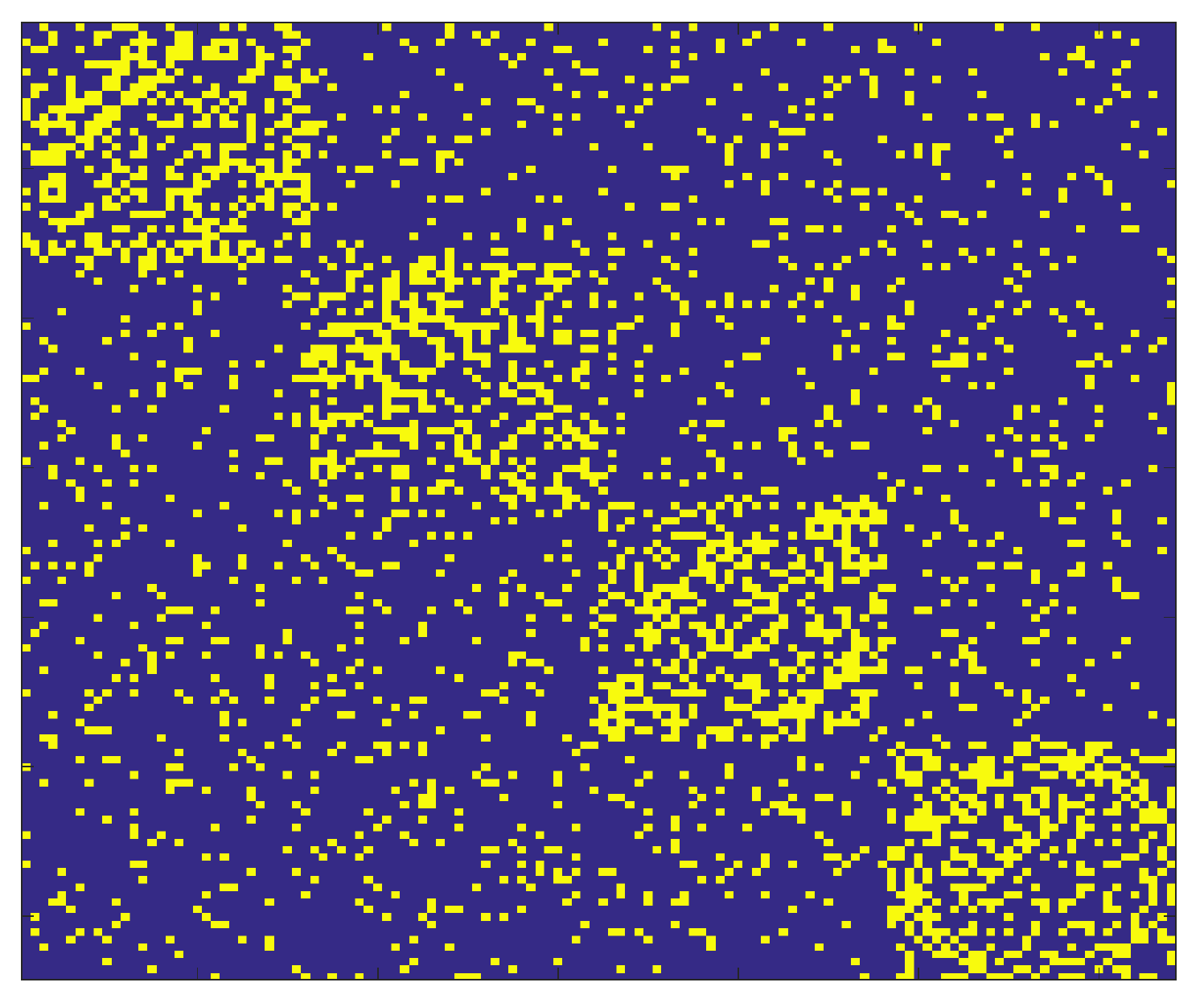}}
\subfigure[]{
\includegraphics[width=0.171\columnwidth]{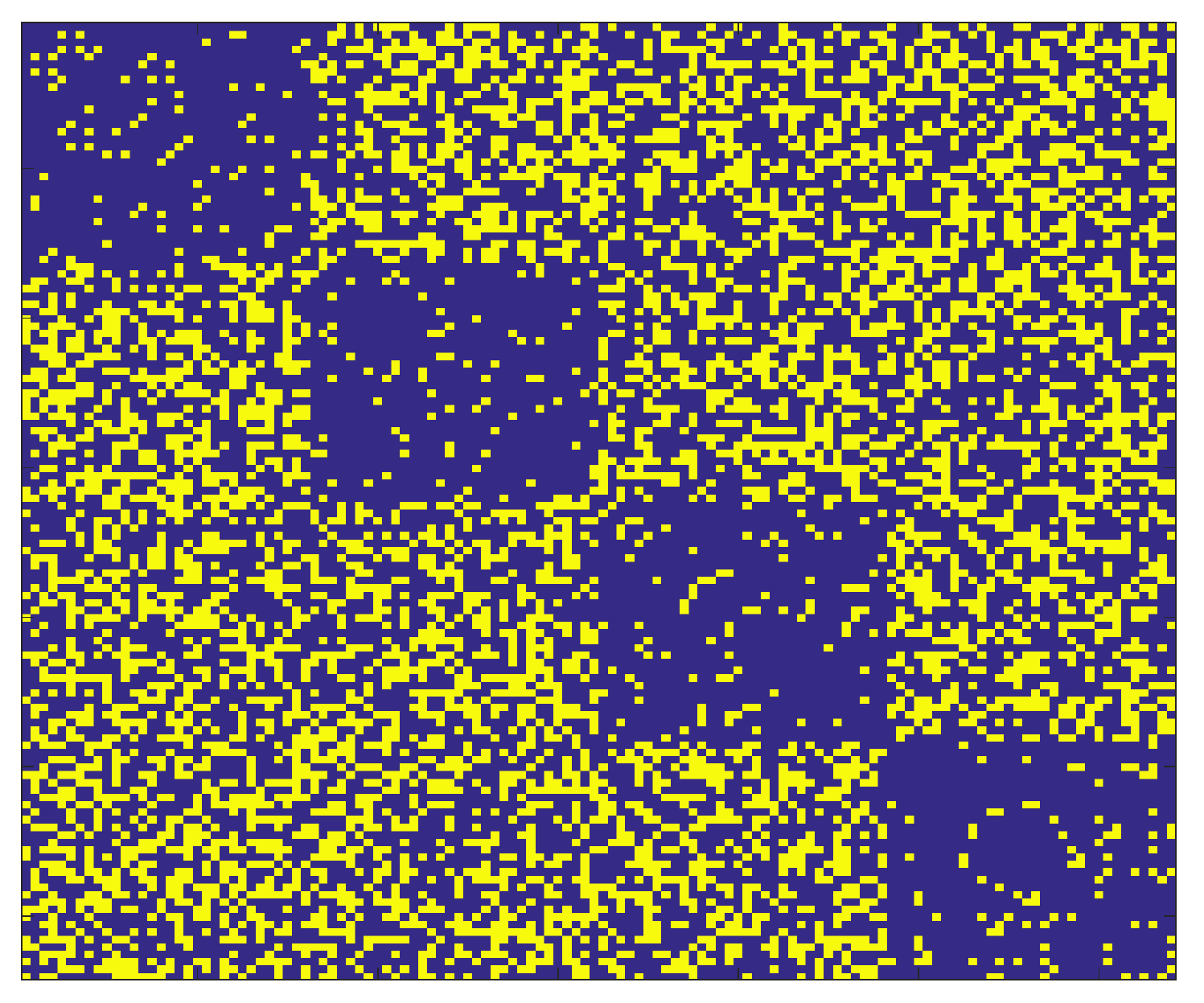}}
\subfigure[]{
\includegraphics[width=0.171\columnwidth]{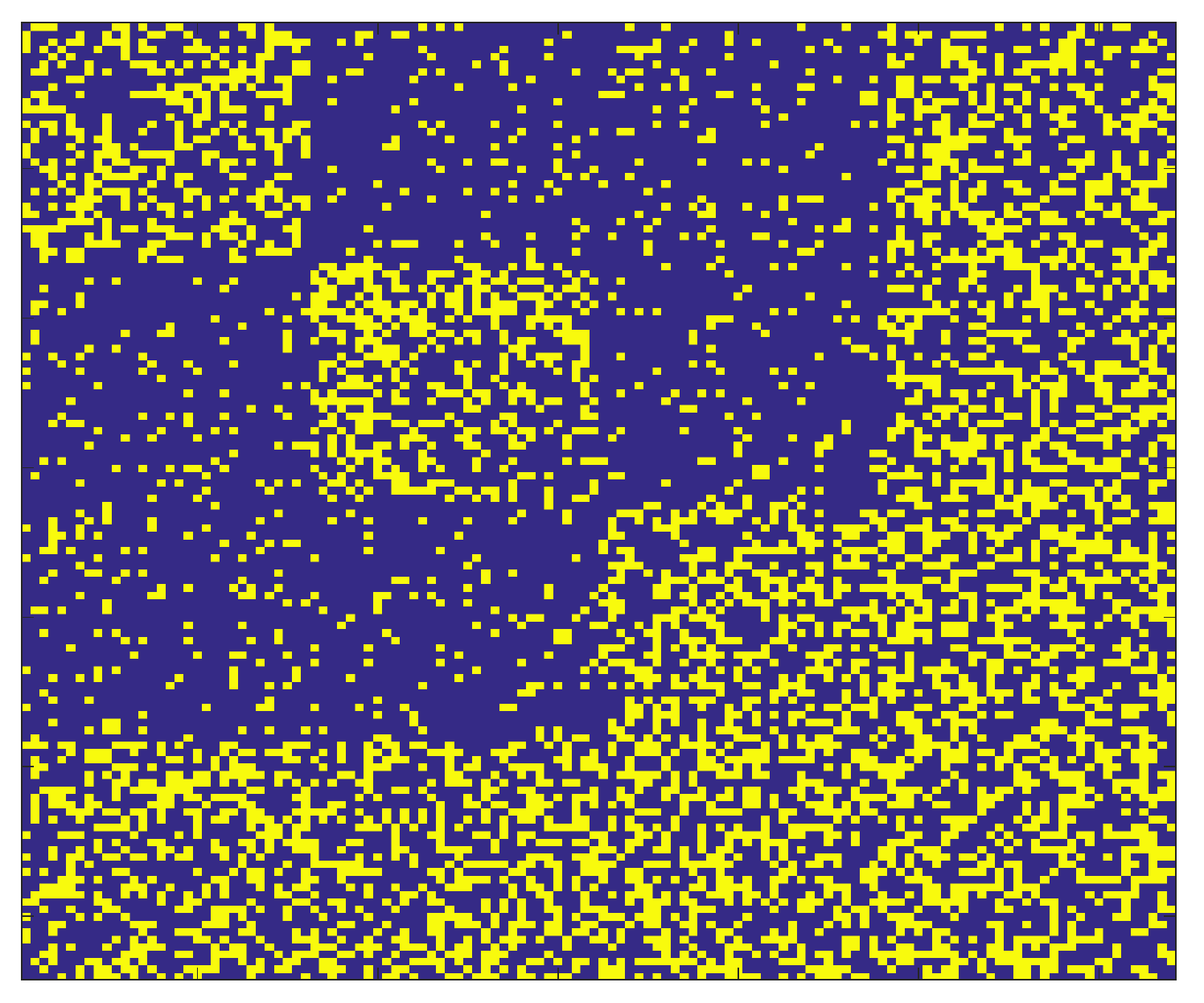}}
\subfigure[]{
\includegraphics[width=0.171\columnwidth]{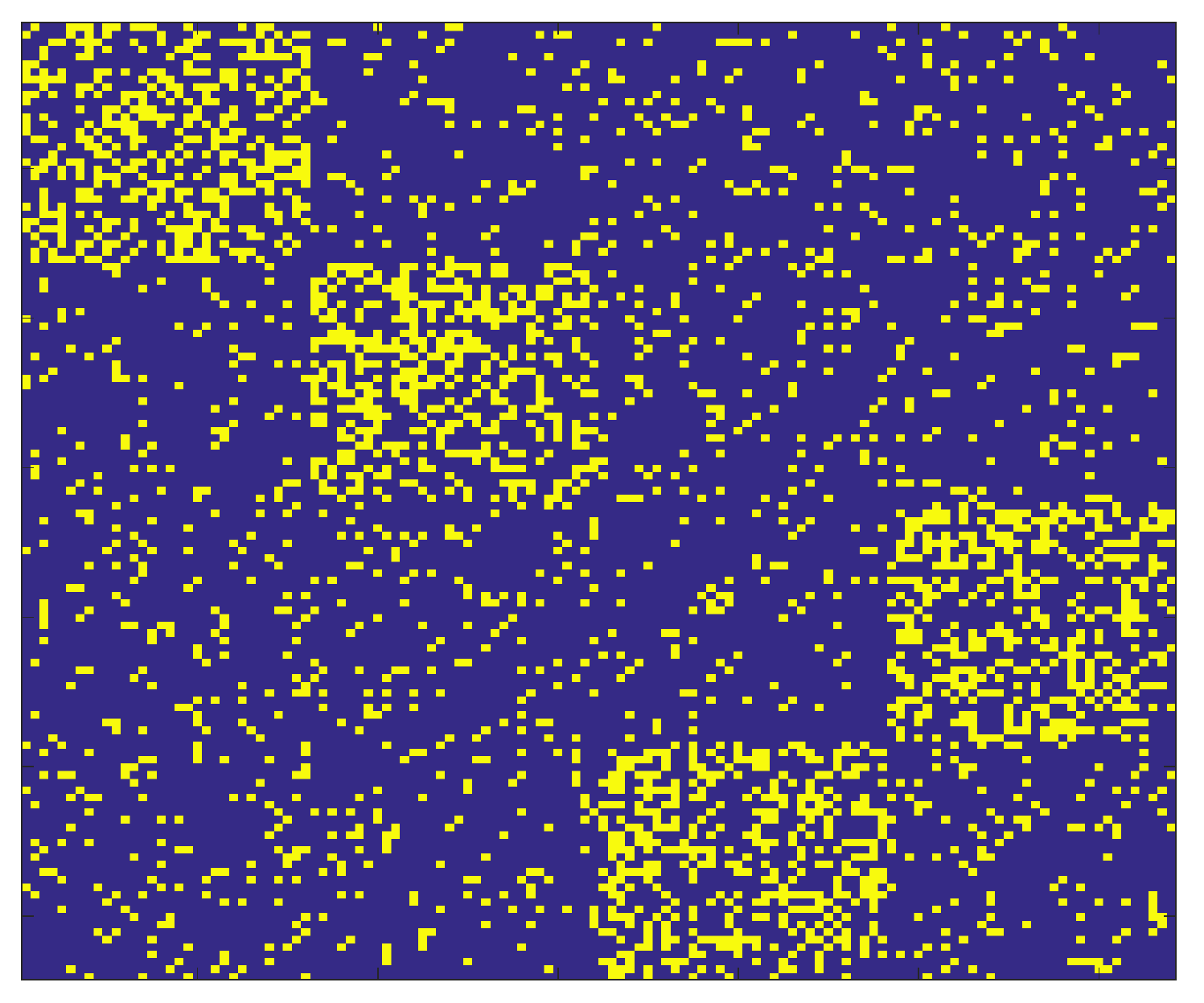}}
\subfigure[]{
\includegraphics[width=0.225\columnwidth]{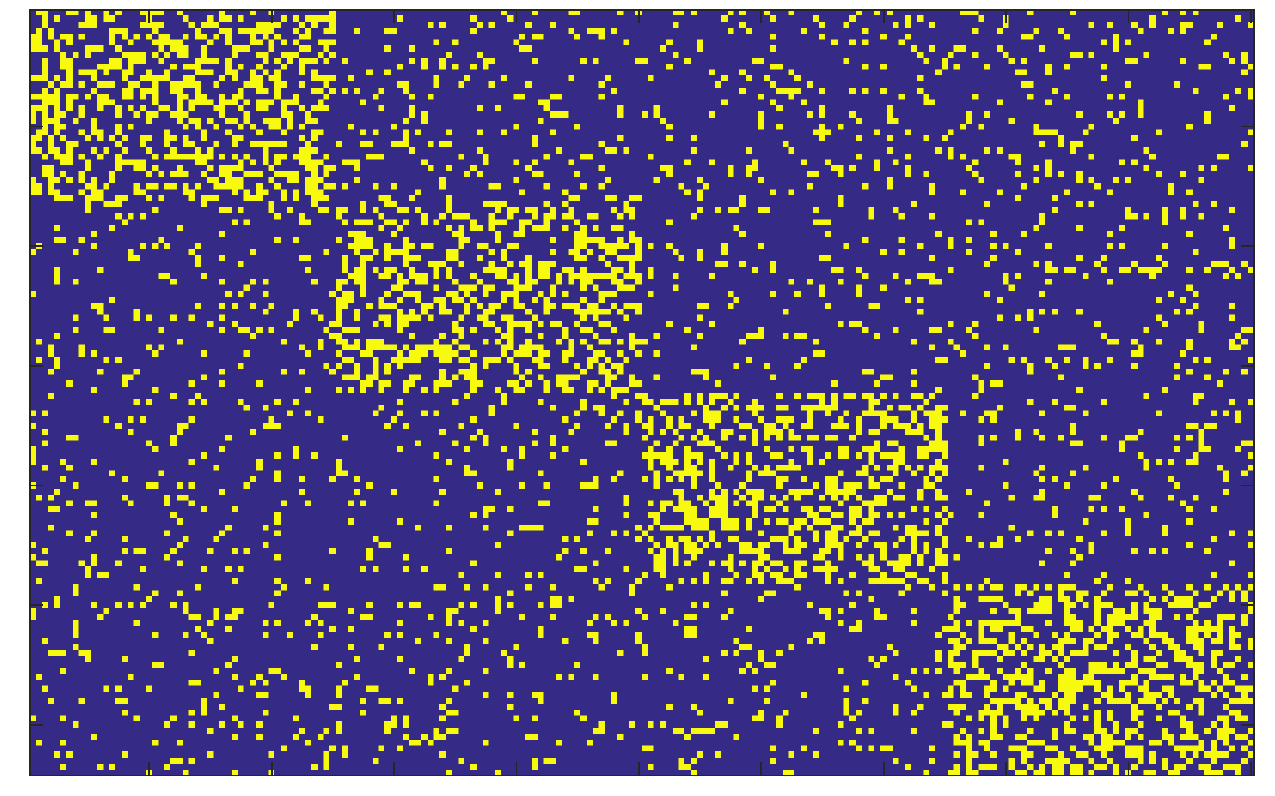}}
\caption{The adjacency and attribute  matrices of attributed networks. (a)-(d) are the adjacency matrices of networks with 4 types of structural patterns: (a) communities; (b) multipartite structures; (c) hubs; and (d) hybrid structures. (e) is the attribute matrix of one of the networks.}
\label{adjmat}
\end{figure}

Then, we give some details for setting the parameters to generate different types of networks. (a) To generate node attributes, we assume that the nodes in the same block have similar attributes by setting $\pmb{\upsilon}$ as follows. If node $i$ belongs to block $k$, we assume that the elements of the $n\times (K\times h)$-dimension matrix $\upsilon$ are  set as $\upsilon_{id} = p_{a_1}$ if $d\in\{(k-1)\times h+1,(k-1)\times h+2, ...,k\times h\}$, otherwise $\upsilon_{id} = p_{a_2}$.  (b) For structural topology, we generate four types of networks as follows. We denote the indices of blocks as $k,l \in\{1,2,...,K\}$. For networks with communities, we set $\pi_{kl} = p_{s_1}$ if $k=l$, otherwise $\pi_{kl} = p_{s_2}$. For networks with multipartite structures, we set $\pi_{kl} = p_{s_2}$ if $k=l$, otherwise $\pi_{kl} = p_{s_1}$. For networks with hubs, we set $\pi_{kl} = p_{s_1}$ if $k=l$ or $k=K$ or $l=K$, otherwise $\pi_{kl} = p_{s_2}$. For networks with hybrid structures containing $k_1$ communities and $k_2$ multipartite networks ($k_1+k_2=K$), we set $\pmb{\Pi}$ as followings:
\begin{equation}
\begin{split}
\pmb{\Pi}=
\begin{bmatrix}
p_{s_1}  &       & p_{s_2}&\cdot &   &   & \\
      &\ddots &        &\cdot &   & p_{s_2}& \\
 p_{s_2} &    &p_{s_1}     &\cdot &   &   &  \\
\cdot &\cdot  &\cdot&\cdot&\cdot &\cdot&\cdot   \\
      &       &   &\cdot  &p_{s_2}&   & p_{s_1}\\
      & p_{s_2}    &   &\cdot  &   &\ddots& \\
      &       &   &\cdot  & p_{s_1}&   &p_{s_2}\\
\end{bmatrix}\notag
\begin{matrix}
 \Bigg\}k_1 \\
   \\
   \\
 \Bigg\}k_2\end{matrix}
\end{split}
\end{equation}
Here, we set $n = 128$, $K = 4$, $k_1 = k_2 = 2$, $\pmb{\omega} = (\frac{1}{4},\frac{1}{4},\frac{1}{4},\frac{1}{4})$, $h = 50$, $p_{s_1}=p_{a_1}=0.4$, and $p_{s_2}=p_{a_2} = 0.1$. Figure \ref{adjmat} shows the adjacency and attribute matrices of the generated networks.

We test the proposed method and the baselines on these generated attributed networks and show the results on Section \ref{Synth-results}.

\begin{figure}[htbp]
\centering
\subfigure[NOBE]{
\includegraphics[width=0.235\textwidth]{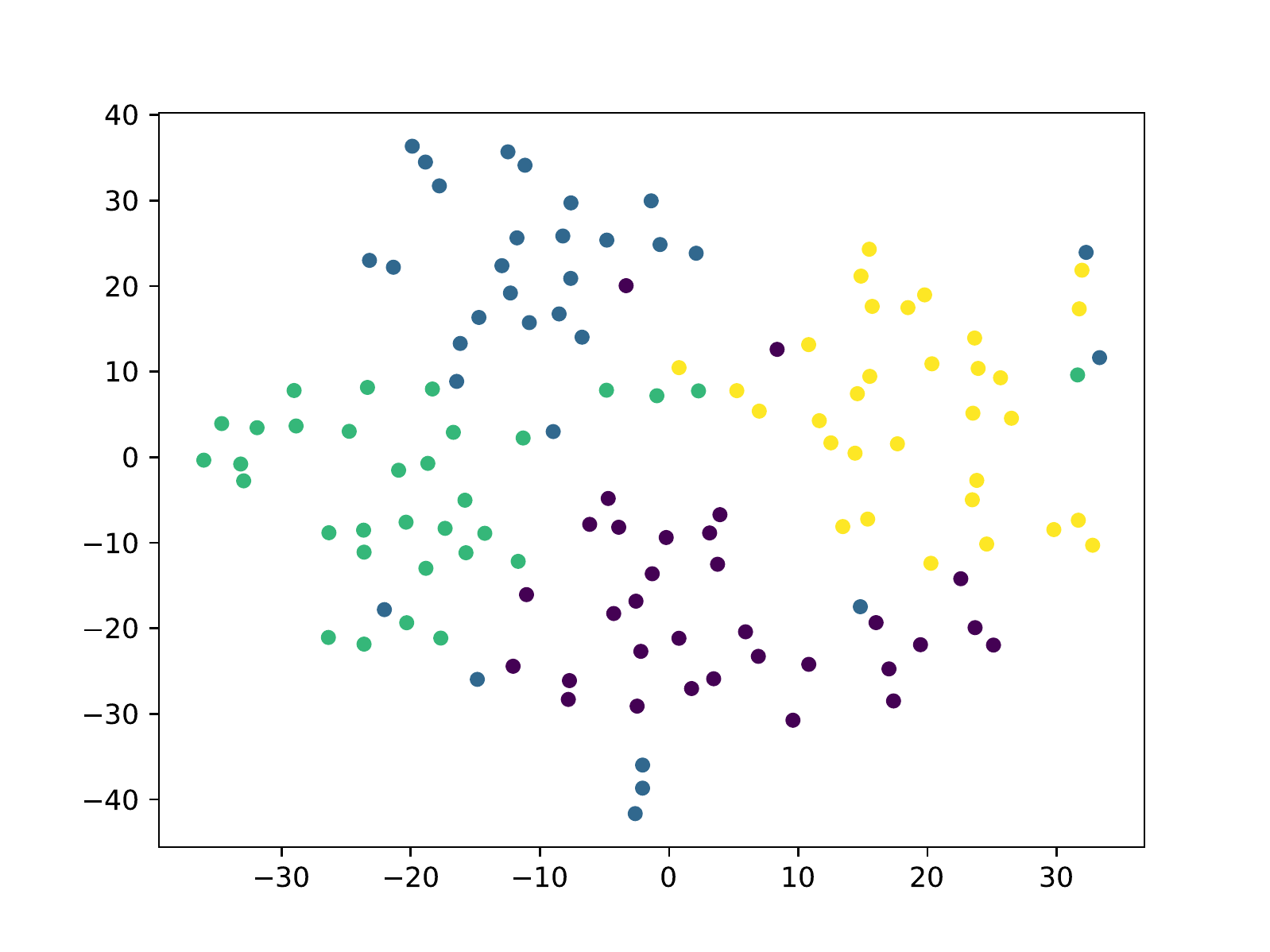}}
\subfigure[Node2Vec]{
\includegraphics[width=0.235\textwidth]{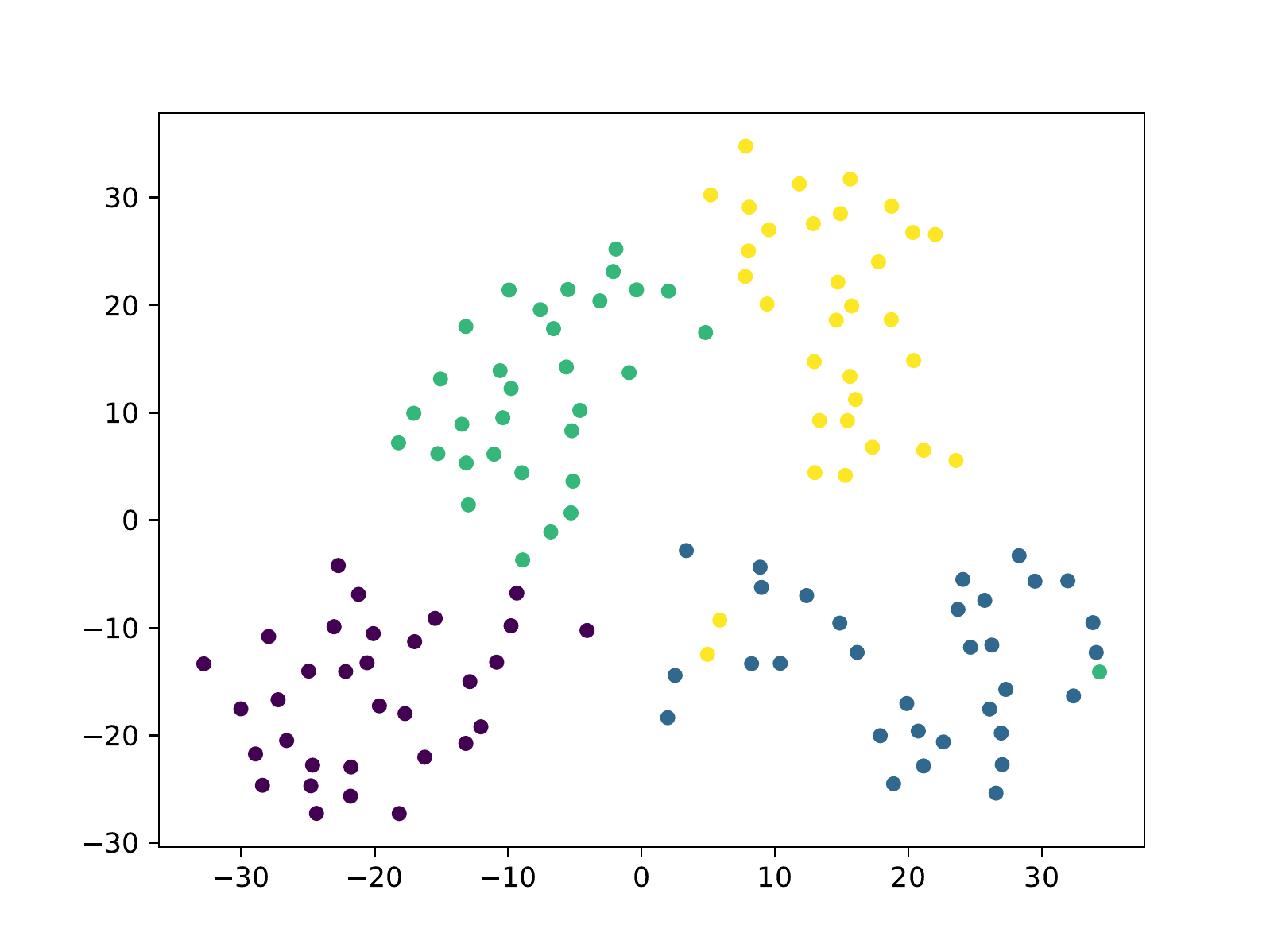}}
\subfigure[BANE]{
\includegraphics[width=0.235\textwidth]{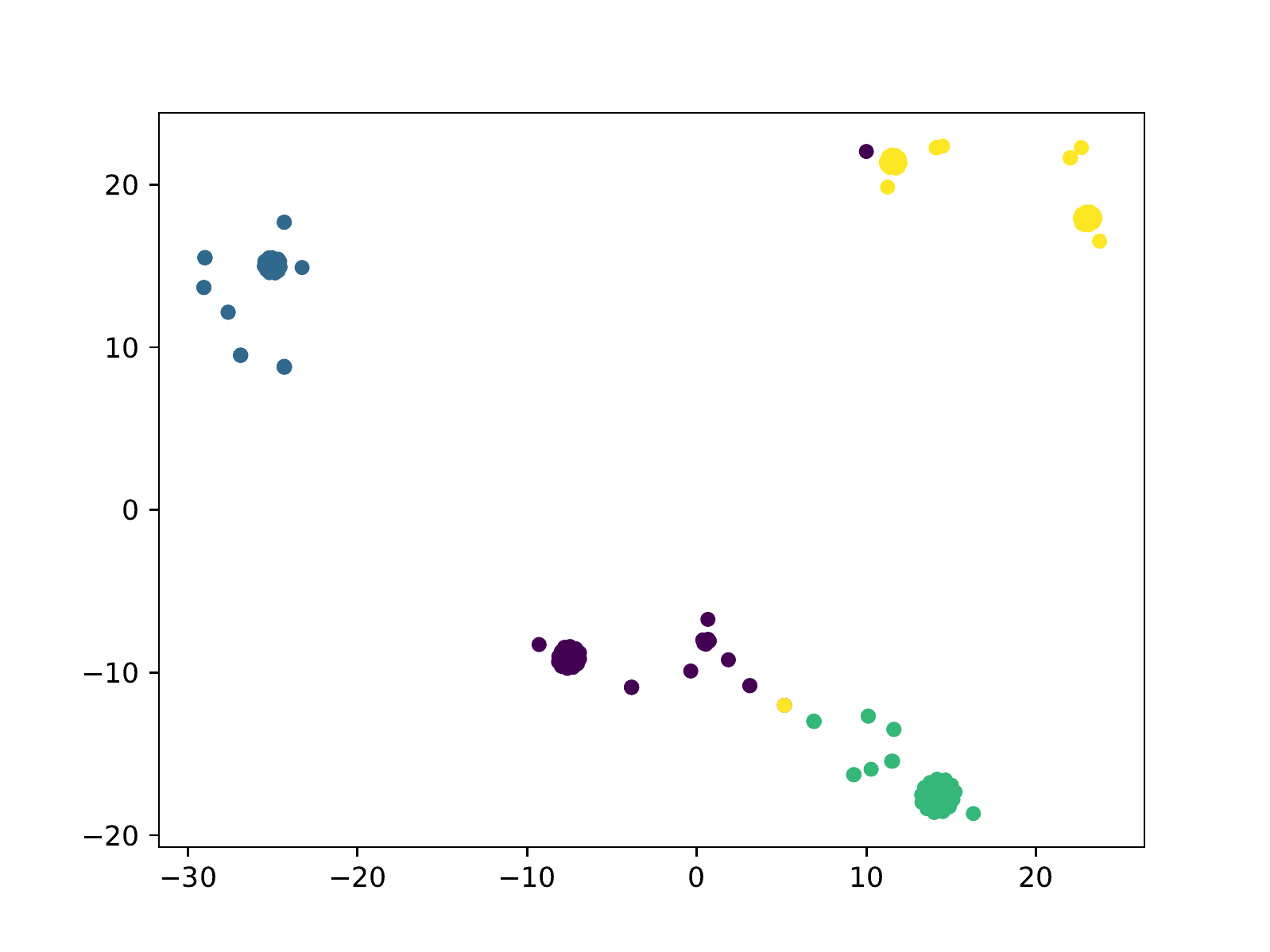}}
\subfigure[ASNE]{
\includegraphics[width=0.235\textwidth]{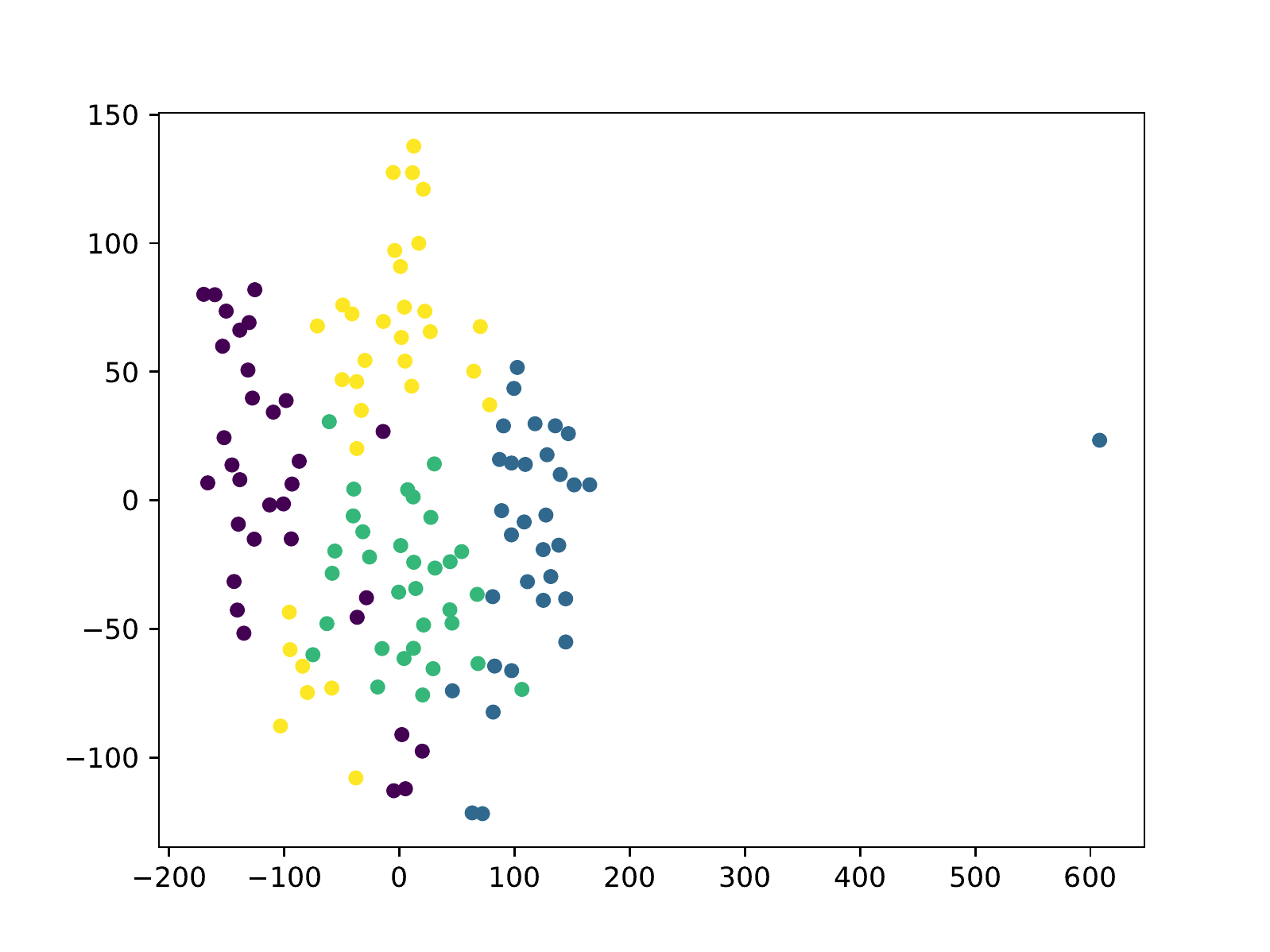}}
\subfigure[ANRL]{
\includegraphics[width=0.235\textwidth]{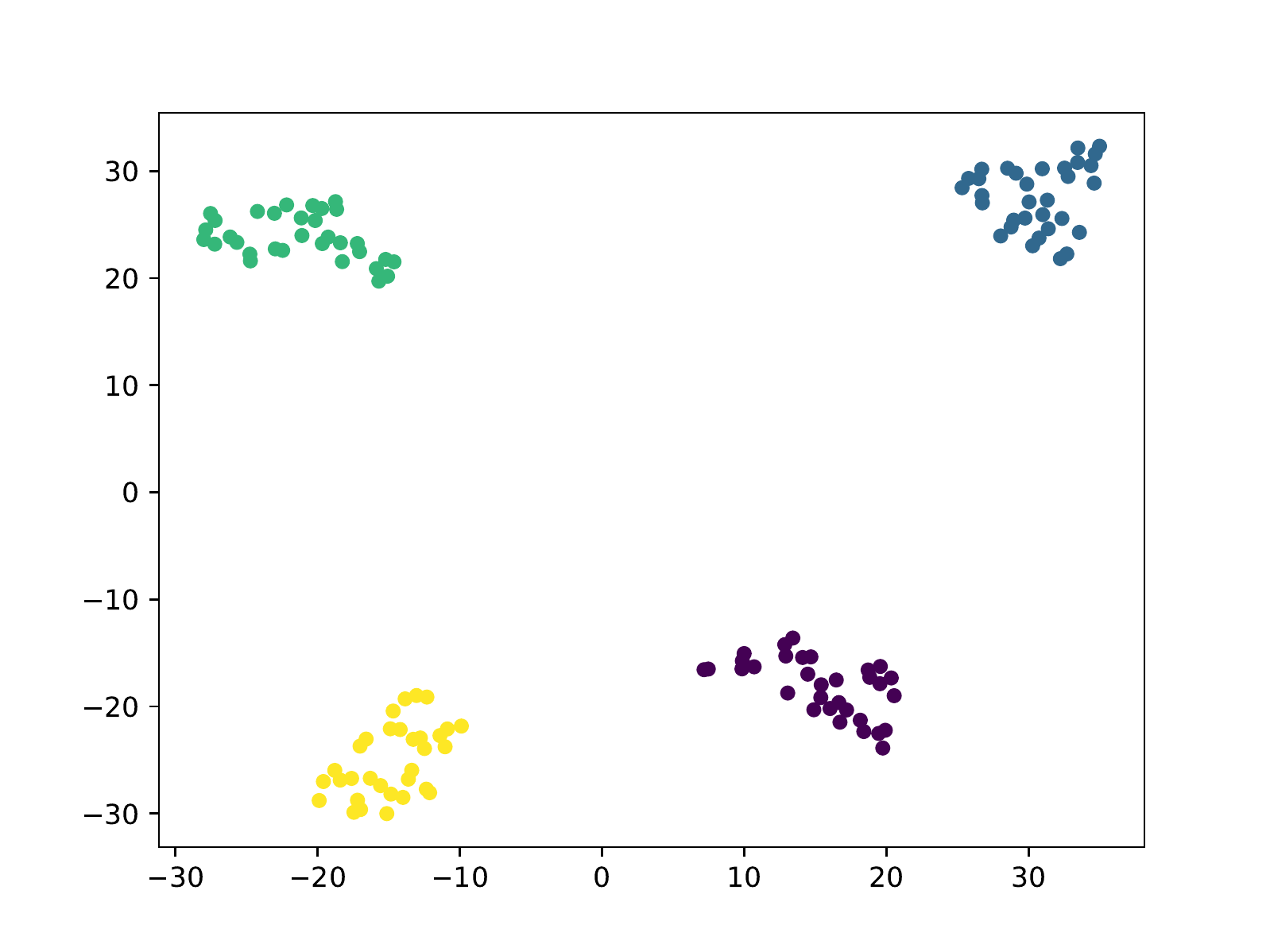}}
\subfigure[VGAE]{
\includegraphics[width=0.235\textwidth]{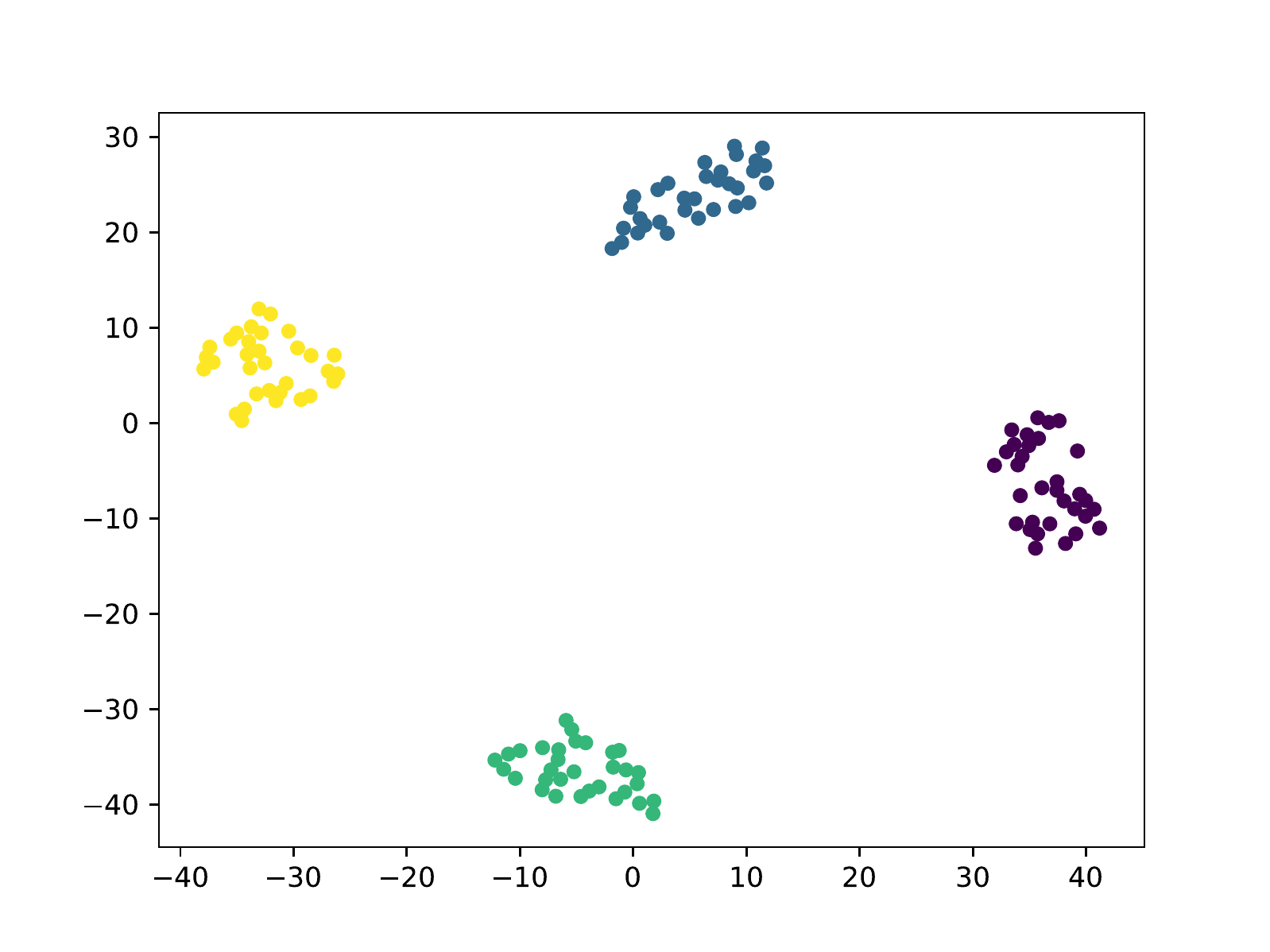}}
\subfigure[ARVGE]{
\includegraphics[width=0.235\textwidth]{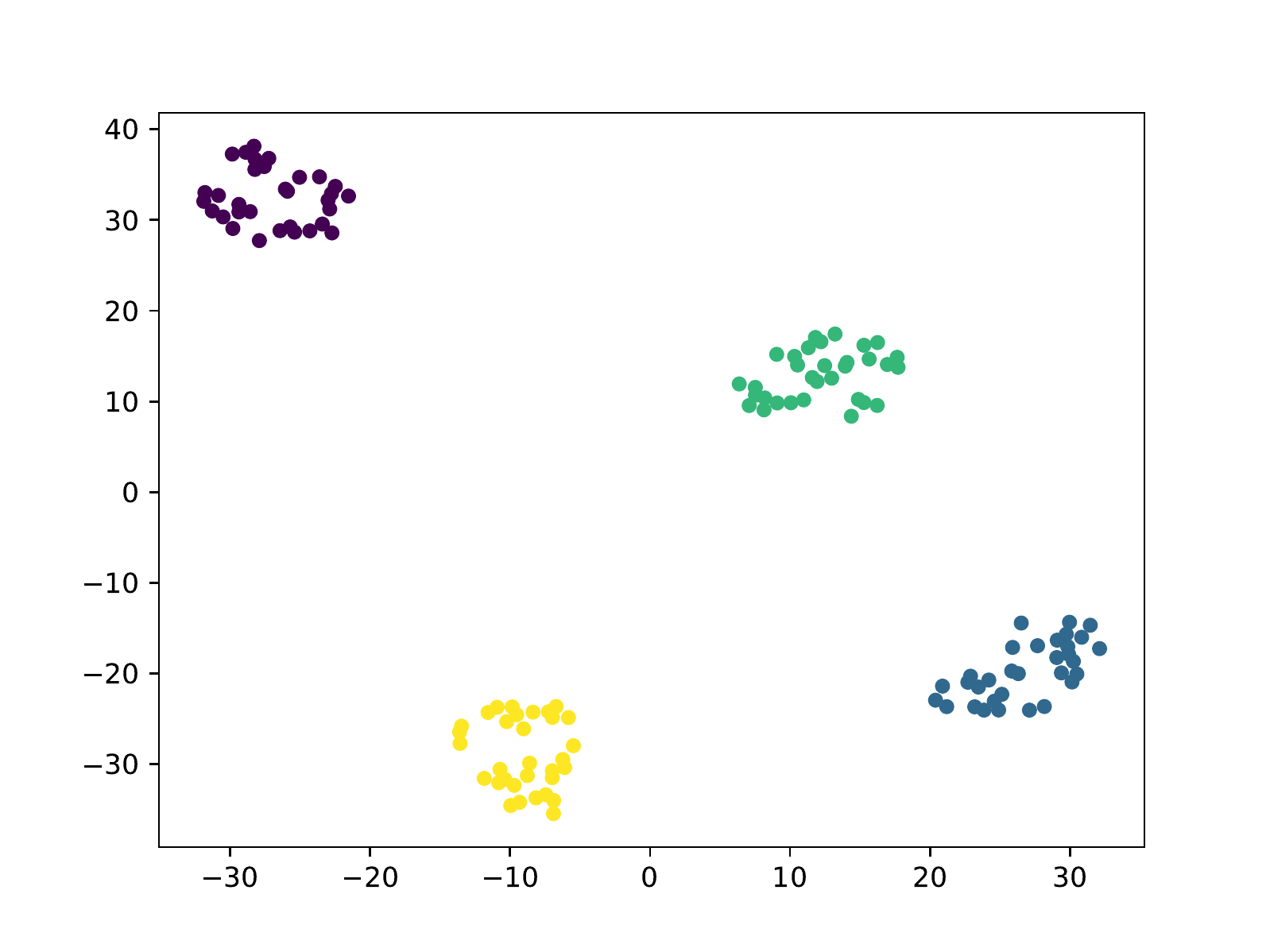}}
\subfigure[G2G]{
\includegraphics[width=0.235\textwidth]{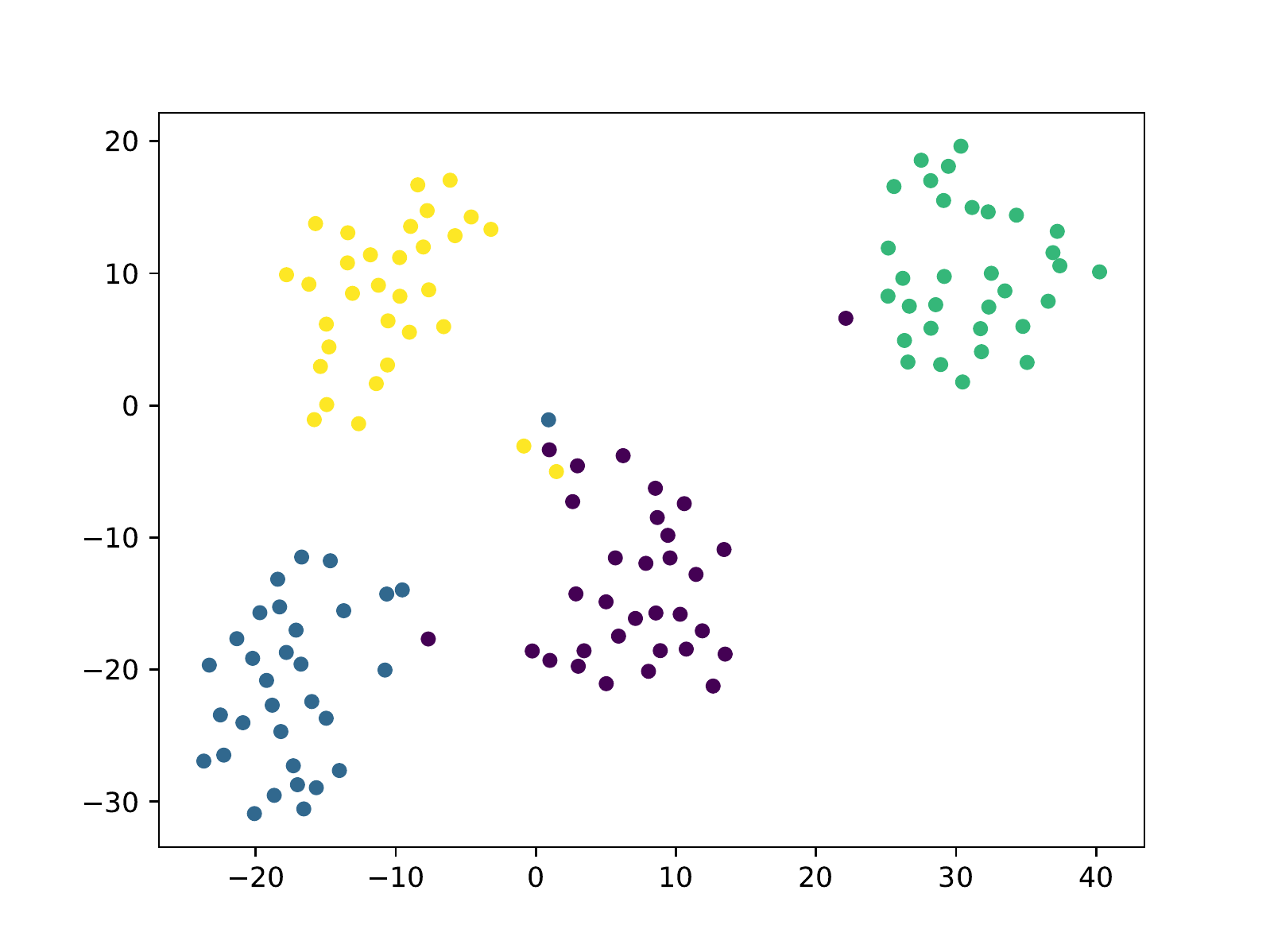}}
\subfigure[GATE]{
\includegraphics[width=0.235\textwidth]{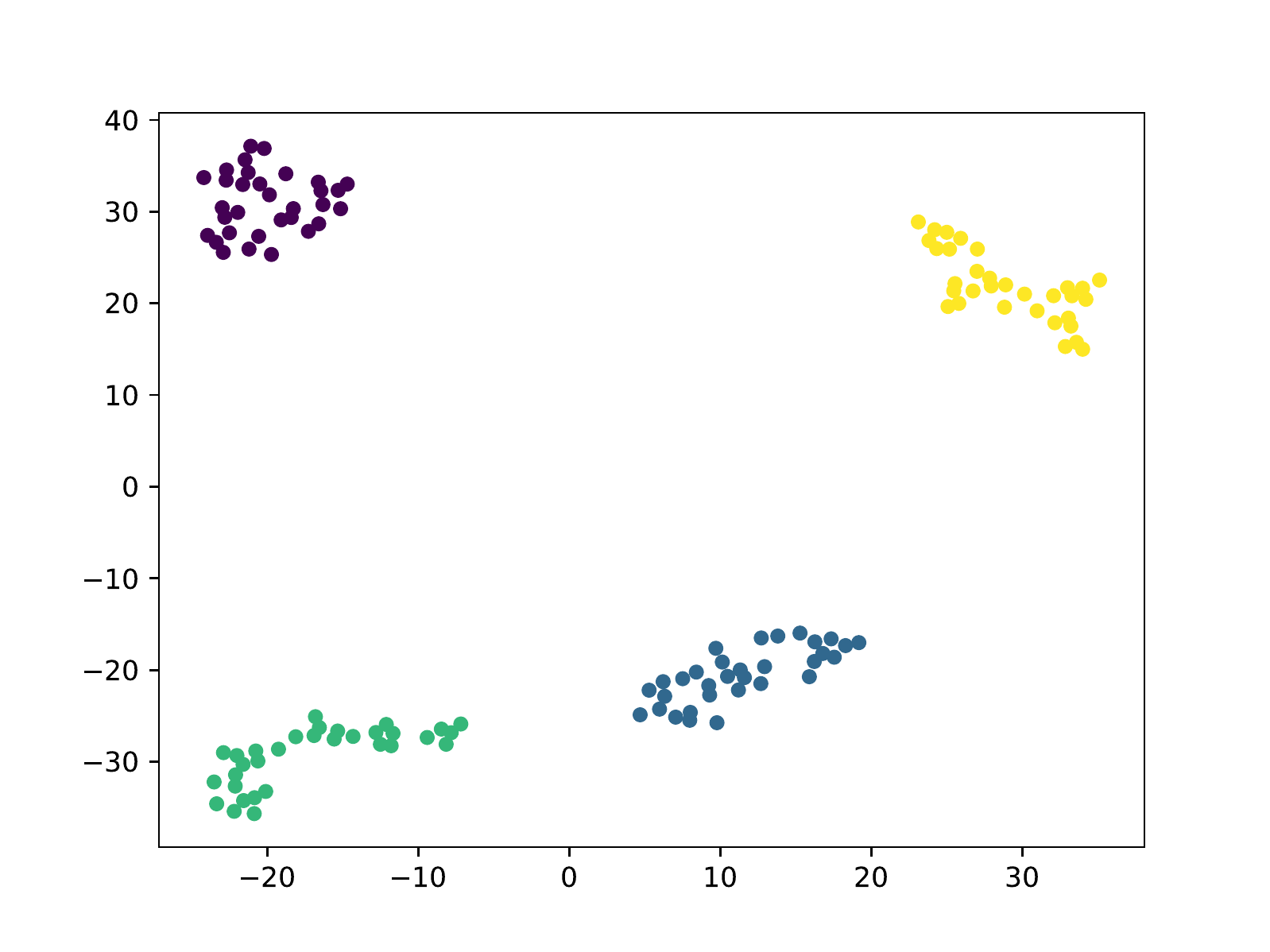}}
\subfigure[ANGM]{
\includegraphics[width=0.235\textwidth]{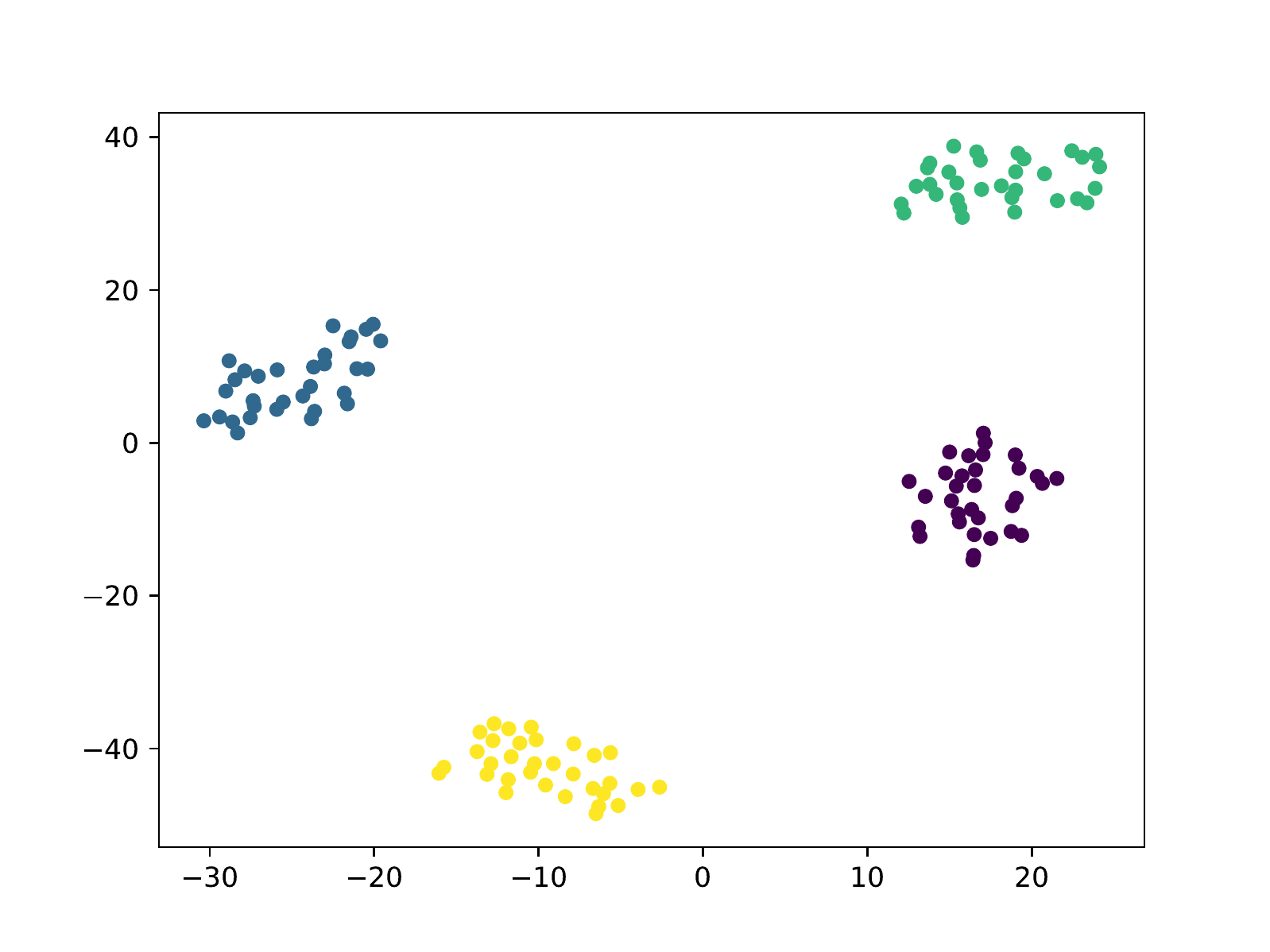}}
\caption{Visualization of representation learned by algorithms on attributed networks with communities.}
\label{fig1}
\end{figure}

\begin{figure}[htbp]
\centering
\subfigure[NOBE]{
\includegraphics[width=0.235\textwidth]{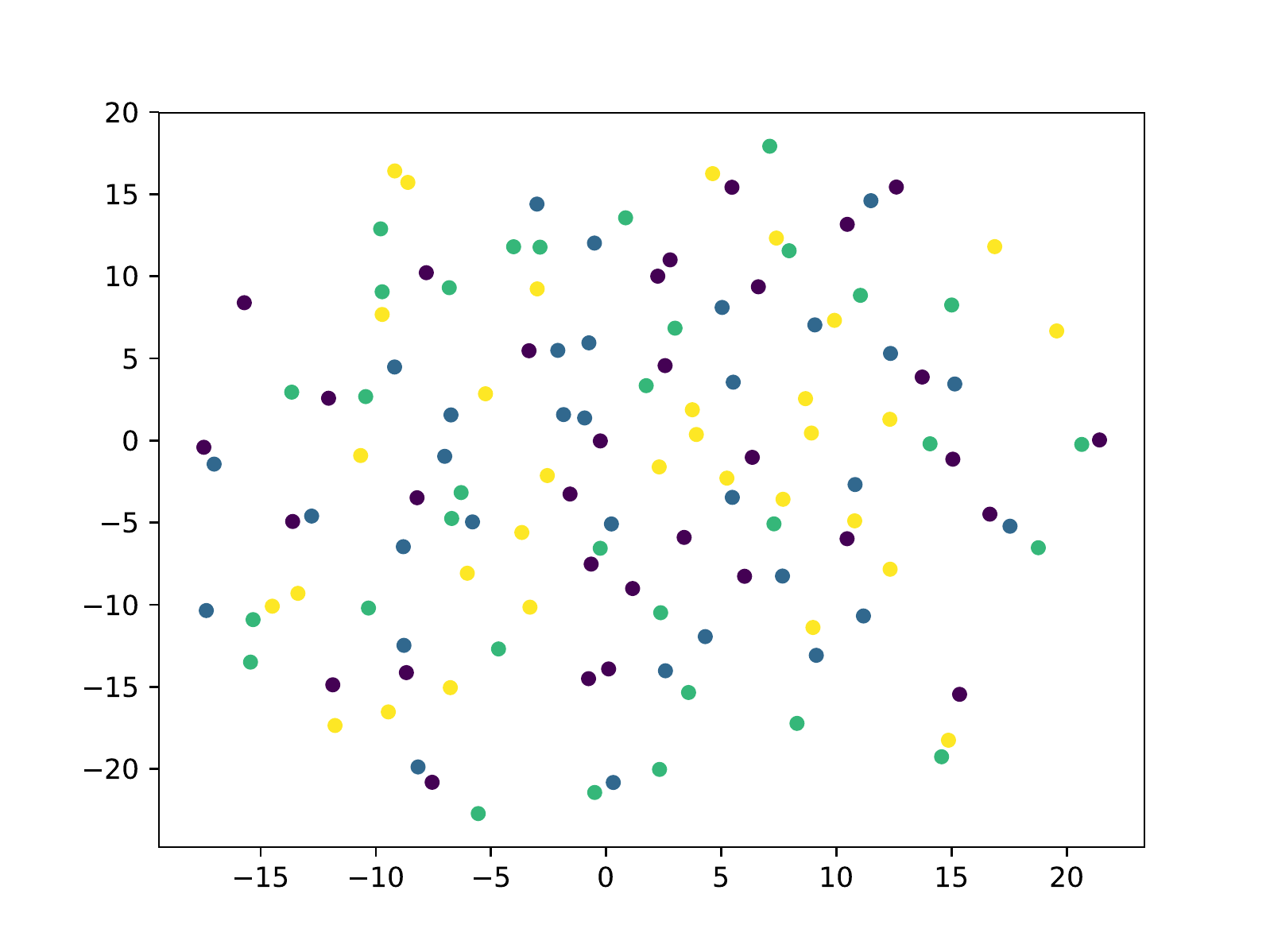}}
\subfigure[Node2Vec]{
\includegraphics[width=0.235\textwidth]{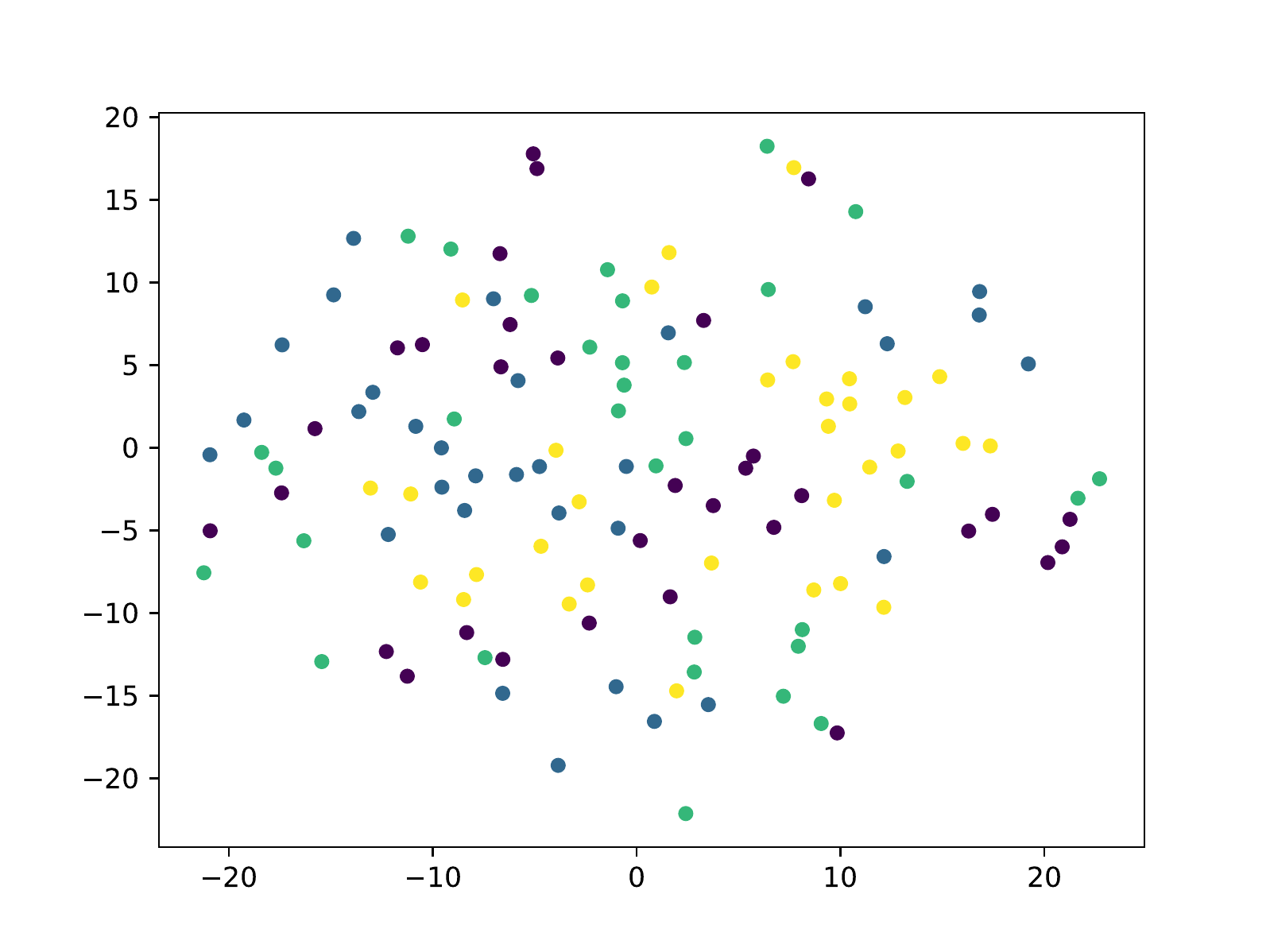}}
\subfigure[BANE]{
\includegraphics[width=0.235\textwidth]{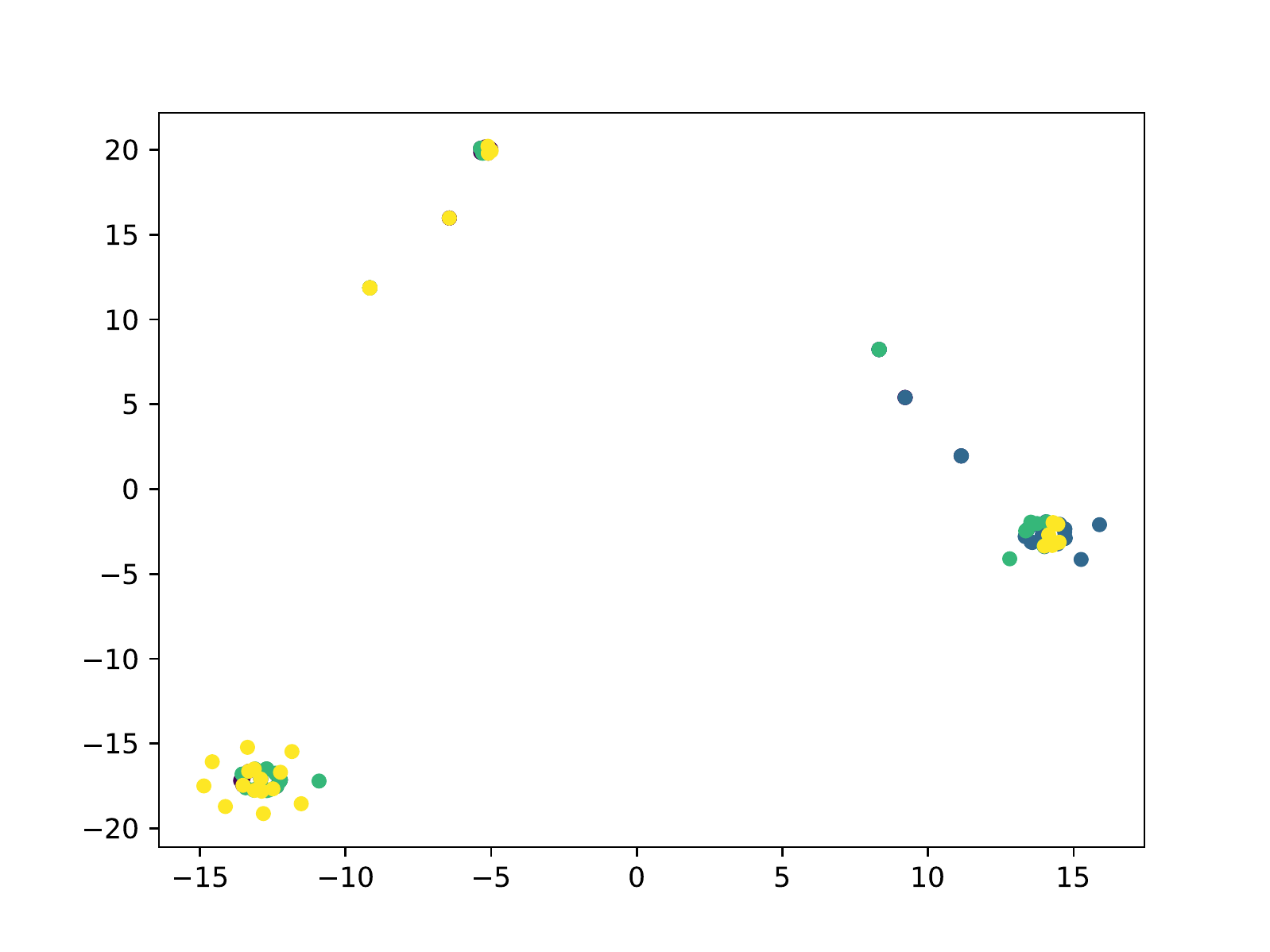}}
\subfigure[ASNE]{
\includegraphics[width=0.235\textwidth]{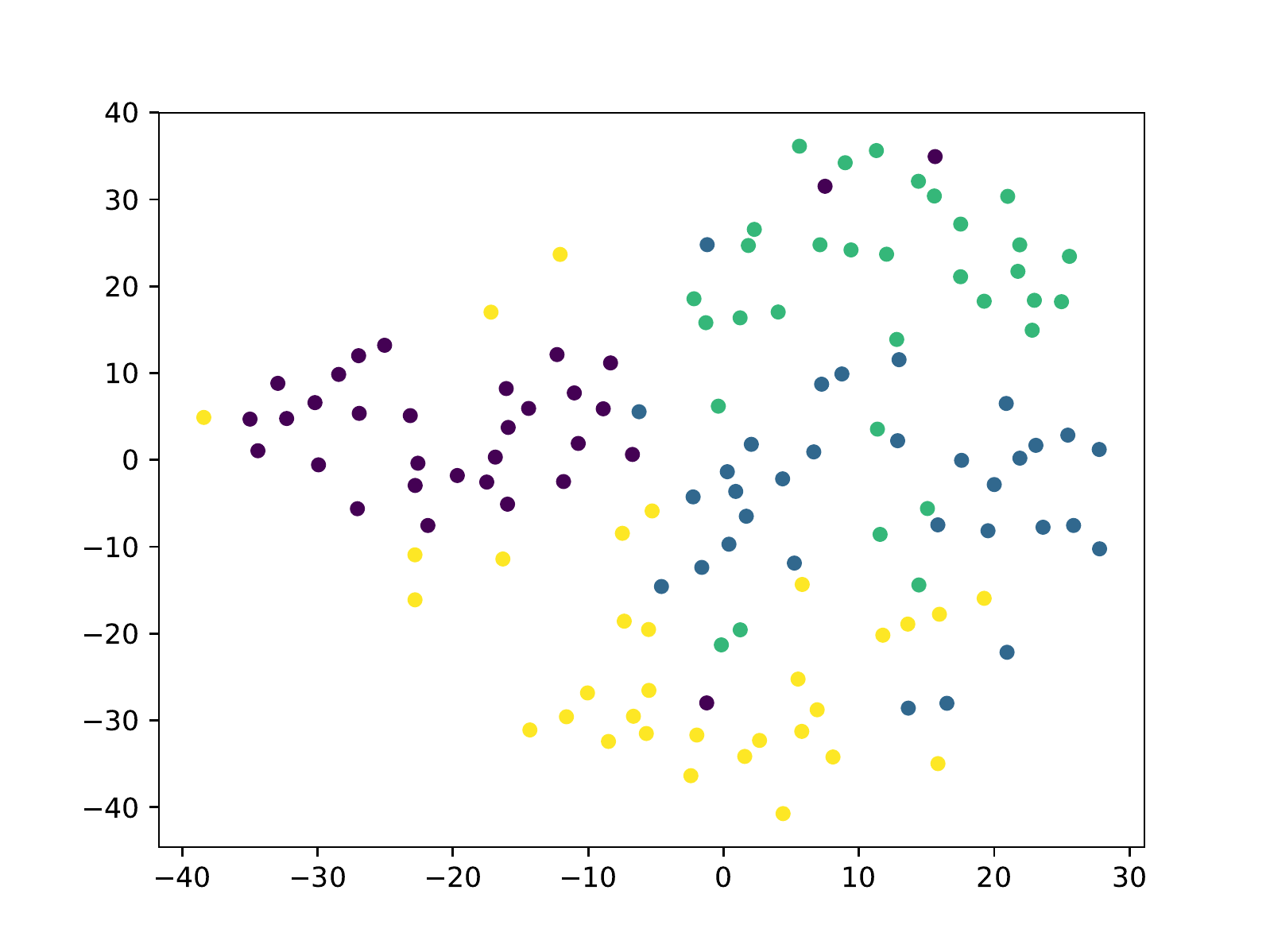}}
\subfigure[ANRL]{
\includegraphics[width=0.235\textwidth]{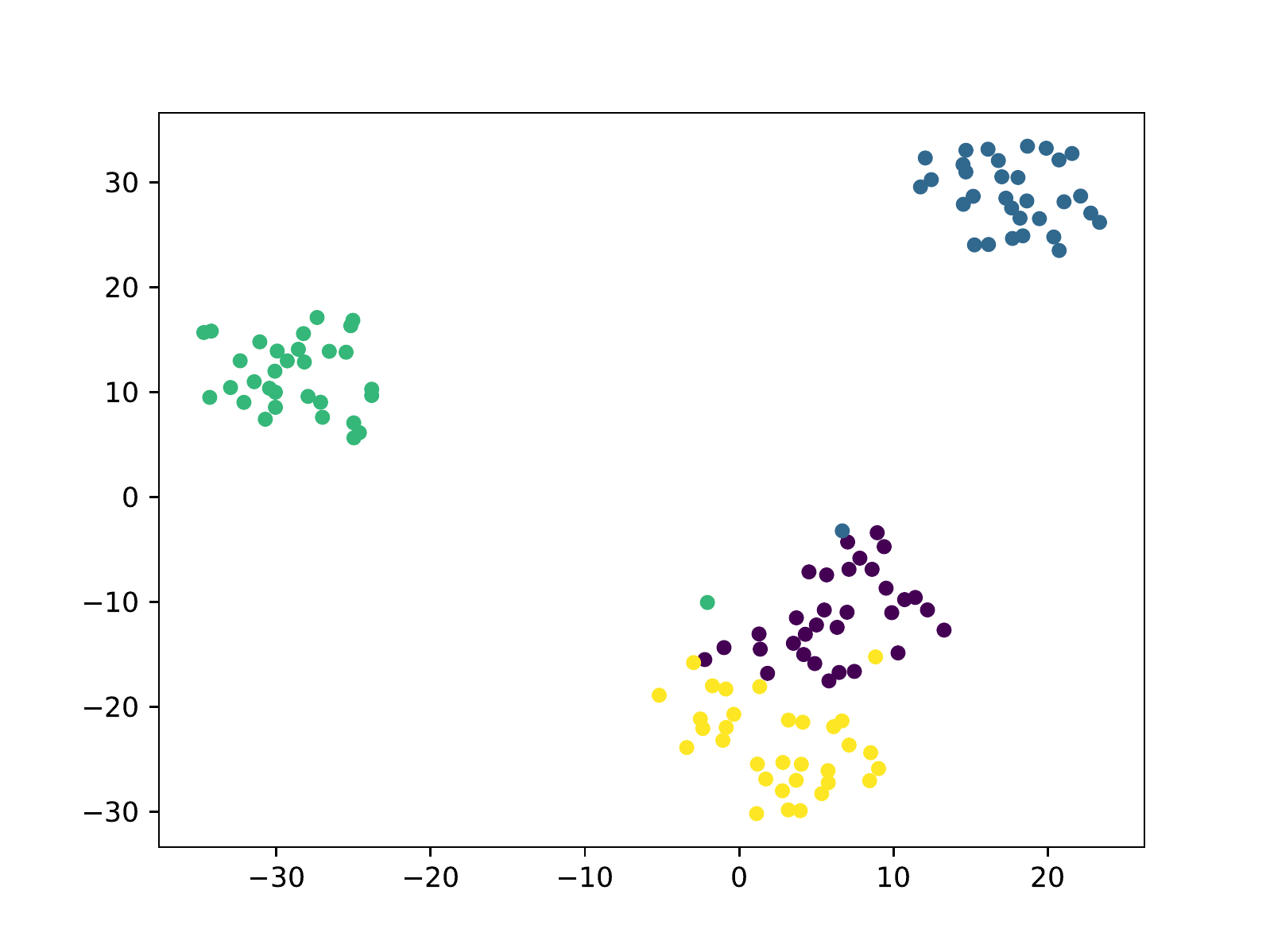}}
\subfigure[VGAE]{
\includegraphics[width=0.235\textwidth]{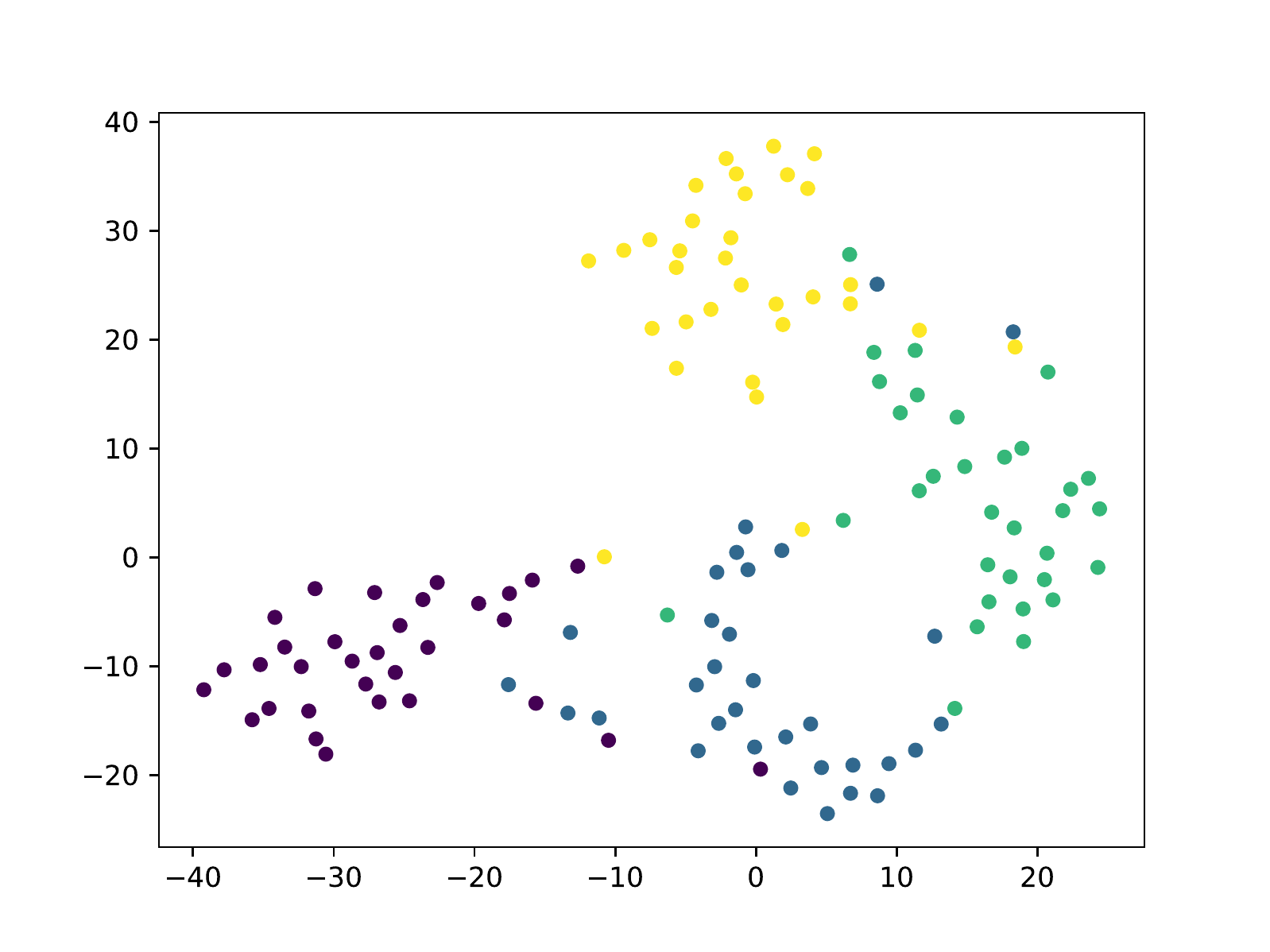}}
\subfigure[ARVGE]{
\includegraphics[width=0.235\textwidth]{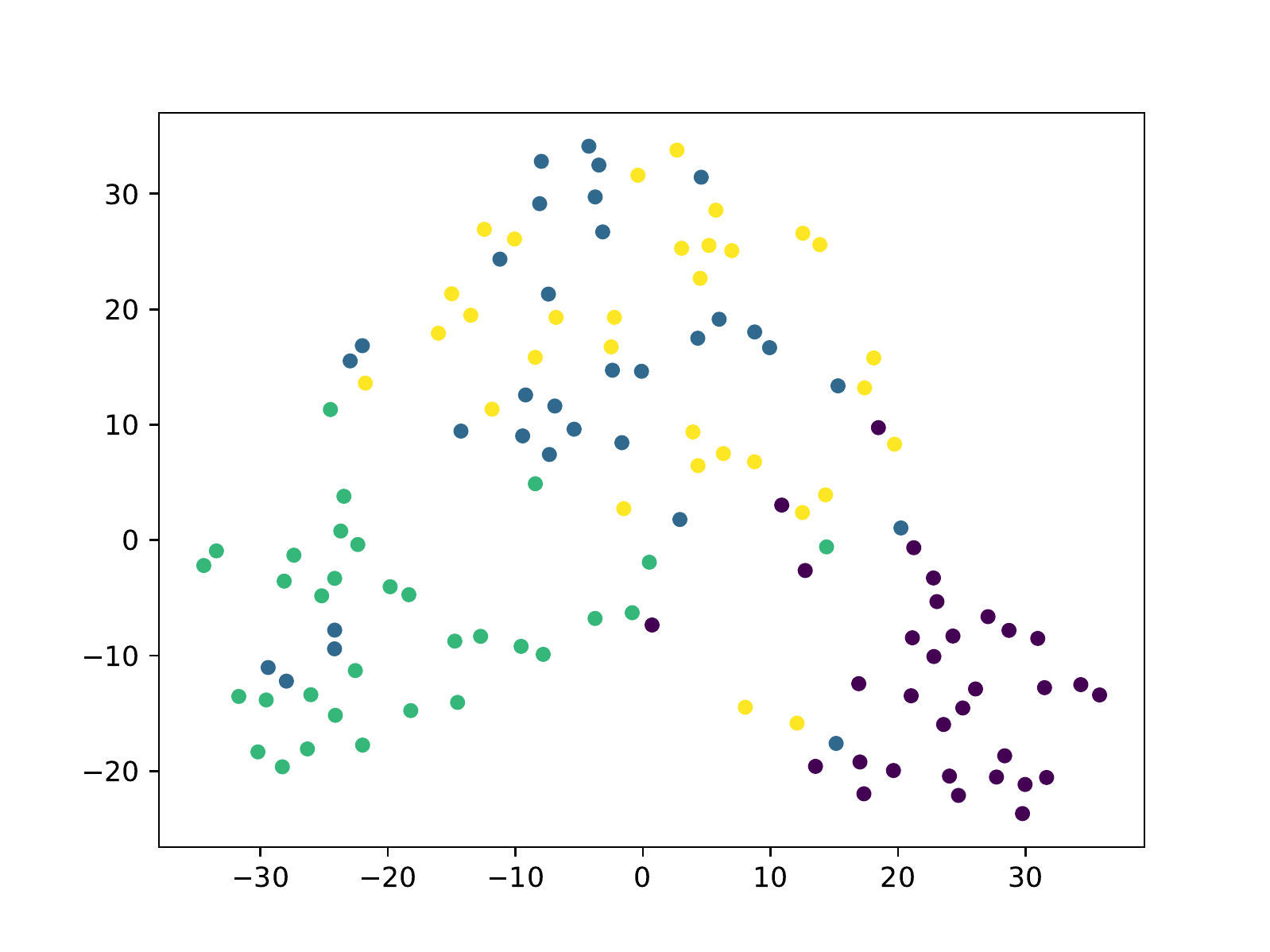}}
\subfigure[G2G]{
\includegraphics[width=0.235\textwidth]{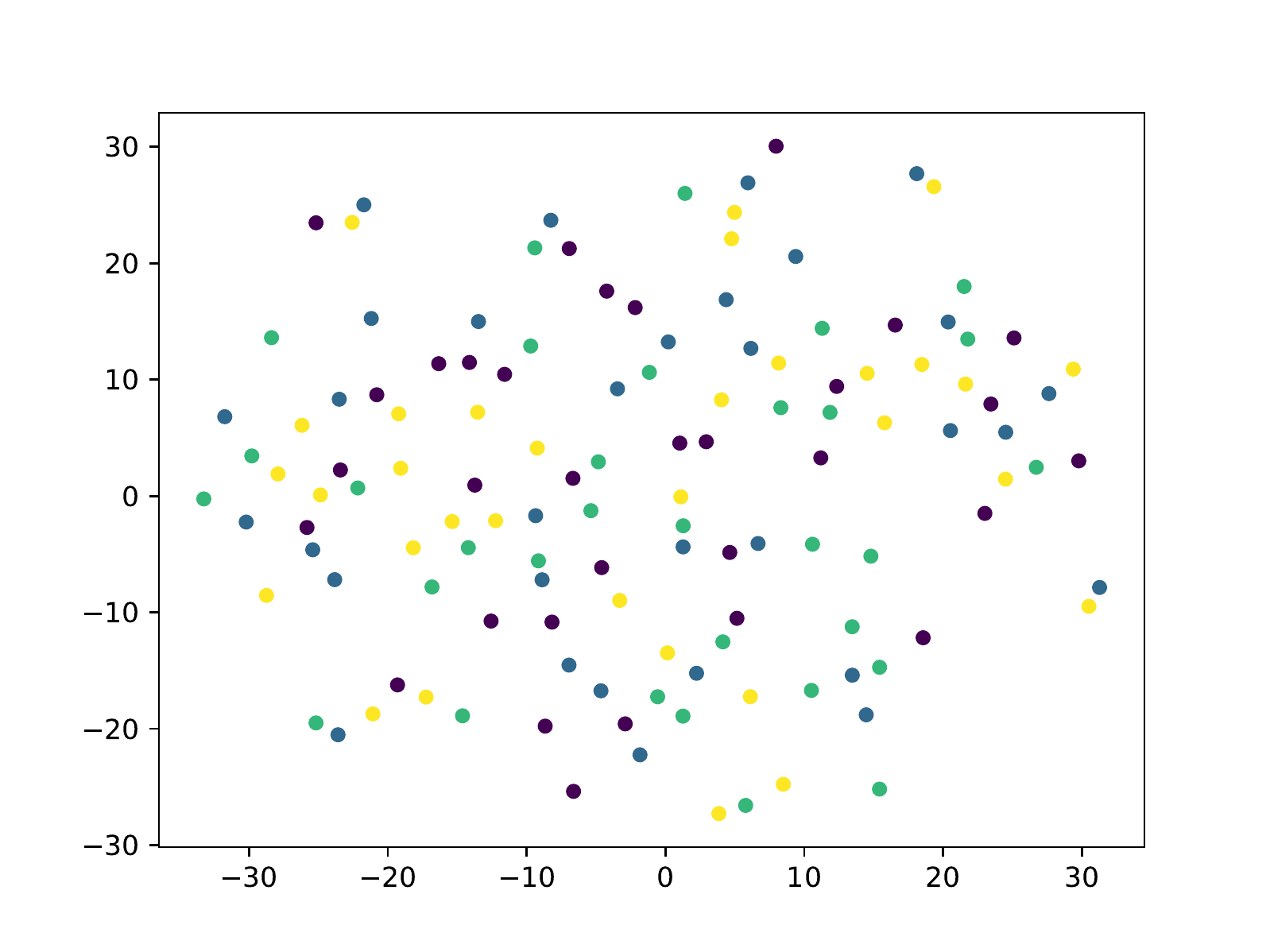}}
\subfigure[GATE]{
\includegraphics[width=0.235\textwidth]{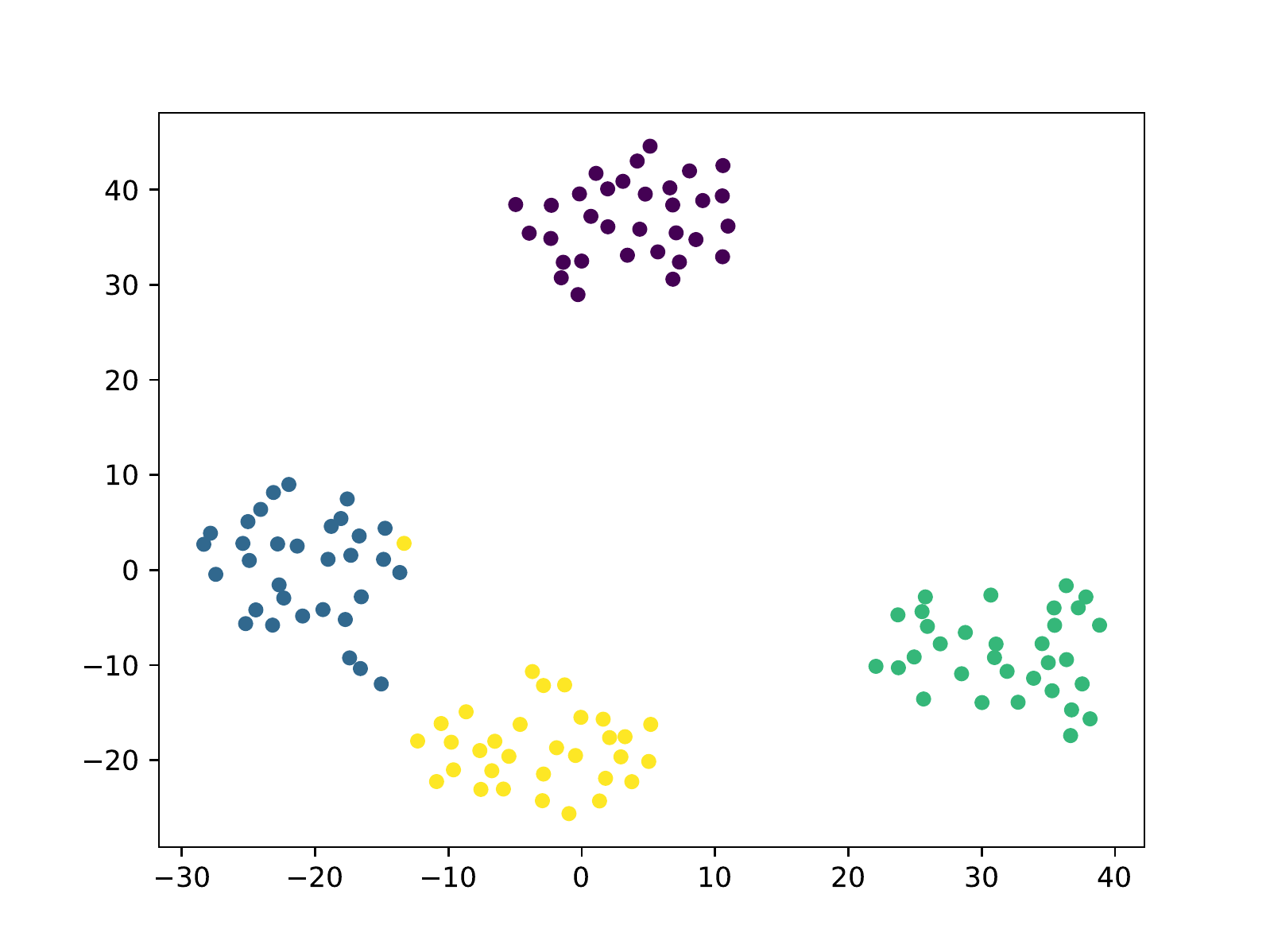}}
\subfigure[ANGM]{
\includegraphics[width=0.235\textwidth]{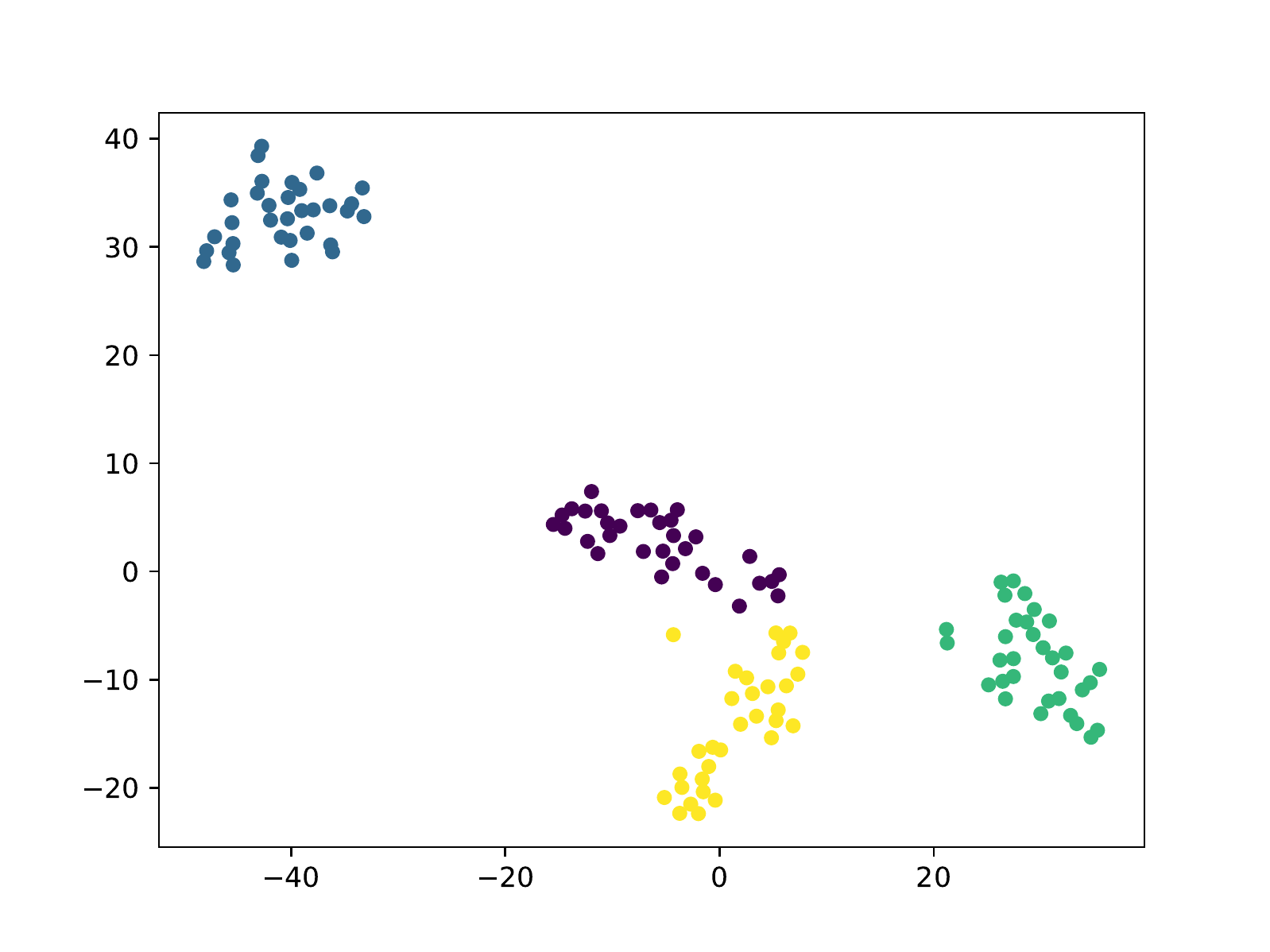}}
\caption{Visualization of representation learned by algorithms on attributed networks with multipartite structures.}
\label{fig2}
\end{figure}

\subsubsection{Experimental results}\label{Synth-results}
\begin{figure}[htbp]
\centering
\subfigure[NOBE]{
\includegraphics[width=0.235\textwidth]{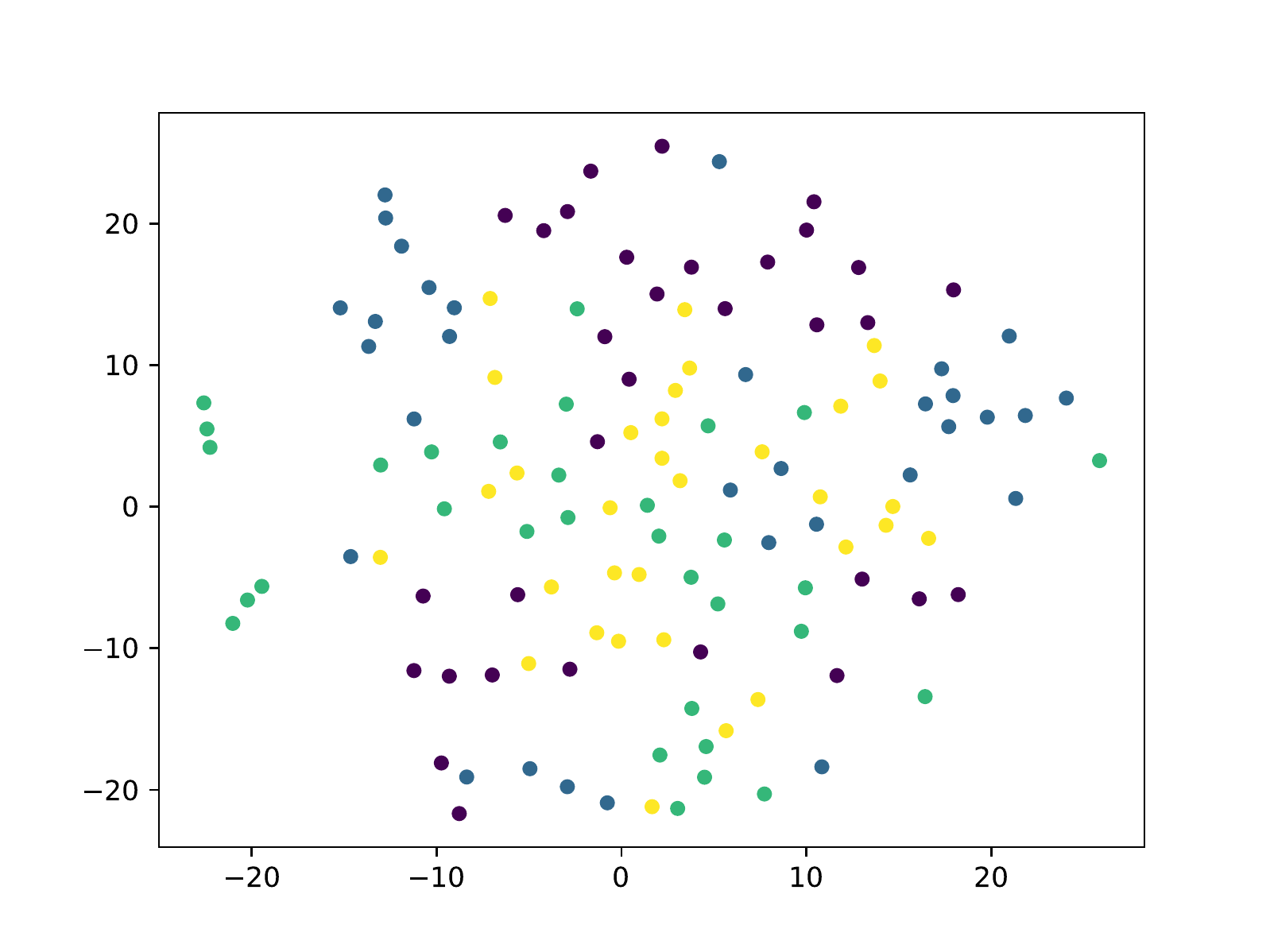}}
\subfigure[Node2Vec]{
\includegraphics[width=0.235\textwidth]{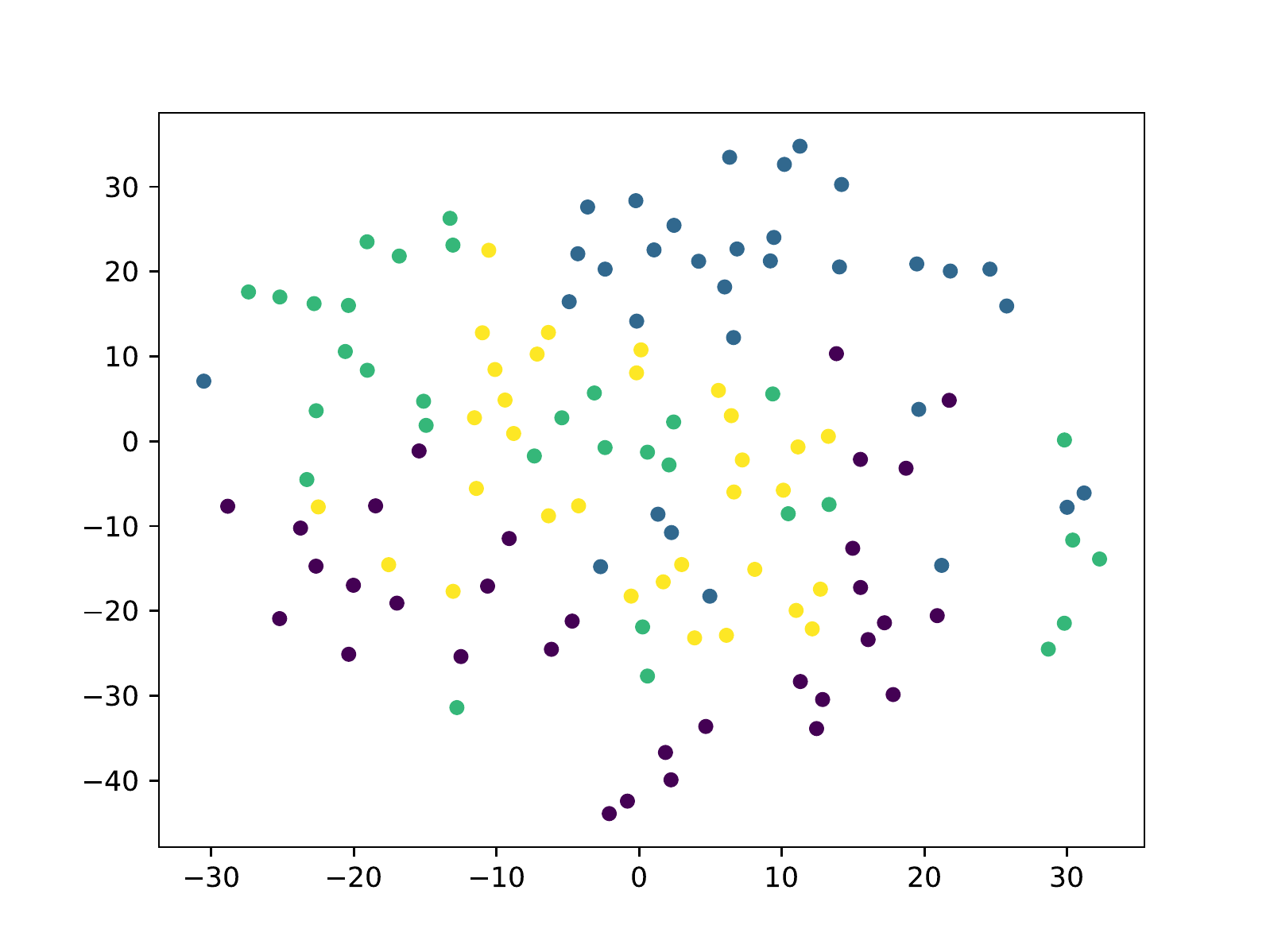}}
\subfigure[BANE]{
\includegraphics[width=0.235\textwidth]{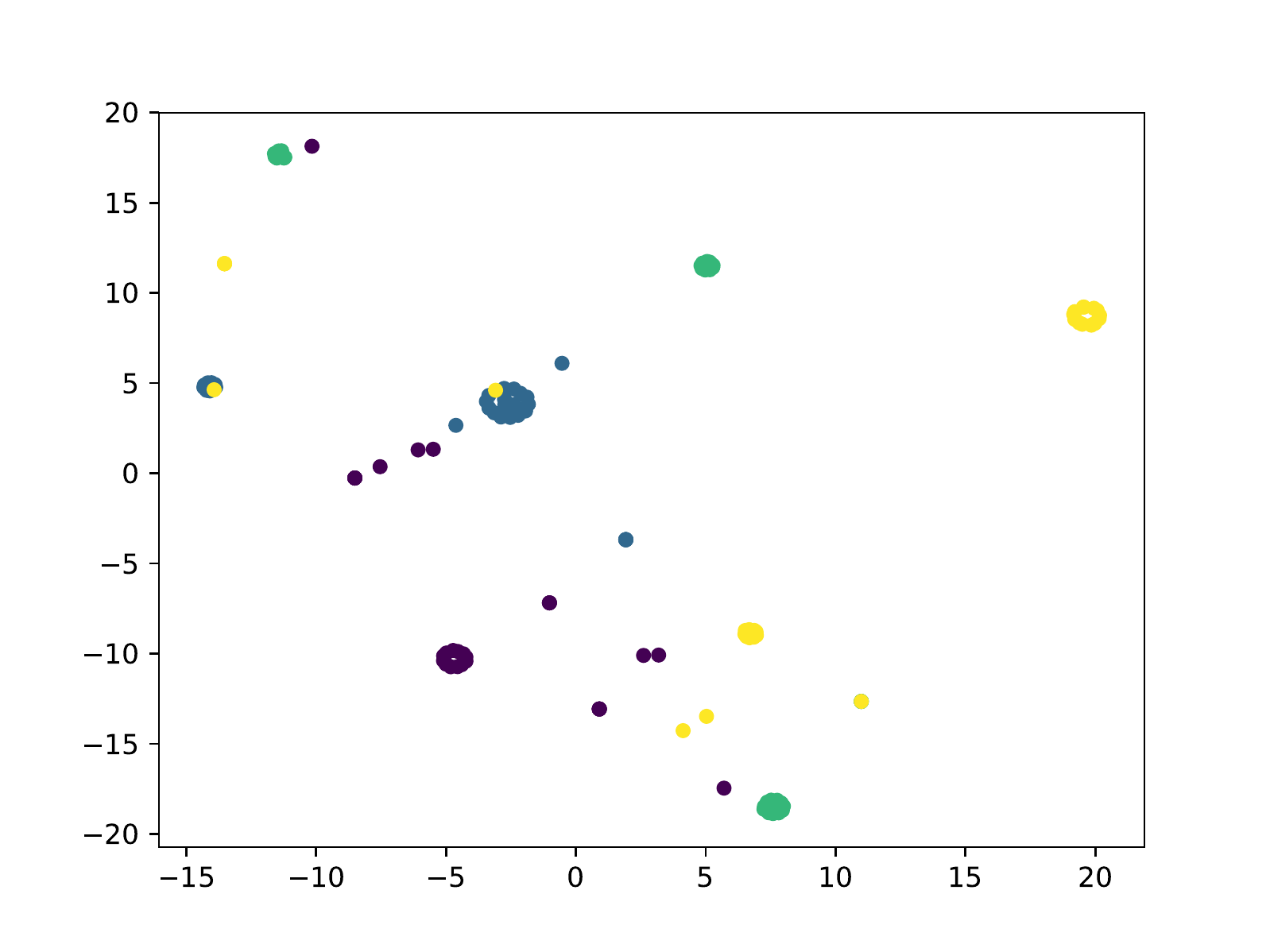}}
\subfigure[ASNE]{
\includegraphics[width=0.235\textwidth]{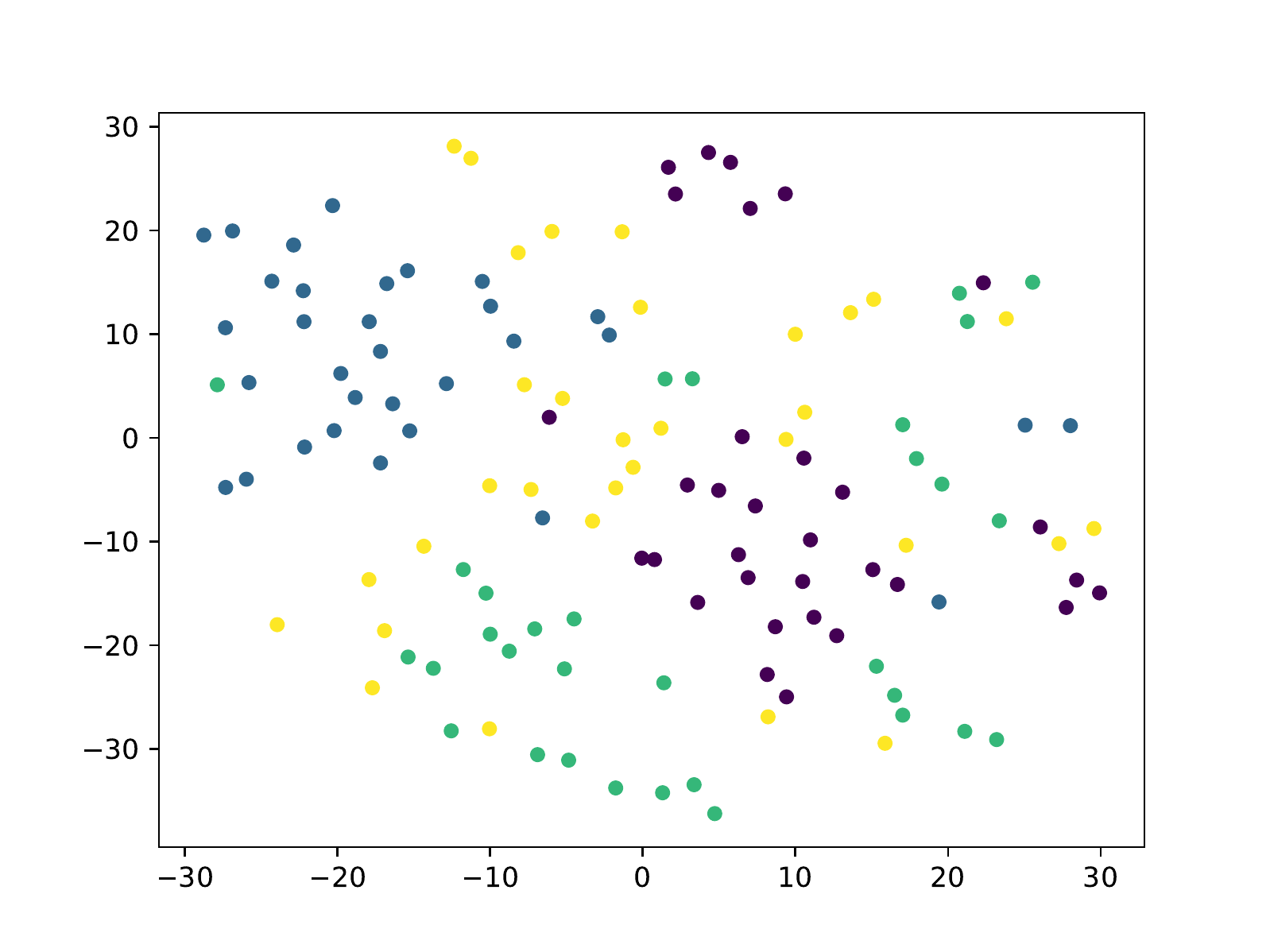}}
\subfigure[ANRL]{
\includegraphics[width=0.235\textwidth]{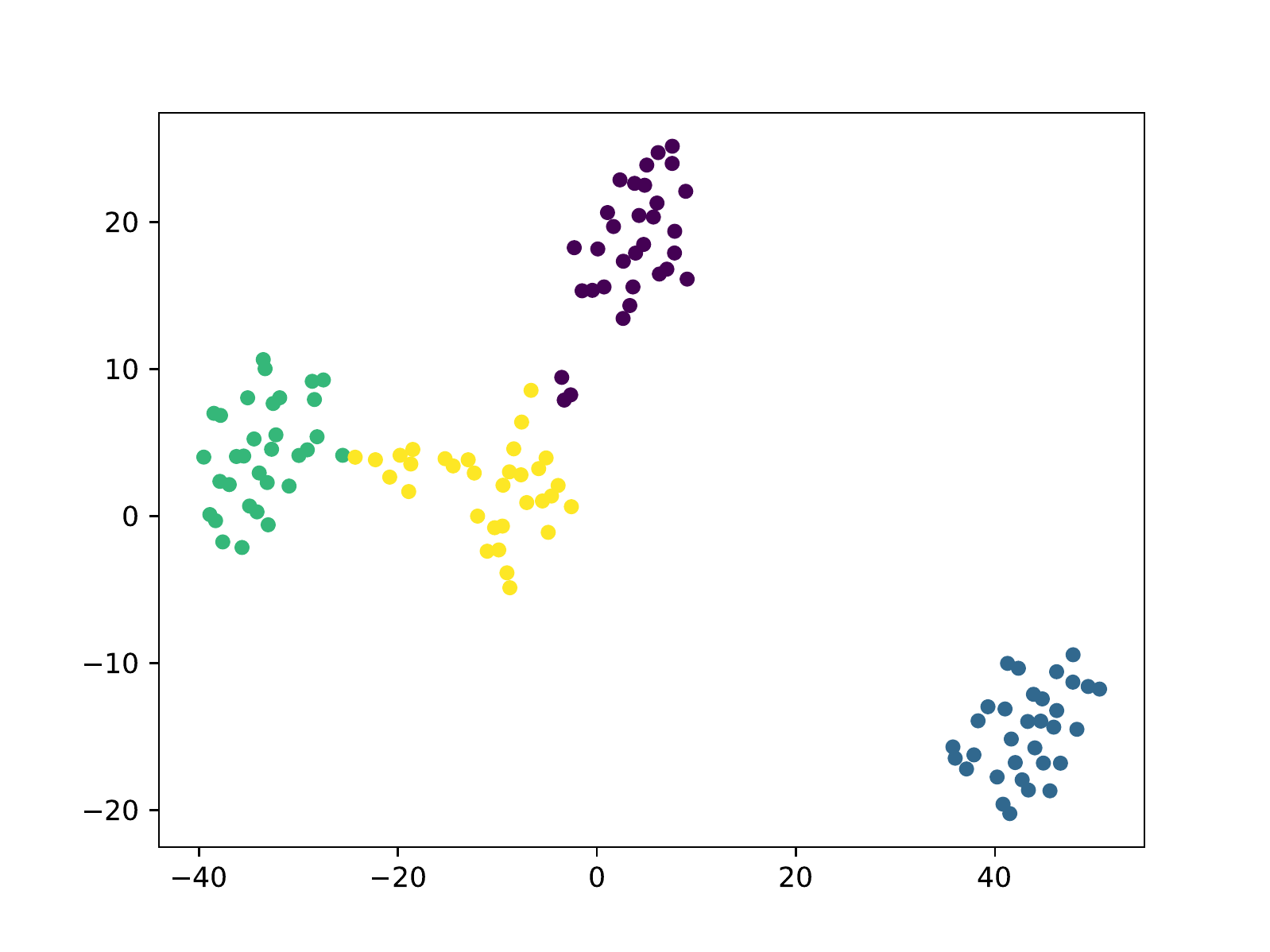}}
\subfigure[VGAE]{
\includegraphics[width=0.235\textwidth]{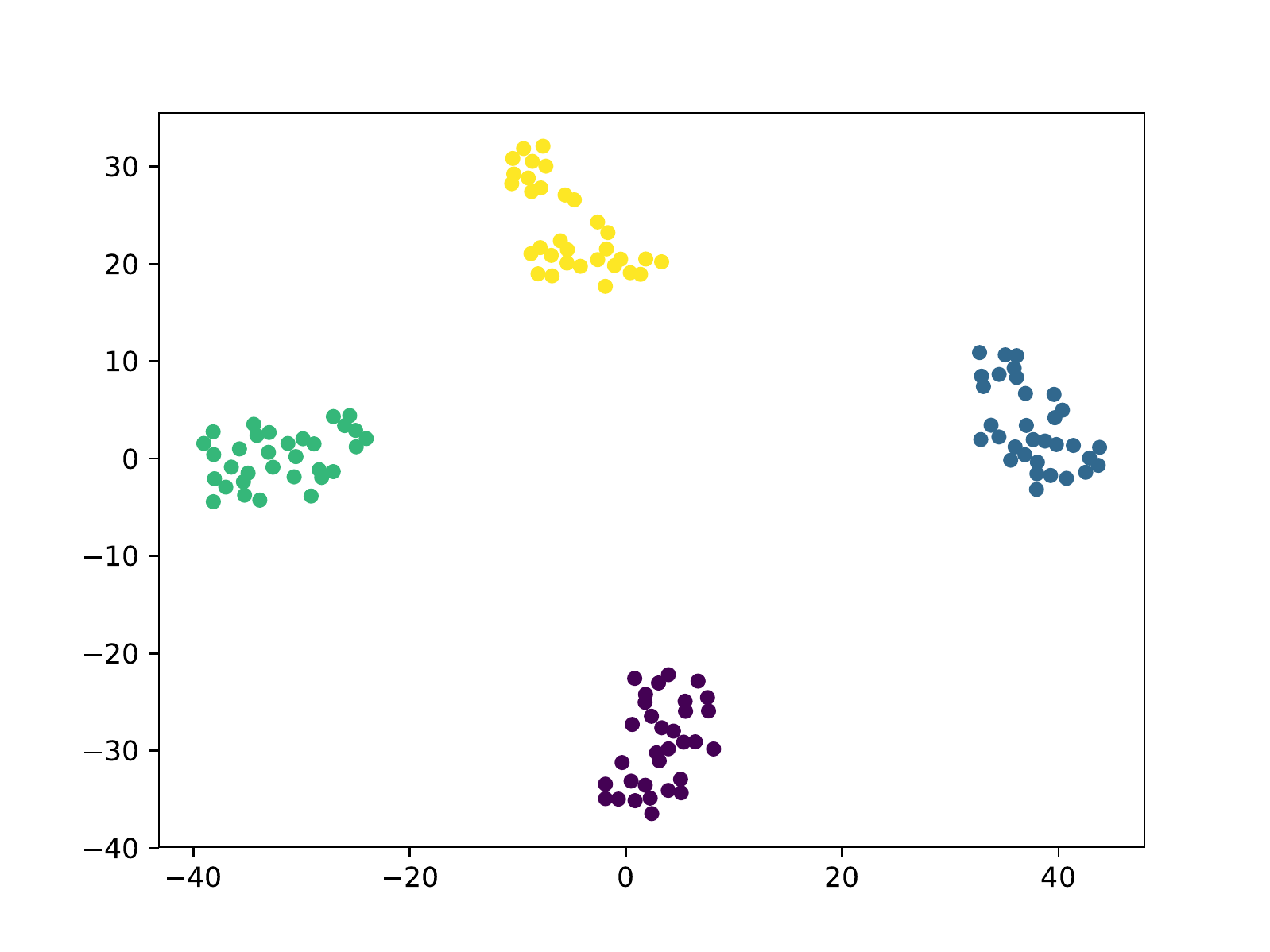}}
\subfigure[ARVGE]{
\includegraphics[width=0.235\textwidth]{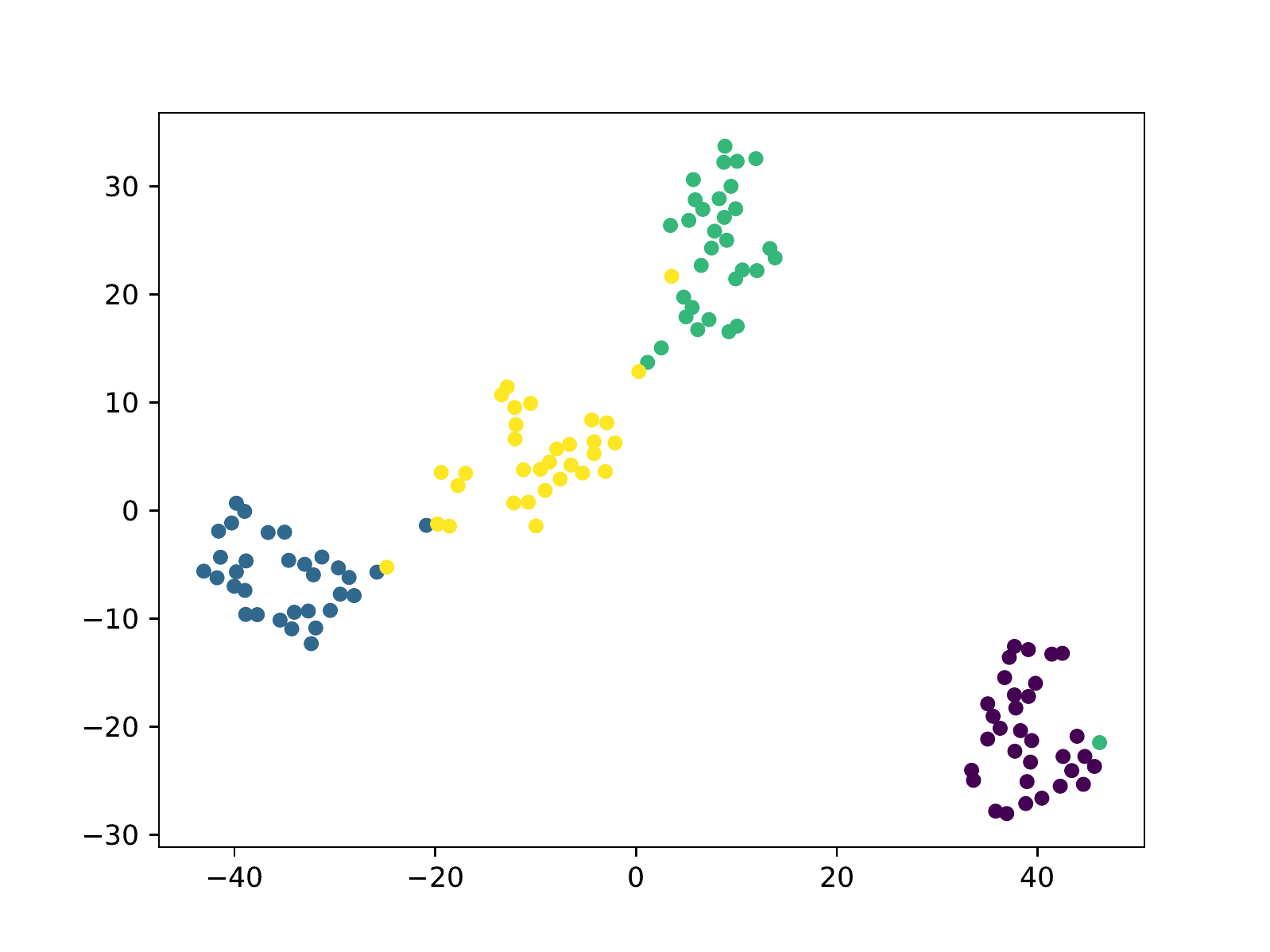}}
\subfigure[G2G]{
\includegraphics[width=0.235\textwidth]{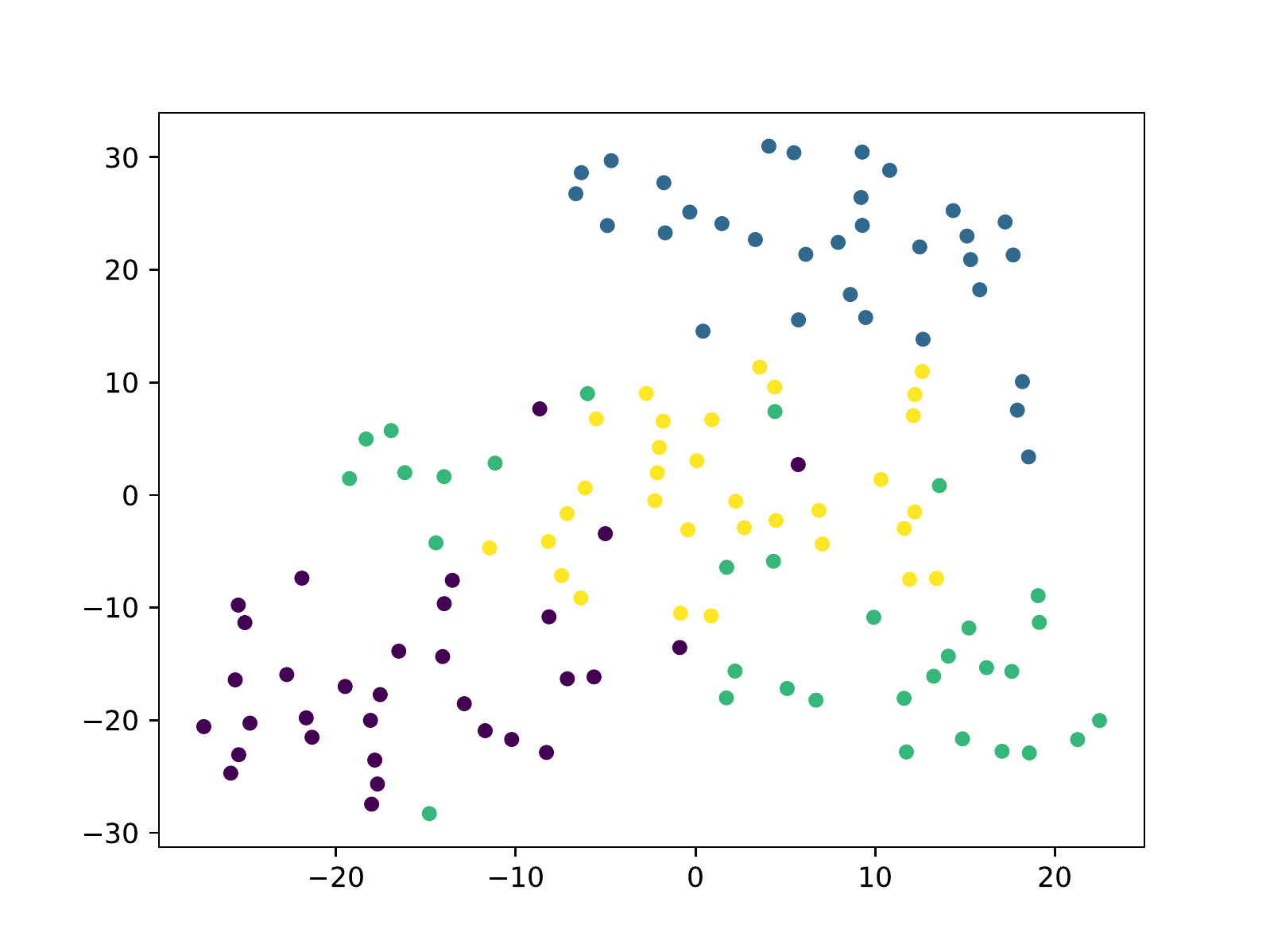}}
\subfigure[GATE]{
\includegraphics[width=0.235\textwidth]{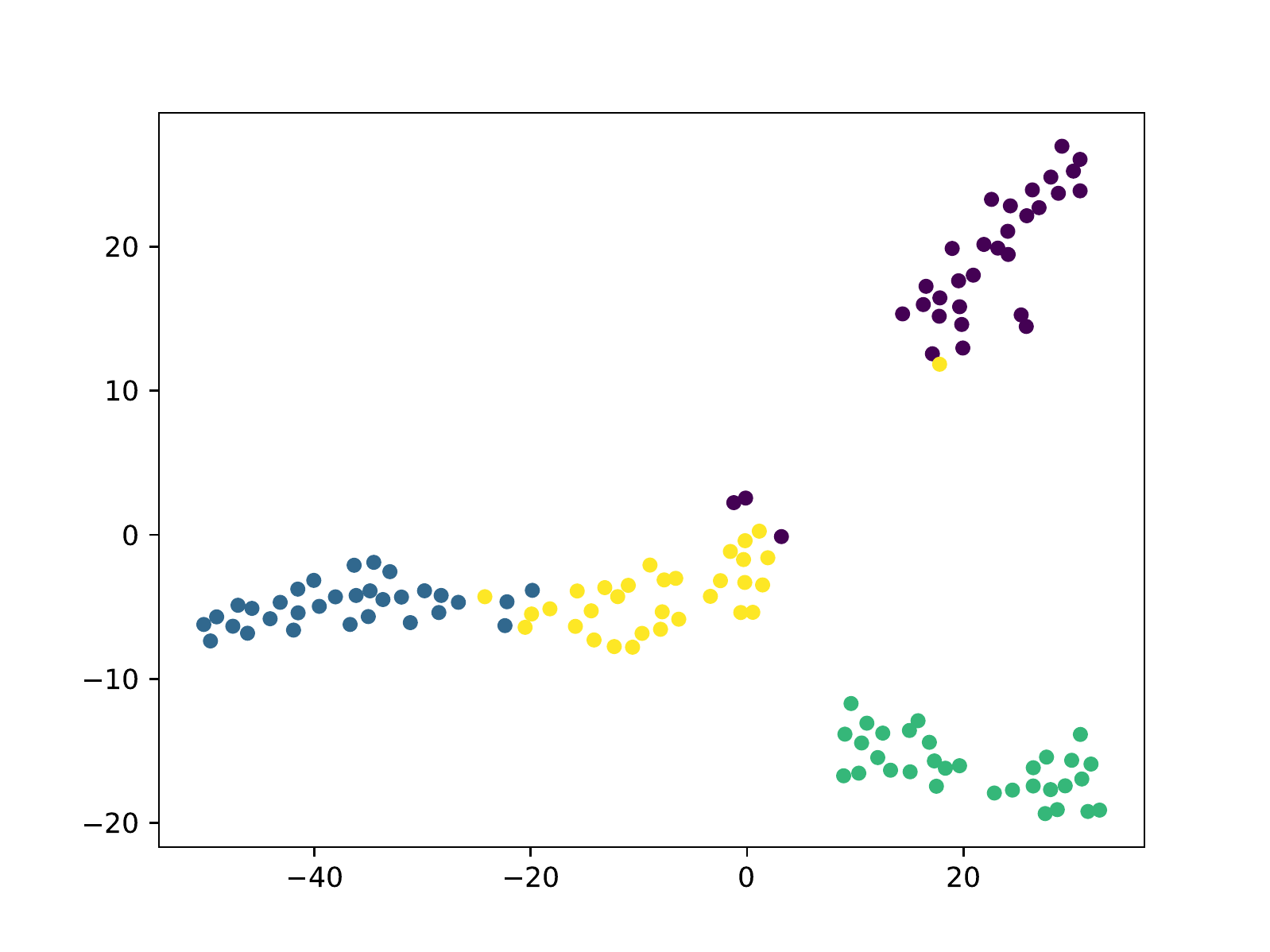}}
\subfigure[ANGM]{
\includegraphics[width=0.235\textwidth]{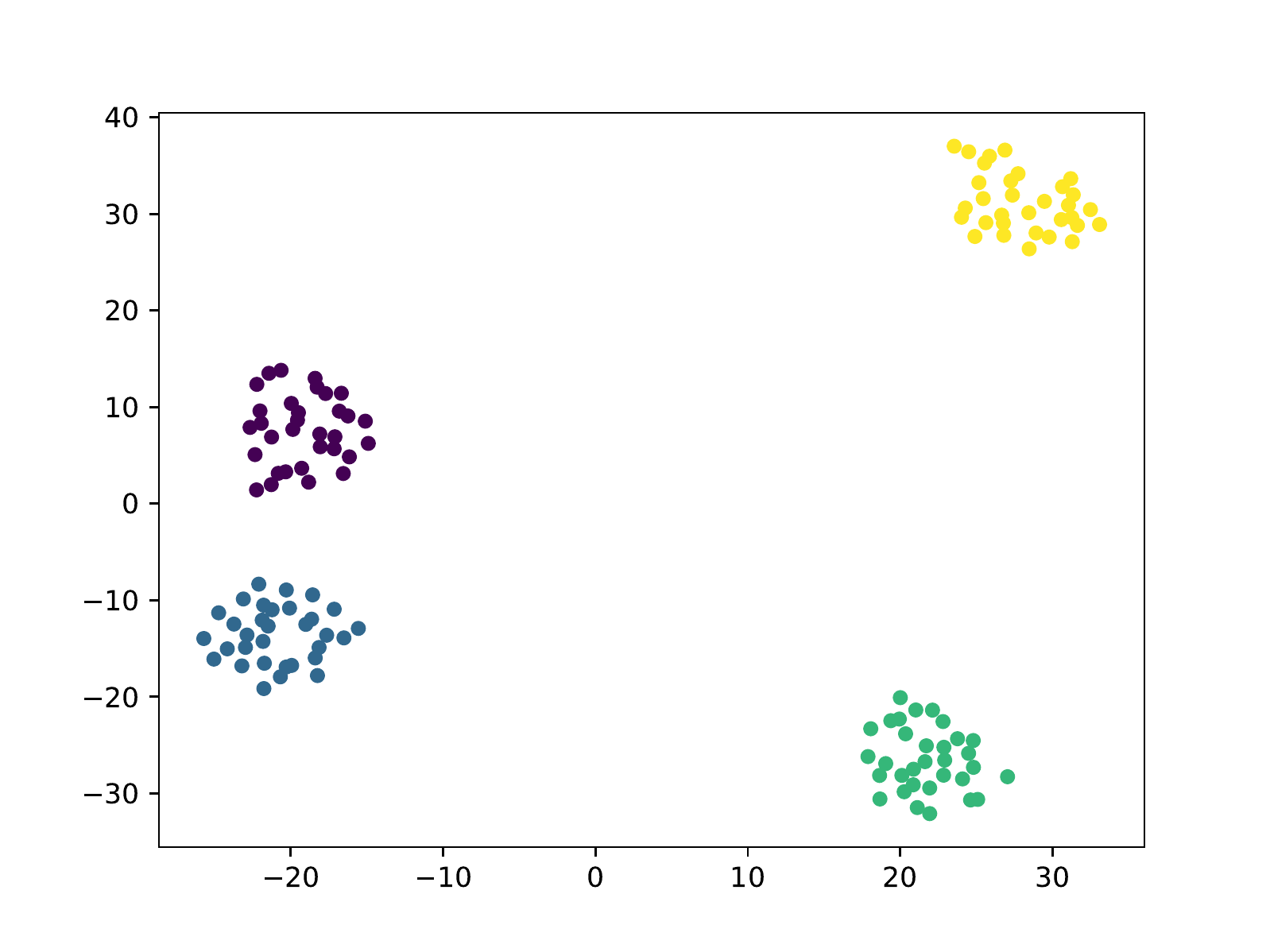}}
\caption{Visualization of representation learned by algorithms on attributed networks with hubs.}
\label{fig3}
\end{figure}

\begin{figure}[htbp]
\centering
\subfigure[NOBE]{
\includegraphics[width=0.235\textwidth]{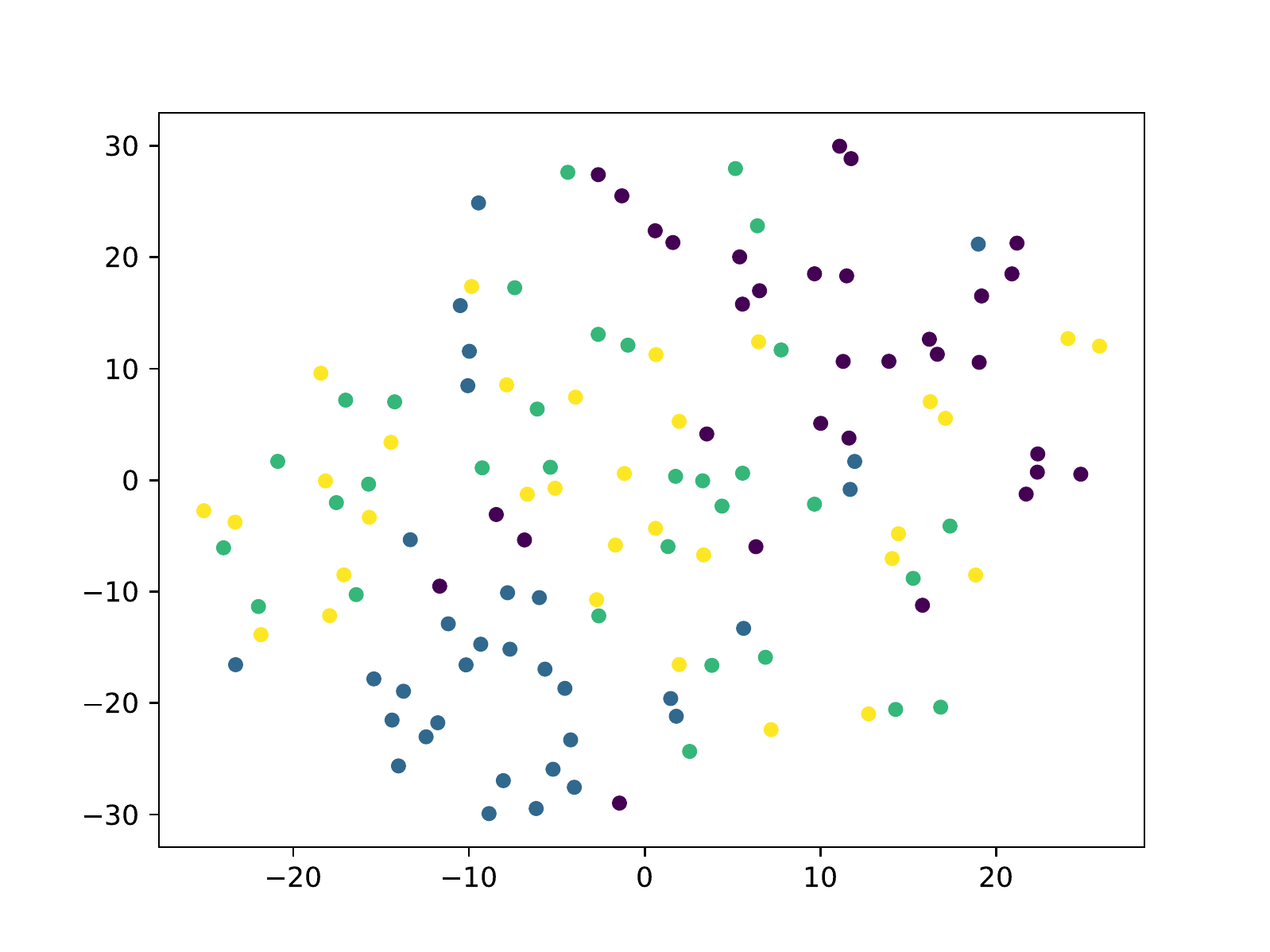}}
\subfigure[Node2Vec]{
\includegraphics[width=0.235\textwidth]{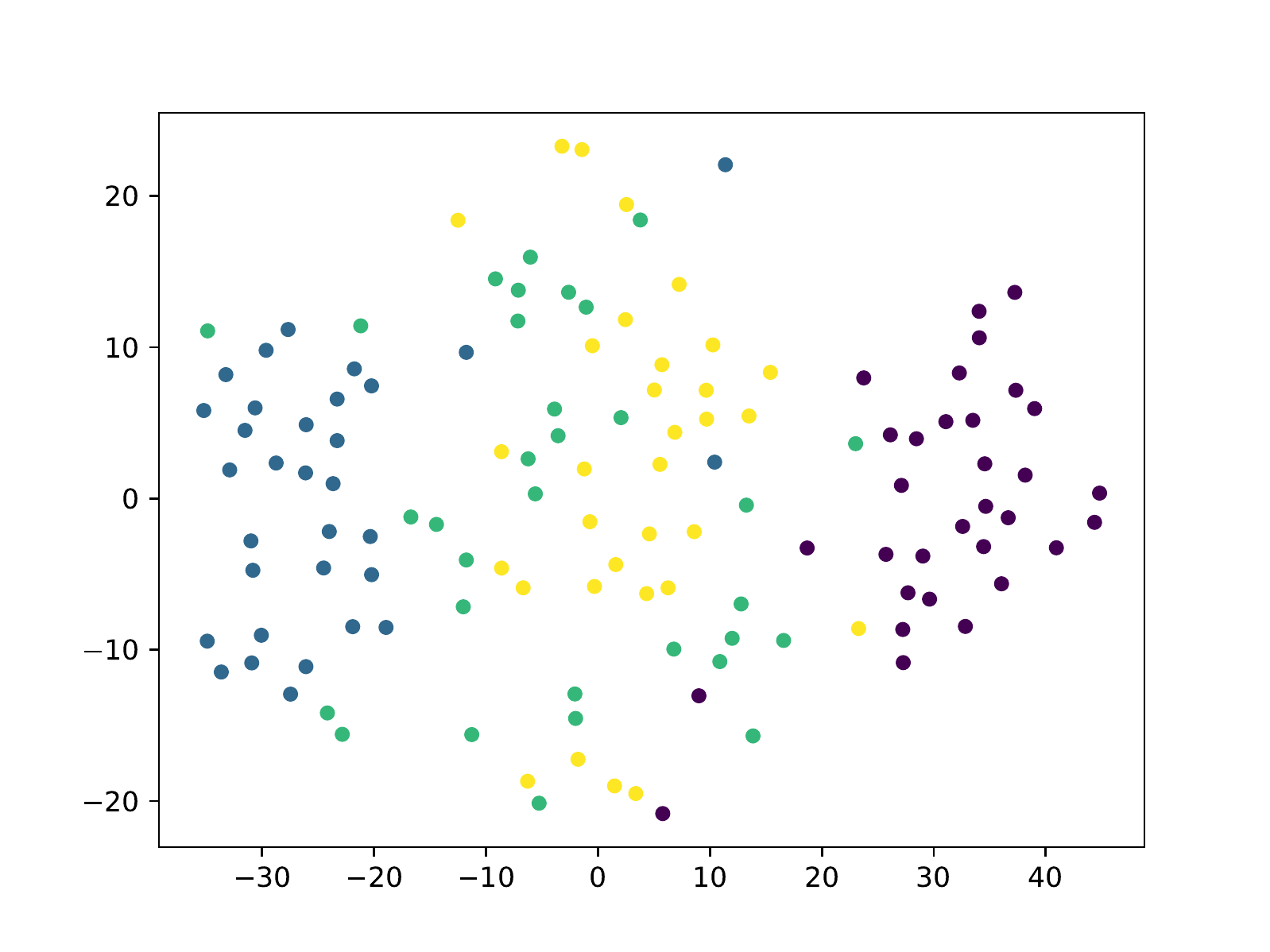}}
\subfigure[BANE]{
\includegraphics[width=0.235\textwidth]{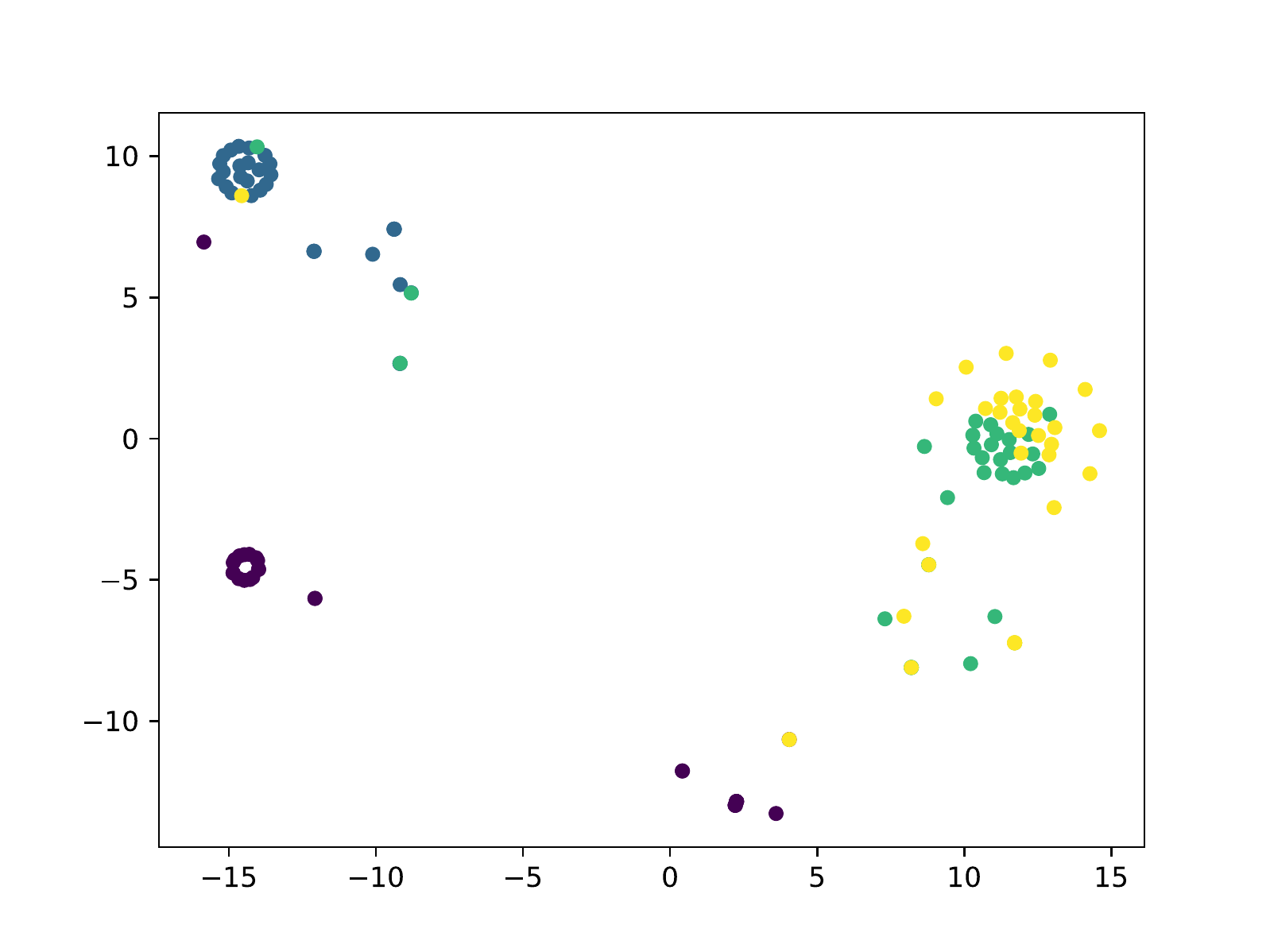}}
\subfigure[ASNE]{
\includegraphics[width=0.235\textwidth]{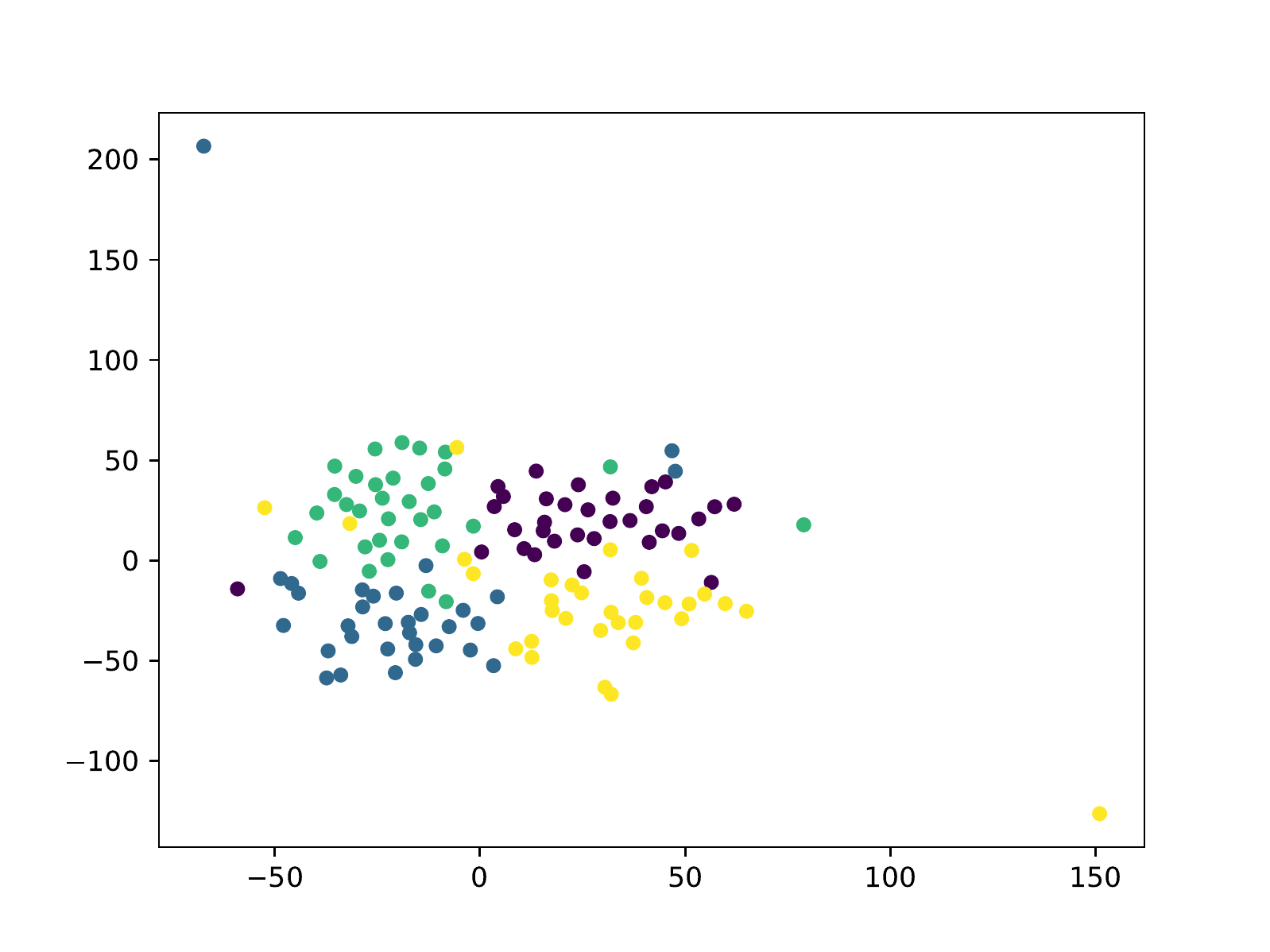}}
\subfigure[ANRL]{
\includegraphics[width=0.235\textwidth]{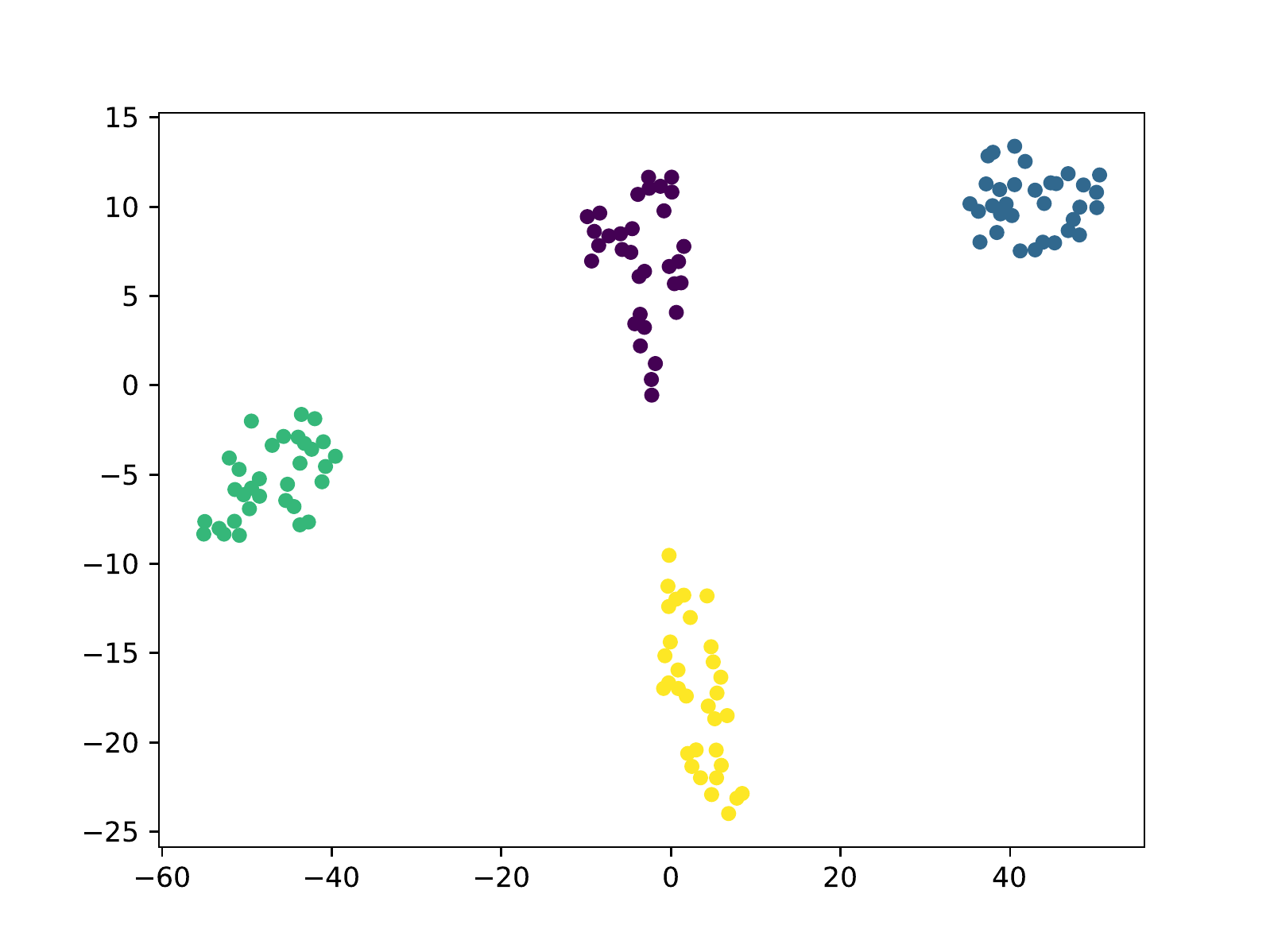}}
\subfigure[VGAE]{
\includegraphics[width=0.235\textwidth]{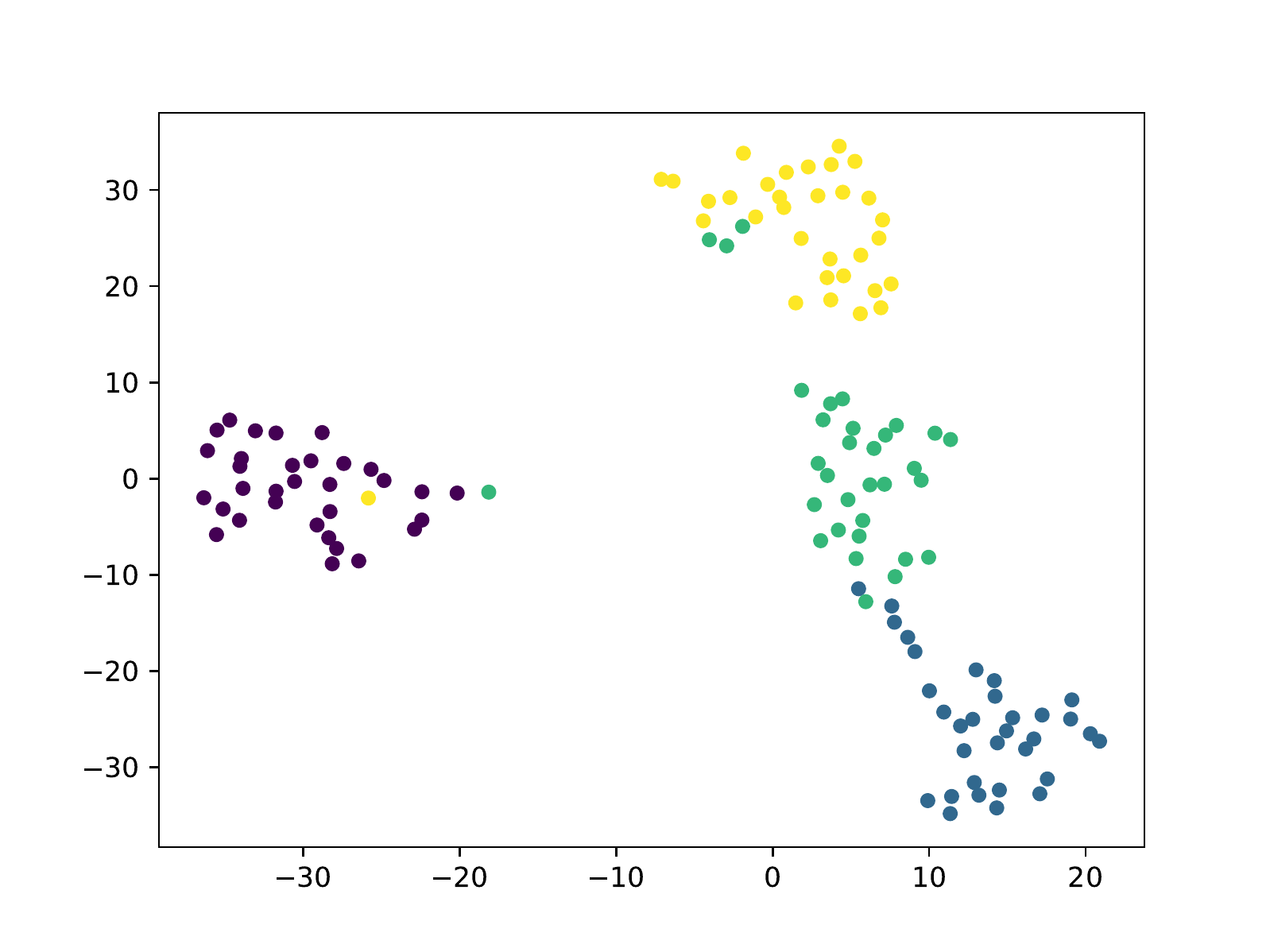}}
\subfigure[ARVGE]{
\includegraphics[width=0.235\textwidth]{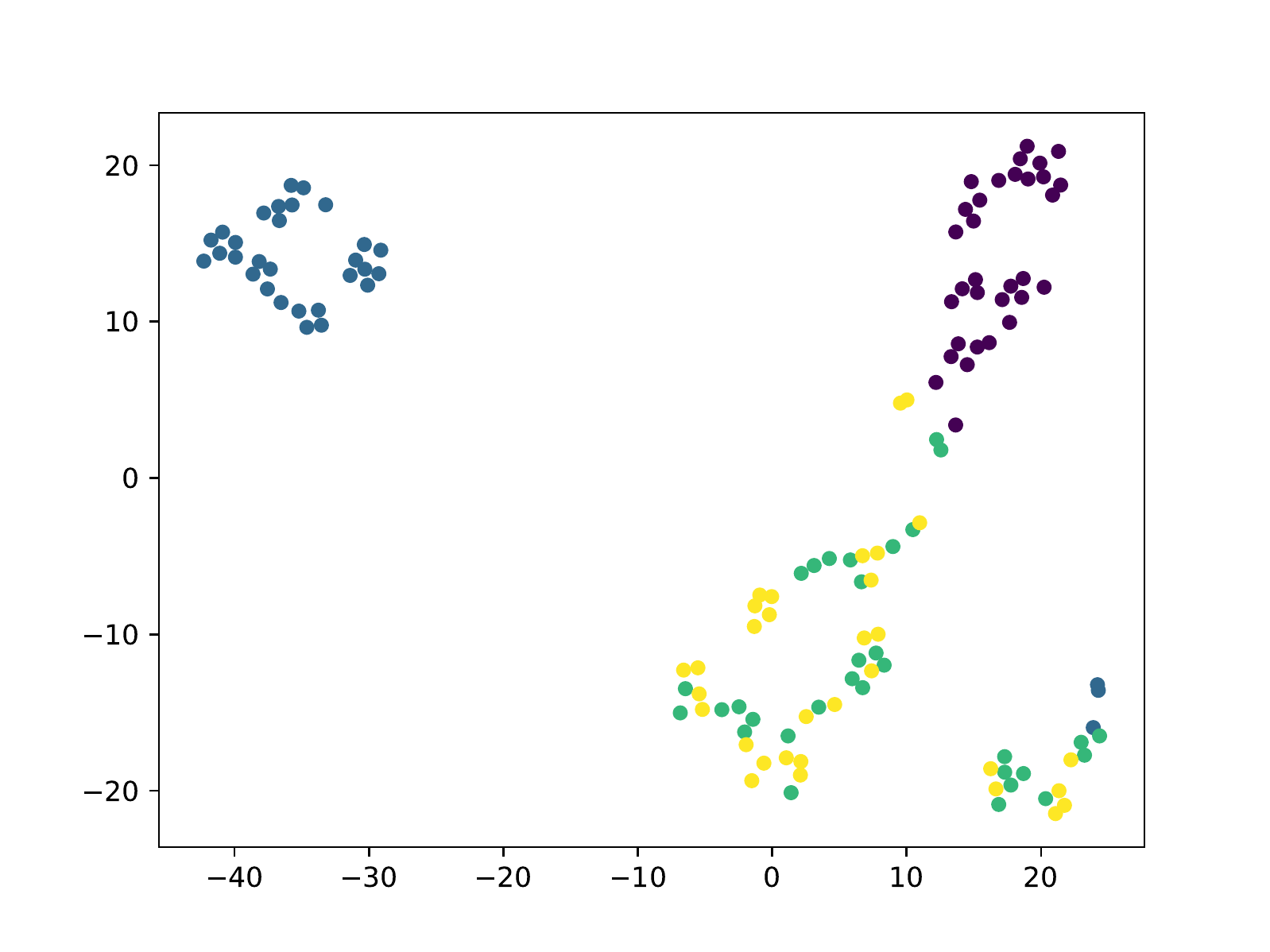}}
\subfigure[G2G]{
\includegraphics[width=0.235\textwidth]{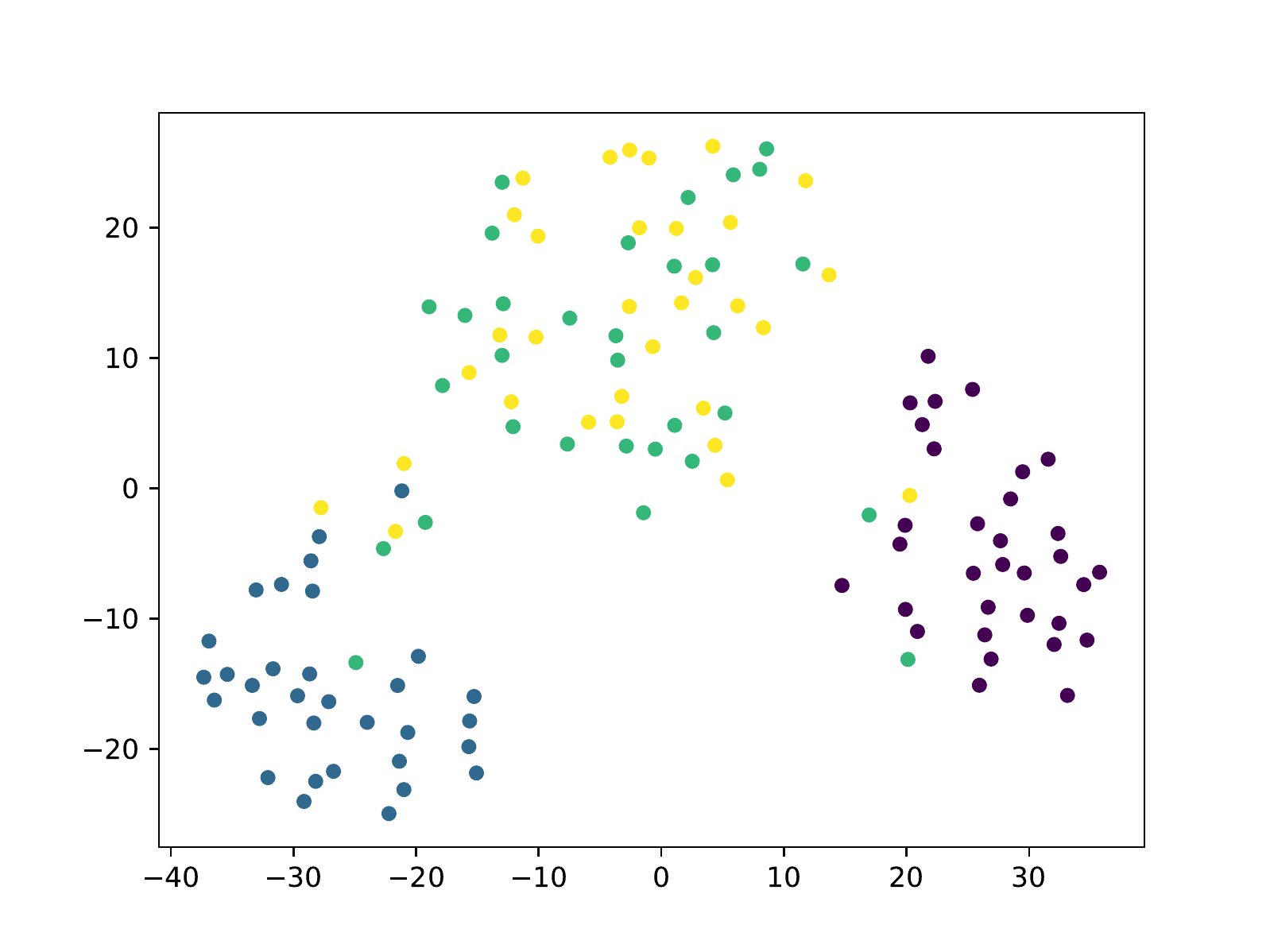}}
\subfigure[GATE]{
\includegraphics[width=0.235\textwidth]{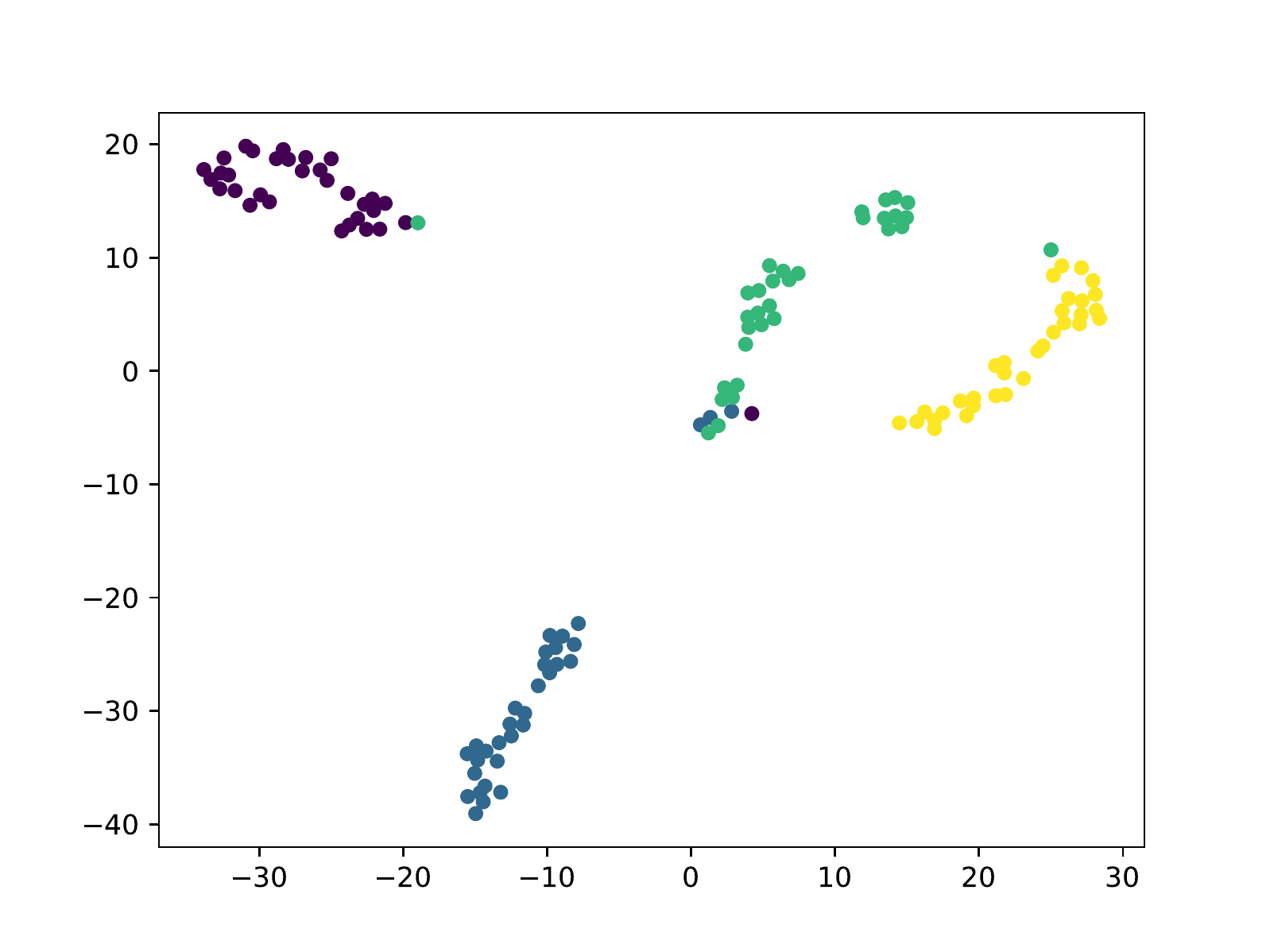}}
\subfigure[ANGM]{
\includegraphics[width=0.235\textwidth]{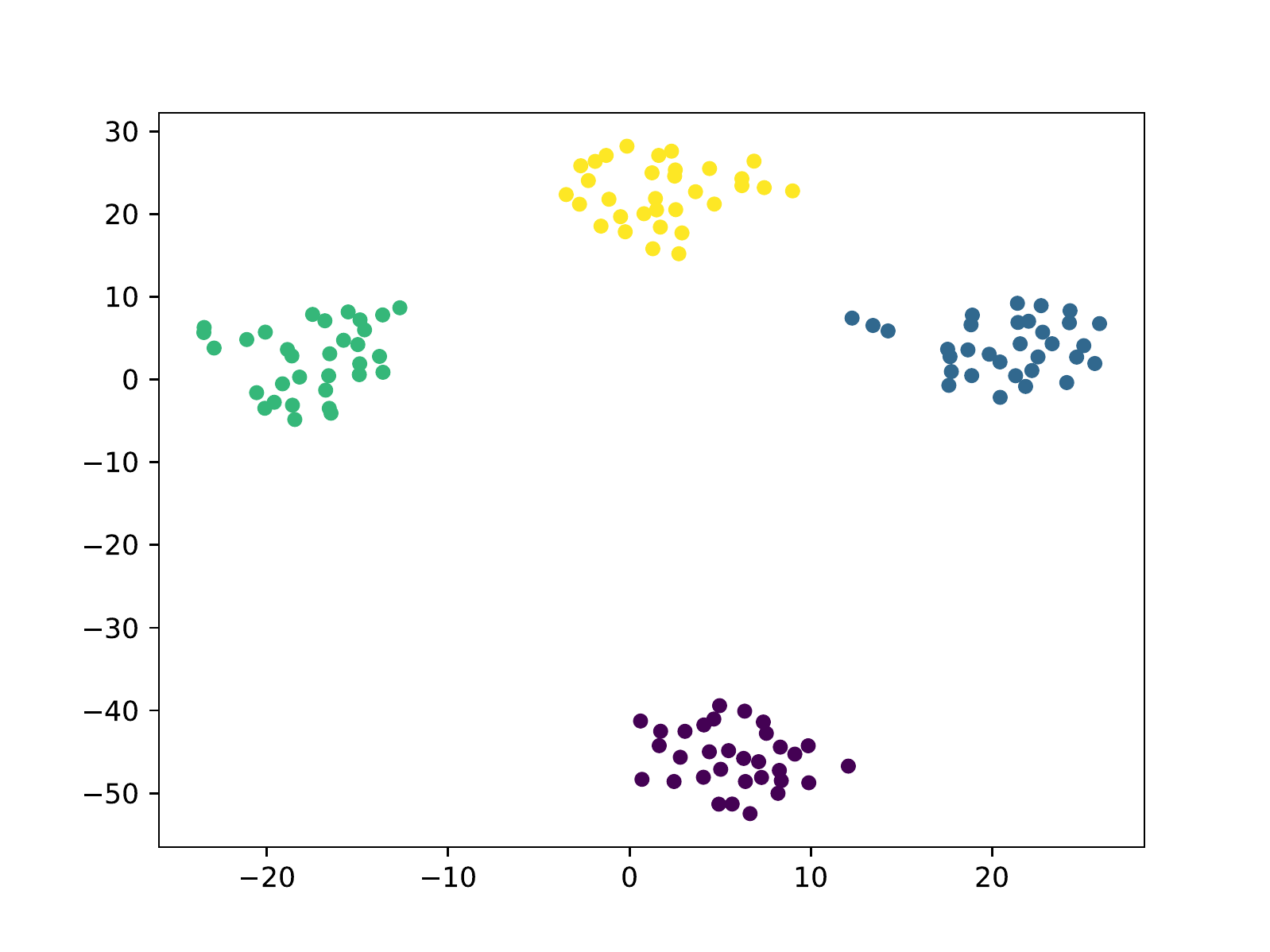}}
\caption{Visualization of representation learned by algorithms attributed networks with hybrid structures.}
\label{fig4}
\end{figure}

We first perform the proposed methods and the baselines on four types of networks and then map the embeddings into 2-dimension space by applying t-SNE \cite{van2014accelerating} and then visualize them as shown in Figures~\ref{fig1}-\ref{fig4}. For t-SNE, the perplexity is set to 10, the number of iterations is 1000. Besides, we use GMM to cluster the nodes. Table~\ref{synthe-clustering:NMI-ACC} shows the clustering NMI and AC of the eight methods on four types of synthetic networks.

\begin{table}[htbp]
\caption{NMI (\%) and ACC (\%) of the methods on node clustering for four synthetic networks}
\centering
\resizebox{\columnwidth}{!}{
\begin{tabular}{ccccccc}
\toprule
Metrics & Method   &   community & multipartite & hub & hybrid \\
\midrule
\multirow{11}{*}{NMI}
& NOBE & 97.48 & 1.80 & 28.35 & 39.31 \\
& Node2Vec  & 90.78 & 7.07  & 51.64 & 57.22 \\
& BANE & 94.96 & 16.48 & 52.97& 50.22 \\
&ASNE      & 94.98 & 86.54 & 60.38 & 97.48\\
&ANRL      & 100   & 95.75 & 94.98   & 100\\
&VGAE      & 100   & 81.54 & 100   & 91.82\\
&ARVGE     & 100   & 57.49 & 92.49 & 78.01\\
& G2G & 92.48 & 1.53 & 92.47 & 61.24\\
& GATE& 100 & 97.48 & 75.83 & 71.38\\
&ANGM  (ours)      & 100   & 100   & 100   & 100\\
\midrule
\multirow{11}{*}{AC}
& NOBE & 99.22 & 30.47 & 41.41 & 39.31\\
&Node2Vec  & 98.88 & 36.72 & 64.06 & 68.75\\
& BANE& 98.44 & 46.09 & 71.88 & 50.03\\
&ASNE      & 97.66 & 89.85 & 72.66 & 99.22\\
&ANRL      & 100   & 98.44 & 98.44   & 100 \\
&VGAE     & 100   & 93.75 & 100   & 96.88\\
&ARVGE    & 100   & 78.91 & 97.66 & 89.85\\
& G2G & 97.66 & 30.47 & 97.66 & 71.09\\
&GATE& 100 & 99.22 & 89.84 & 83.59\\
&ANGM  (ours)      & 100   & 100   & 100   & 100\\
\bottomrule
\end{tabular}}\label{synthe-clustering:NMI-ACC}
\end{table}

From Table \ref{synthe-clustering:NMI-ACC} and Figures \ref{fig1}-\ref{fig4}, we can conclude several observations. (1) ANGM finds all blocks on four types of networks, and both NMI and AC of ANGM  achieve 100\%. Because the parameter $\pmb{\Pi}$ in ANGM is capable of characterizing networks with various structural patterns. (2) NOBE and Node2Vec perform worse than others on most networks, especially on networks with multipartite structures, because they only use the structural topology information but not the attribute information. It indicates that the additional node attributes can help the network representation methods to learn node embeddings with higher quality. (3) Among the four types of networks, most state-of-the-art attributed network embedding algorithms, like BANE, VGAE, ARVGE, and G2G, perform worst on the network with multipartite structures and perform best on that with communities. Since they assume that the attributes propagate based on the links, they are suitable in the case of linked nodes sharing similar embeddings. However, the nodes in different blocks are more likely to connect to each other in networks with multipartite structures.

In summary, ANGM outperforms most of the baselines on these four types of synthetic networks, especially on disassortative networks. It indicates that ANGM can deal with both assortative networks and disassortative networks.

\section{Conclusion}\label{Conclusion}
In this paper, we propose a novel block-based generative model for attributed network representation learning. Accordingly, we introduce ``block'' concept to attributed network embedding methods. The connection patterns related to blocks can define assortative networks with communities as well as disassortative networks with multipartite structures, hubs, or any hybrid of them. Then, we use neural networks to depict the nonlinearity between the node embeddings and the node attributes. The topology information and the attribute information are combined by assuming that the nodes in the same blocks share similar embeddings and similar linkage patterns. Finally, the variational inference is introduced for learning the parameters of the proposed model. Experiments show that our proposed model consistently outperforms state-of-the-art methods on both real-word and synthetic attributed networks with various structural patterns.

So far, the proposed method's time complexity is square to the network scale. In future work, we will try to use a Poisson distribution \cite{gopalan2015scalable} to generate the links between the nodes to reduce the time complexity. Under the Poisson distribution, the time complexity will be relative to the number of edges, which is linear to the network scale in most real-world networks.

\section*{Appendix}
\setcounter{equation}{0}
\renewcommand\theequation{A.\arabic{equation}}
In this section, we give some details for the derivation of likelihood of complete-data and the update rules of the parameters of our model.

\subsection*{\textbf{Derivation of likelihood of complete-data}}
According to the generative process of attributed networks, the joint probability or the likelihood of complete-data is
\begin{equation}
    p(\pmb{X},\pmb{A},\pmb{Z},\pmb{c}|\pmb{\Pi},\pmb{\omega},\pmb{\sigma},\pmb{\mu})=p(\pmb{A}|\pmb{c},\pmb{\Pi})p(\pmb{X}|\pmb{Z})p(\pmb{Z}|\pmb{c},\pmb{\sigma},\pmb{\mu})p(\pmb{C}|\pmb{\omega})\label{app-joint-prob}
\end{equation}
and each factor is defined as follows.

First, we know that the node assignment follows a multinomial distribution. The probability of node $i$ belongs to block $k$ is $\omega_k$ and the assignment for each node is independent. Thus, the probability of assigning all nodes, i.e., obtaining vector $\pmb{c}=<c_1,c_2,...c_n>$, is
\begin{equation}\label{p-C}
  p(\pmb{c}|\pmb{\omega}) = \prod_i\omega_{c_i}.
\end{equation}

Then, the embedding of node $i$ follows a Gaussian distribution with mean $\pmb{\mu}_{c_i}$ and standard derivation $\pmb{\sigma}_{c_i}$ if we know that node $i$ belongs to block $c_i$. Thus, we have
\begin{equation}\label{p-Z}
  p(\pmb{Z}|\pmb{c},\pmb{\sigma},\pmb{\mu}) = \prod_{id}\frac{1}{\sqrt{2\pi}\sigma_{c_id}}e^{-\frac{(z_{id}-\mu_{c_id})^2}{2\sigma_{c_id}^2}}.
\end{equation}
As for the probability of generating node attributes,
if $\pmb{X} \in \{0,1\}^{n\times M}$,  it follows a Bernoulli distribution, i.e., the probability of node $i$ having $m$-th attribute is $\upsilon_{im}$. Thus,
\begin{equation}\label{p-X-b}
  p(\pmb{X}|\pmb{Z}) = \prod_{im}\upsilon_{im}^{x_{im}}(1-\upsilon_{im})^{1-x_{im}},
\end{equation}
Similarly, if $\pmb{X} \in \mathbb{R}^{n\times M}$, we can obtain
\begin{equation}\label{p-X-g}
  p(\pmb{X}|\pmb{Z}) = \prod_{im}\frac{1}{\sqrt{2\pi}\lambda_{im}}e^{-\frac{(x_{im}-\upsilon_{im})^2}{2\lambda_{im}^2}}.
\end{equation}
Finally, generating links between each pair of nodes follows a Bernoulli distribution and the generation process of each pair nodes is independent. The probability of node $i$ connecting to node $j$ is $\pi_{c_ic_j}$ if the node assignment is known. Thus, the probability of generating links is
\begin{equation}\label{p-A}
  p(\pmb{A}|\pmb{c},\pmb{\Pi}) = \prod_{ij}\pi_{c_ic_j}^{a_{ij}}(1-\pi_{c_ic_j})^{1-a_{ij}}.
\end{equation}
Using a network with binary attributes as an example, we substitute Eqs. (\ref{p-C})-(\ref{p-X-b}) and (\ref{p-A}) to Eq.(\ref{app-joint-prob}), we obtain
\begin{equation}
\begin{split}
 &p(\pmb{X},\pmb{A},\pmb{Z},\pmb{c}|\pmb{\Pi},\pmb{\omega},\pmb{\sigma},\pmb{\mu})\\
 &=\prod_{ij}\pi_{c_ic_j}^{a_{ij}}(1-\pi_{c_ic_j})^{1-a_{ij}}\times\prod_{im}\upsilon_{im}^{x_{im}}(1-\upsilon_{im})^{(1-x_{im})}\\
 &\quad\times\prod_{id}\frac{1}{\sqrt{2\pi}\sigma_{c_id}}e^{-\frac{(z_{id}-\mu_{c_id})^2}{2\sigma_{c_id}^2}}\times\prod_i\omega_{c_i}.
\end{split}\label{-app-likelihood}
\end{equation}

\subsection*{\textbf{Derivation of update rules of the parameters}}
First, the items related to $\pmb{\tau}$ on Eq.~(\ref{ELBO}) are:
\begin{equation}
\begin{split}
\mathcal{L}_{[\tau_{ik}]} = &\sum_{j}\sum_{l}\tau_{ik}\tau_{jl}[a_{ij}\log\pi_{kl}+(1-a_{ij})\log(1-\pi_{kl})]\\
& -\frac{1}{2}\sum_{d=1}^D\tau_{ik}(\log\sigma_{kd}^2+\frac{\hat{\sigma}_{id}^2}{\sigma_{kd}^2} +\frac{(\hat{\mu}_{id}-\mu_{kd})^2}{\sigma_{kd}^2})\\
&+ \tau_{ik}\log\frac{\omega_k}{\tau_{ik}}.
\end{split}\notag
\end{equation}
Set $\frac{\partial\mathcal{L}_{[\tau_{ik}]}}{\partial\tau_{ik}}=0$, then we can update $\tau_{ik}$ by
\begin{equation}
\begin{split}
  \tau_{ik} \propto &\exp (\sum_{j}\sum_{l}\tau_{jl}[a_{ij}\log\pi_{kl}+(1-a_{ij})\log(1-\pi_{kl})]\\
& \quad-\frac{1}{2}\sum_{d}^D(\log\sigma_{kd}^2+\frac{\hat{\sigma}_{id}^2}{\sigma_{kd}^2} +\frac{(\hat{\mu}_{id}-\mu_{kd})^2}{\sigma_{kd}^2}) + \log\omega_k).
\end{split}
\end{equation}

Then, we optimize $\pi_{kl}$:
\begin{equation}
\begin{split}
\mathcal{L}_{[\pi_{kl}]} = \sum_{ij}\tau_{ik}\tau_{jl}[A_{ij}\log\pi_{kl}+(1-A_{ij})\log(1-\pi_{kl})].
\end{split}\notag
\end{equation}
Set $\frac{\partial\mathcal{L}_{[\pi_{kl}]}}{\partial\pi_{kl}}=0$, we obtain
\begin{equation}
 \pi_{kl} = \frac{\sum_{ij}\tau_{ik}\tau_{jl}A_{ij}}{\sum_{ij}\tau_{ik}\tau_{jl}}.
\end{equation}

Next, the items related to $\omega_k$ are
\begin{equation}
\mathcal{L}_{[\omega_{k}]}= \sum_{i}\gamma_{ik}\log\omega_{k}.
\end{equation}
Since $\sum_{k=1}^K=1$, we take the derivative of $\mathcal{L}_{[\omega_{k}]} +  \beta(\sum_k\omega_k - 1)$ of $\omega_k$, and make the derivative to zero. Then, we can obtain the update formula for $\omega_k$ as follows:
\begin{equation}
  \omega_k = \frac{1}{n}\sum_{i}\gamma_{ik}.
\end{equation}

In the same way, we can obtain the items related to $\mu_{kd}$ and $\sigma_{kd}$ as follows:
\begin{equation}
  \mathcal{L}_{[\mu_{kd}]}=-\frac{1}{2}\sum_i^n\gamma_{ik}\frac{(\hat{\mu}_{id}-\mu_{kd})^2}{\sigma_{kd}^2},
\end{equation}
and
\begin{equation}
  \mathcal{L}_{[\sigma_{kd}]}=-\frac{1}{2}\sum_i^n\gamma_{ik}(\log\sigma_{kd}^2+\frac{\hat{\sigma}_{id}^2}{\sigma_{kd}^2}+\frac{(\hat{\mu}_{id}-\mu_{kd})^2}{\sigma_{kd}^2}).
\end{equation}
We set $\frac{\partial\mathcal{L}_{[\mu_{kd}]}}{\partial\mu_{kd}}=0$ and $\frac{\partial\mathcal{L}_{[\sigma_{kd}]}}{\partial\sigma_{kd}}=0$, then we derive the update rules for  $\mu_{kd}$ and $\sigma_{kd}$ are
\begin{equation}
 \mu_{kd}= \frac{\sum_i^n\tau_{ik}\hat{\mu}_{id}}{\sum_i^n\tau_{ik}},
\end{equation}
and
\begin{equation}
 \sigma_{kd}= \frac{\sum_i^n\tau_{ik}(\hat{\sigma}_{id}+(\hat{\mu}_{id}-\mu_{kd})^2)}{\sum_i^n\tau_{ik}},
\end{equation}
respectively.

\section*{Acknowledgment}
This work was supported  by the National Natural Science Foundation of China under grant number 61876069; Jilin Province Key Scientific and Technological Research and Development project under grant numbers 20180201067GX, 20180201044GX; Jilin Province Natural Science Foundation under grant number 20200201036JC; China Scholarship Council under grant number 201906170205, 201906170208; Australian Research Council under grant number DP190101087.

\section*{Conflict of interest}

The authors declare that they have no conflict of interest.

\bibliographystyle{spmpsci}
\bibliography{references}

\begin{thebibliography}{10}
\providecommand{\url}[1]{{#1}}
\providecommand{\urlprefix}{URL }
\expandafter\ifx\csname urlstyle\endcsname\relax
  \providecommand{\doi}[1]{DOI~\discretionary{}{}{}#1}\else
  \providecommand{\doi}{DOI~\discretionary{}{}{}\begingroup
  \urlstyle{rm}\Url}\fi

\bibitem{abbe2017community}
Abbe, E.: Community detection and stochastic block models: recent developments.
\newblock The Journal of Machine Learning Research \textbf{18}(1), 6446--6531
  (2017)

\bibitem{barbieri2014follow}
Barbieri, N., Bonchi, F., Manco, G.: Who to follow and why: link prediction
  with explanations.
\newblock In: KDD, pp. 1266--1275 (2014)

\bibitem{bojchevski2018deep}
Bojchevski, A., G{\"u}nnemann, S.: Deep gaussian embedding of graphs:
  Unsupervised inductive learning via ranking.
\newblock In: International Conference on Learning Representations (2018)

\bibitem{craveny1998learning}
Craveny, M., DiPasquoy, D., Freitagy, D., McCallumzy, A., Mitchelly, T.,
  Nigamy, K., an~Slatteryy, S.: Learning to extract symbolic knowledge from the
  world wide web.
\newblock In: AAAI, pp. 509--516 (1998)

\bibitem{gama2019diffusion}
Gama, F., Ribeiro, A., Bruna, J.: Diffusion scattering transforms on graphs.
\newblock In: International Conference on Learning Representations (2019)

\bibitem{gao2018deep}
Gao, H., Huang, H.: Deep attributed network embedding.
\newblock In: IJCAI, pp. 3364--3370 (2018)

\bibitem{gao2019graph}
Gao, H., Ji, S.: Graph u-nets.
\newblock In: Proceedings of the 36th International Conference on Machine
  Learning (2019)

\bibitem{gao2010community}
Gao, J., Liang, F., Fan, W., Wang, C., Sun, Y., Han, J.: On community outliers
  and their efficient detection in information networks.
\newblock In: KDD, pp. 813--822. ACM (2010)

\bibitem{gao2018bine}
Gao, M., Chen, L., He, X., Zhou, A.: Bine: Bipartite network embedding.
\newblock In: SIGIR, pp. 715--724 (2018)

\bibitem{gopalan2015scalable}
Gopalan, P., Hofman, J.M., Blei, D.M.: Scalable recommendation with
  hierarchical poisson factorization.
\newblock In: UAI, pp. 326--335 (2015)

\bibitem{node2vec-kdd2016}
Grover, A., Leskovec, J.: node2vec: Scalable feature learning for networks.
\newblock In: KDD (2016)

\bibitem{guimera2009missing}
Guimer{\`a}, R., Sales-Pardo, M.: Missing and spurious interactions and the
  reconstruction of complex networks.
\newblock PNAS \textbf{106}(52), 22073--22078 (2009)

\bibitem{hamilton2017inductive}
Hamilton, W., Ying, Z., Leskovec, J.: Inductive representation learning on
  large graphs.
\newblock In: Advances in Neural Information Processing Systems, pp. 1024--1034
  (2017)

\bibitem{holland1983stochastic}
Holland, P.W., Laskey, K.B., Leinhardt, S.: Stochastic blockmodels: First
  steps.
\newblock Social networks \textbf{5}(2), 109--137 (1983)

\bibitem{huang2017accelerated}
Huang, X., Li, J., Hu, X.: Accelerated attributed network embedding.
\newblock In: Proceedings of the 2017 SIAM international conference on data
  mining, pp. 633--641 (2017)

\bibitem{huang2019large}
Huang, X., Song, Q., Yang, F., Hu, X.: Large-scale heterogeneous feature
  embedding.
\newblock In: AAAI (2019)

\bibitem{jiang2018spectral}
Jiang, F., He, L., Zheng, Y., Zhu, E., Xu, J., Yu, P.S.: On spectral graph
  embedding: A non-backtracking perspective and graph approximation.
\newblock In: Proceedings of the 2018 SIAM International Conference on Data
  Mining, pp. 324--332. SIAM (2018)

\bibitem{jiang2015stochastic}
Jiang, J.Q.: Stochastic block model and exploratory analysis in signed
  networks.
\newblock Physical Review E \textbf{91}(6), 062805 (2015)

\bibitem{kendrick2018change}
Kendrick, L., Musial, K., Gabrys, B.: Change point detection in social
  networks—critical review with experiments.
\newblock Computer Science Review \textbf{29}, 1--13 (2018)

\bibitem{kingma2013auto}
Kingma, D.P., Welling, M.: Auto-encoding variational bayes.
\newblock ICLR  (2014)

\bibitem{kipf2016variational}
Kipf, T.N., Welling, M.: Variational graph auto-encoders.
\newblock NIPS Workshop on Bayesian Deep Learning  (2016)

\bibitem{kipf2016semi}
Kipf, T.N., Welling, M.: Semi-supervised classification with graph
  convolutional networks.
\newblock ICLR  (2017)

\bibitem{knyazev2019understanding}
Knyazev, B., Taylor, G.W., Amer, M.: Understanding attention and generalization
  in graph neural networks.
\newblock In: Advances in Neural Information Processing Systems, pp. 4202--4212
  (2019)

\bibitem{kuncheva2004using}
Kuncheva, L.I., Hadjitodorov, S.T.: Using diversity in cluster ensembles.
\newblock In: IEEE International Conference on Systems, Man and Cybernetics,
  vol.~2, pp. 1214--1219 (2004)

\bibitem{li2015unsupervised}
Li, J., Hu, X., Tang, J., Liu, H.: Unsupervised streaming feature selection in
  social media.
\newblock In: Proceedings of the 24th ACM International on Conference on
  Information and Knowledge Management, pp. 1041--1050 (2015)

\bibitem{liao2018attributed}
Liao, L., He, X., Zhang, H., Chua, T.S.: Attributed social network embedding.
\newblock IEEE TKDE \textbf{30}(12), 2257--2270 (2018)

\bibitem{mehta2019stochastic}
Mehta, N., Duke, L.C., Rai, P.: Stochastic blockmodels meet graph neural
  networks.
\newblock In: ICML, pp. 4466--4474 (2019)

\bibitem{namata2012query}
Namata, G., London, B., Getoor, L., Huang, B., EDU, U.: Query-driven active
  surveying for collective classification.
\newblock In: 10th International Workshop on Mining and Learning with Graphs,
  vol.~8 (2012)

\bibitem{newman2016structure}
Newman, M.E., Clauset, A.: Structure and inference in annotated networks.
\newblock Nature communications \textbf{7}(1), 1--11 (2016)

\bibitem{pan2018adversarially}
Pan, S., Hu, R., Long, G., Jiang, J., Yao, L., Zhang, C.: Adversarially
  regularized graph autoencoder for graph embedding.
\newblock In: IJCAI, pp. 2609--2615 (2018)

\bibitem{pan2016tri}
Pan, S., Wu, J., Zhu, X., Zhang, C., Wang, Y.: Tri-party deep network
  representation.
\newblock In: IJCAI, pp. 1895--1901 (2016)

\bibitem{ICLR2020GeomGCN}
Pei, H., Wei, B., Chang, K.C.C., Lei, Y., Yang, B.: Geom-gcn: Geometric graph
  convolutional networks.
\newblock In: ICLR (2020)

\bibitem{perozzi2014focused}
Perozzi, B., Akoglu, L., Iglesias~S{\'a}nchez, P., M{\"u}ller, E.: Focused
  clustering and outlier detection in large attributed graphs.
\newblock In: KDD, pp. 1346--1355. ACM (2014)

\bibitem{perozzi2014deepwalk}
Perozzi, B., Al-Rfou, R., Skiena, S.: Deepwalk: Online learning of social
  representations.
\newblock In: KDD, pp. 701--710 (2014)

\bibitem{pillai2012f}
Pillai, I., Fumera, G., Roli, F.: F-measure optimisation in multi-label
  classifiers.
\newblock In: Proceedings of the 21st International Conference on Pattern
  Recognition (ICPR2012), pp. 2424--2427 (2012)

\bibitem{ribeiro2017struc2vec}
Ribeiro, L.F., Saverese, P.H., Figueiredo, D.R.: struc2vec: Learning node
  representations from structural identity.
\newblock In: KDD, pp. 385--394 (2017)

\bibitem{ruan2013efficient}
Ruan, Y., Fuhry, D., Parthasarathy, S.: Efficient community detection in large
  networks using content and links.
\newblock In: WWW, pp. 1089--1098. ACM (2013)

\bibitem{salehi2019graph}
Salehi, A., Davulcu, H.: Graph attention auto-encoders.
\newblock arXiv preprint arXiv:1905.10715  (2019)

\bibitem{silva2018social}
Silva, T.H., Laender, A.H., de~Melo, P.O.V.: Social-based classification of
  multiple interactions in dynamic attributed networks.
\newblock In: 2018 IEEE International Conference on Big Data (Big Data), pp.
  4063--4072 (2018)

\bibitem{van2014accelerating}
Van Der~Maaten, L.: Accelerating t-sne using tree-based algorithms.
\newblock The Journal of Machine Learning Research \textbf{15}(1), 3221--3245
  (2014)

\bibitem{velivckovic2017graph}
Veli{\v{c}}kovi{\'c}, P., Cucurull, G., Casanova, A., Romero, A., Lio, P.,
  Bengio, Y.: Graph attention networks.
\newblock ICLR  (2018)

\bibitem{wahid2019predict}
Wahid-Ul-Ashraf, A., Budka, M., Musial, K.: How to predict social
  relationships—physics-inspired approach to link prediction.
\newblock Physica A: Statistical Mechanics and its Applications \textbf{523},
  1110--1129 (2019)

\bibitem{xu2018graph}
Xu, B., Shen, H., Cao, Q., Qiu, Y., Cheng, X.: Graph wavelet neural network.
\newblock In: International Conference on Learning Representations (2018)

\bibitem{xu2003document}
Xu, W., Liu, X., Gong, Y.: Document clustering based on non-negative matrix
  factorization.
\newblock In: SIGIR, pp. 267--273 (2003)

\bibitem{yang2011characterizing}
Yang, B., Liu, J., Liu, D.: Characterizing and extracting multiplex patterns in
  complex networks.
\newblock IEEE Transactions on Systems, Man, and Cybernetics, Part B
  (Cybernetics) \textbf{42}(2), 469--481 (2011)

\bibitem{yang2012characterizing}
Yang, B., Liu, J., Liu, D.: Characterizing and extracting multiplex patterns in
  complex networks.
\newblock IEEE transactions on systems, man, and cybernetics. Part B,
  Cybernetics: a publication of the IEEE Systems, Man, and Cybernetics Society
  \textbf{42}(2), 469 (2012)

\bibitem{yang2017stochastic}
Yang, B., Liu, X., Li, Y., Zhao, X.: Stochastic blockmodeling and variational
  bayes learning for signed network analysis.
\newblock IEEE TKDE \textbf{29}(9), 2026--2039 (2017)

\bibitem{yang2018binarized}
Yang, H., Pan, S., Zhang, P., Chen, L., Lian, D., Zhang, C.: Binarized
  attributed network embedding.
\newblock In: 2018 IEEE International Conference on Data Mining (ICDM), pp.
  1476--1481. IEEE (2018)

\bibitem{yang2018enhanced}
Yang, S., Yang, B.: Enhanced network embedding with text information.
\newblock In: 2018 24th International Conference on Pattern Recognition (ICPR),
  pp. 326--331. IEEE (2018)

\bibitem{yang2011detecting}
Yang, T., Chi, Y., Zhu, S., Gong, Y., Jin, R.: Detecting communities and their
  evolutions in dynamic social networks--a bayesian approach.
\newblock Machine learning \textbf{82}(2), 157--189 (2011)

\bibitem{ijcai2018-438}
Zhang, Z., Yang, H., Bu, J., Zhou, S., Yu, P., Zhang, J., Ester, M., Wang, C.:
  Anrl: Attributed network representation learning via deep neural networks.
\newblock In: IJCAI, pp. 3155--3161 (2018)

\end{thebibliography}

\end{document}